\documentclass{article}

\usepackage[preprint]{neurips_2026}

\usepackage[utf8]{inputenc} 
\usepackage[T1]{fontenc}    
\usepackage{hyperref}       
\usepackage{url}            
\usepackage{booktabs}       
\usepackage{amsfonts}       
\usepackage{nicefrac}       
\usepackage{microtype}      
\usepackage{xcolor}         
\usepackage{graphicx}
\usepackage{enumitem}
\usepackage{amsmath}
\usepackage{array}
\usepackage{wrapfig}
\usepackage{floatflt}
\usepackage{lipsum}
\usepackage{booktabs}
\usepackage{multirow}
\usepackage{makecell}
\usepackage{xspace}
\usepackage[most]{tcolorbox}
\usepackage{listings}
\usepackage{subcaption}
\usepackage{rotating}
\hypersetup{
    colorlinks=true,
    linkcolor=black,
    citecolor=black,
    filecolor=black,
    urlcolor=black,
}

\usepackage{booktabs}
\usepackage{multirow}

\newcommand{\best}[1]{\textbf{\boldmath #1}}   
\newcommand{\second}[1]{\underline{#1}}
\definecolor{seagreen}{RGB}{46, 139, 87}

\definecolor{CustomGreen}{HTML}{1DA35F}
\newcommand{\framework}{{\textsc{Flame}}\xspace}
\newcommand{\solutionrationale}{solution rationale\xspace}
\newcommand{\geminipro}{Gemini 3.1 Pro\xspace}
\newcommand{\geminiflashlite}{Gemini 2.5 Flash Lite\xspace}
\newcommand{\claudeopus}{Claude Opus 4.6\xspace}
\newcommand{\claudehaiku}{Claude Haiku 4.5\xspace}
\newcommand{\gptlarge}{GPT-5.4\xspace}
\newcommand{\gptmini}{GPT-4.1-mini\xspace}

\newcommand{\qwenmed}{Qwen3-8B\xspace}
\newcommand{\qwenlarge}{Qwen3-32B\xspace}
\newcommand{\gemmamed}{Gemma3-12b-it\xspace}
\newcommand{\gemmalarge}{Gemma3-27b-it\xspace}
\newcommand{\deepseekmed}{DeepSeek-R1-Distill-Qwen-14b\xspace}
\newcommand{\deepseeklarge}{DeepSeek-R1-Distill-Qwen-32b\xspace}

\title{Fine-Grained Benchmark Generation for Comprehensive Evaluation of Foundation Models}

\author{%
  Mohammed Saidul Islam$^{1}$\,\,
  Negin Baghbanzadeh$^{1}$\,\,
  Farnaz Kohankhaki$^{1}$ \,\,
  \textbf{Afshin Cheraghi}$^{2}$ \\[0.5em]
  \textbf{Ali Kore}$^{1}$ \,\,
  \textbf{Shayaan Mehdi}$^{1}$ \,\,
  \textbf{Elham Dolatabadi}$^{1,2}$ \,\,
  \textbf{Arash Afkanpour}$^{1}$\footnotemark[2] \\[0.5em]
  $^1$Vector Institute \qquad
  $^2$York University
}

\begin{document}
\renewcommand{\thefootnote}{}
\footnotetext{\textsuperscript{$\dagger$} Corresponding author: \texttt{arash.afkanpour@vectorinstitute.ai}}

\maketitle

\begin{abstract}
Evaluation of foundation models often rely on aggregate scores from benchmarks that lack comprehensive coverage and metadata for a fine-grained evaluation. We introduce a framework for automated benchmark generation. Our framework generates evaluation problems grounded in reference material, such as textbooks, producing benchmarks with broad coverage, rich metadata, and robustness to contamination.
The pipeline employs a multi-agent architecture for problem generation and a solution-graph-driven strategy that significantly improves the reliability of ground truth solutions.
Using the framework, we generate three benchmarks in Machine Learning, Corporate Finance, and Personal Finance. Expert review finds a significantly lower ground-truth error rate than previous benchmarks such as MMLU and GSM8K. Evaluation of 12 commercial and open-source  models shows that our benchmarks achieve near-uniform competency coverage and surface performance differences across models that existing benchmarks fail to capture. We will open-source the framework and our curated benchmarks soon.
\end{abstract}

\section{Introduction}



Foundation models have demonstrated remarkable and rapidly advancing capabilities across a broad spectrum of domains, from natural language understanding and code generation to mathematical reasoning and scientific problem solving. This progress, coupled with the proliferation of both commercial and open-source models spanning a wide range of scales, has driven widespread adoption across many industry applications. In many cases, however, model selection is guided by coarse-grained, aggregate performance metrics derived from a limited set of established benchmarks. This approach neglects critical differences in model capabilities at the granularity important for informed decision-making.

Improvements in foundation models have outpaced the pace of developing evaluation benchmarks. Existing benchmarks, including recent efforts such as Humanity's Last Exam \citep{phan2025lastexam}, SuperGPQA \citep{du2025supergpqa}, and MMLU-Pro \citep{wang2024mmlu} have made valuable contributions by curating challenging problems intended to probe the frontiers of model capabilities. However, these benchmarks exhibit several structural limitations that hinder their utility for fine-grained and comprehensive evaluation. First, in each domain they typically consist of relatively small, sparsely sampled sets of problems that, while individually difficult, do not provide adequate coverage over the space of capabilities within the domain. When viewed through the lens of the underlying capability distribution, large regions remain unrepresented, meaning that evaluation on such benchmarks yields an incomplete picture of a model's true competency profile. Second, these benchmarks are frequently designed to produce a single aggregate score or a small number of high-level summary statistics, lacking the rich metadata and taxonomic structure necessary to attribute performance to specific skills, concepts, or reasoning abilities. As a result, existing benchmarks can show that one model outperforms another overall, but cannot identify which specific competencies drive the difference. This limits their utility for model development, where developers need to know which areas to improve, and for model selection, where practitioners need to choose the best model for a specific application.

Beyond coverage and granularity, existing benchmarks face additional well-documented challenges. Static benchmark datasets are susceptible to data contamination, as benchmark tasks are used in pretraining and fine-tuning of foundation models over time, eroding the validity of measured performance \citep{zhang2024careful, deng2024investigating, golchin2023time}. Moreover, high quality benchmark construction is time consuming and requires expensive curation processes. Such processes scale poorly and cannot keep pace with rapid development of foundation models.
A comprehensive discussion of related work is provided in Appendix~\ref{sec:related_work}.

In this work, we introduce \textbf{F}ine-grained, \textbf{LA}rge-scale \textbf{M}odel \textbf{E}valuation (\framework), a framework for automated, comprehensive benchmark generation grounded in external knowledge sources such as textbooks and technical references. Our framework addresses the aforementioned limitations. By systematically generating evaluation tasks from authoritative sources with well-defined topical and conceptual structure, \framework enables the generation of benchmarks that provide broad and verifiable coverage over the competency space of a target domain. Each generated item is associated with rich, structured metadata enabling fine-grained evaluation that localizes model strengths and weaknesses at the level of individual competencies. Because the generation process is automated, benchmarks can be constructed reproducibly and at scale, and can be regenerated to mitigate contamination risks. We demonstrate that benchmarks produced by \framework yield evaluation profiles that are substantially more informative than those obtained from existing benchmarks, enabling model developers to discover strengths and weaknesses of their models at a fine-grained level, and practitioners to make informed decisions about model selection and deployment.

Our contributions are as follows:
\begin{itemize}
    \item We introduce \framework, a framework for automated generation of comprehensive, fine-grained evaluation benchmarks grounded in  external knowledge sources. By generating evaluation tasks grounded in references such as textbooks, \framework produces benchmarks with broad capability coverage, rich metadata, and robustness to data contamination.

    \item Using \framework, we generate and open-source three comprehensive benchmarks in \emph{Machine Learning}, \emph{Corporate Finance}, and \emph{Personal Finance}. The reference books are selected and the benchmarks are reviewed by domain experts to validate their comprehensiveness, coverage, diversity, and correctness of problem statements and solutions.


    \item We evaluate a diverse set of foundation models on the resulting benchmarks and demonstrate that \framework reveals fine-grained differences across specific competencies and skills that are not surfaced by existing benchmarks. We compare the \framework-generated benchmarks against prior benchmarks on aspects such as knowledge coverage, discriminative power to distinguish subject models of various capability, and robustness to data contamination.
\end{itemize}

\begin{figure}
    \centering
    \includegraphics[width=\linewidth]{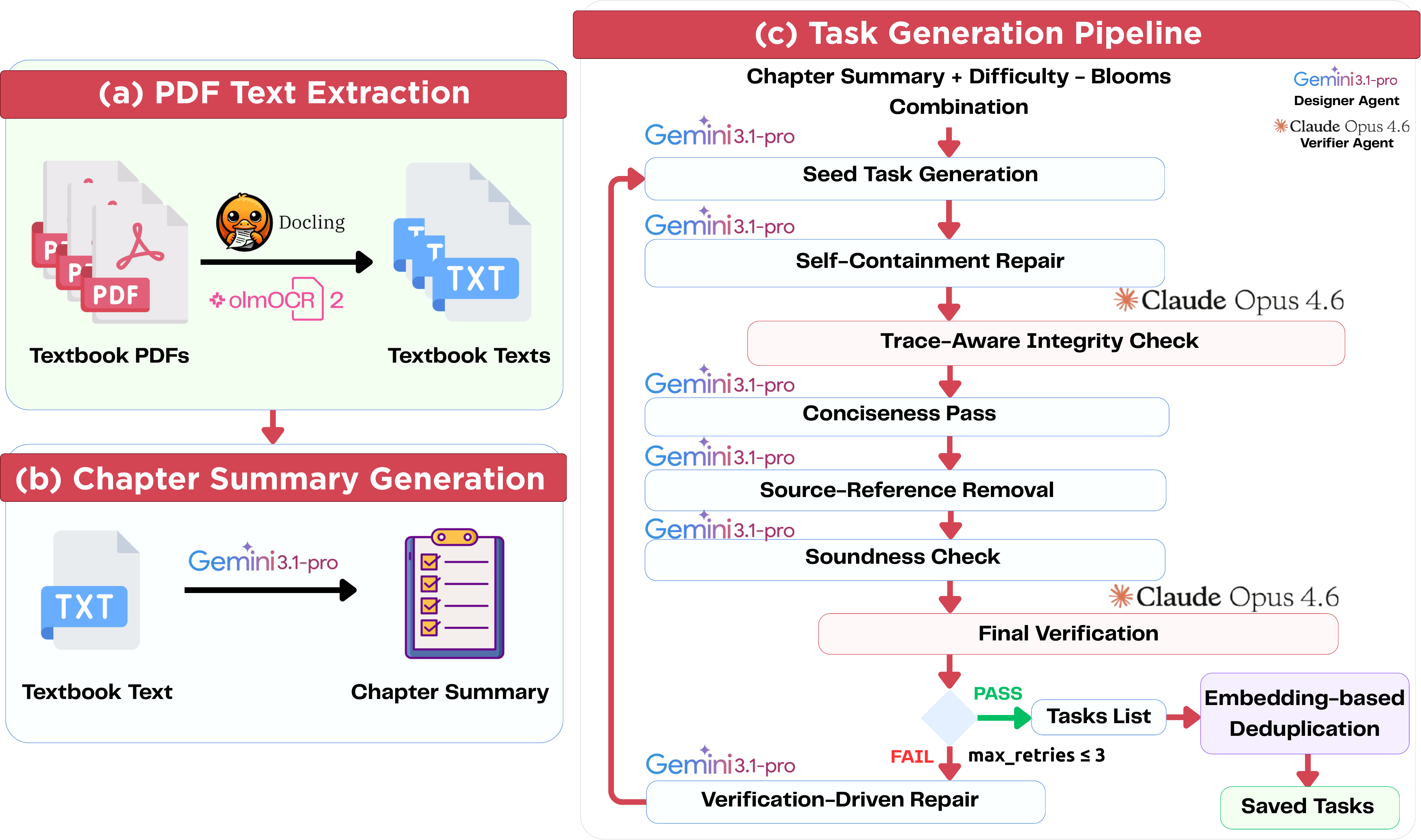}
    \caption{An overview of the major components of the \framework Framework, such as, source content preprocessing that includes: (a) PDF text extraction, (b) Chapter summary generation, and (c) Task Generation Pipeline.}
    \label{fig:flame}
\end{figure}

\section{\framework Framework}

The \framework benchmark generation pipeline consists of two stages: \textit{(i) source curation and preprocessing}, and \textit{(ii) task generation}. Figure~\ref{fig:flame} shows an overall view. Below we describe each stage in detail.

\subsection{Source Curation and Preprocessing}
\label{sec:curation}

The first stage of the pipeline prepares the domain knowledge that grounds problem generation. Given a target domain, a domain expert selects a collection of external sources, such as textbooks, that collectively provide comprehensive coverage of the domain knowledge. The expert then reviews the selected sources to identify the relevant sections for problem generation.

From these sources, we construct a two-level taxonomy of competencies in consultation with the domain expert. Textbooks are naturally organized into parts and chapters that reflect a decomposition of domain knowledge: major parts map to \emph{high-level areas}, and individual chapters map to \emph{competencies} within each area. This taxonomy serves as the backbone of the benchmark: each generated problem is linked to a specific competency and area, enabling evaluation at different levels of granularity.
When multiple textbooks are selected, their structures may partially overlap. The domain expert reviews the merged taxonomy to resolve such overlaps and verify that it represents the competency space. This step only takes a few hours of expert time, since it builds on the  structure already present in the source material.


After curating the taxonomy, source documents are converted to plain text and
segmented by chapter. We automatically remove ancillary content such as prefaces, tables of contents, and end-of-chapter exercises. Exercise sections are
intentionally excluded to avoid verbatim reproduction of existing problems and to reduce bias toward textbook exercises.
We then manually verify the quality and completeness of the extracted text. Implementation details of the extraction pipeline are provided in Appendix~\ref{app:preprocessing}.

\subsection{Task Generation}
\label{sec:task_generation}

Given the preprocessed chapter text and the structured competency taxonomy from the curation stage, the task generation pipeline produces multiple-choice questions (MCQs) grounded in each chapter's content. We adopt the MCQ format because it enables deterministic, automated evaluation without requiring an LLM judge. This avoids the subjectivity and inconsistency that affect evaluation of open-ended responses~\citep{zheng2023judging}. The pipeline is implemented as a multi-agent system with two specialized roles: a \emph{designer agent} responsible for generating and refining candidate problems, and a \emph{verifier agent} responsible for assessing correctness, coherence, and constraint compliance. We separate generation from verification because prior work has shown that decoupling these roles improves output reliability and enables targeted iterative repair~\citep{madaan2023selfrefine, shinn2023reflexion}. For generating our benchmarks, we use \geminipro and \claudeopus as the underlying models for the designer and verifier agents, respectively.

The pipeline proceeds through three phases: (i)~knowledge structuring, which prepares a compact representation of the chapter to guide generation; (ii)~seed task generation, which produces initial candidate problems via a solution-graph-driven strategy; and (iii)~iterative refinement and verification, which applies a sequence of targeted repair and quality-control stages until each candidate either passes all checks or is discarded. We describe each phase below.

\paragraph{Knowledge structuring.}
Before generating any problem, the pipeline constructs a structured summary of the chapter. The designer agent is prompted to extract key elements from the chapter text, including core concepts and definitions, rules and procedures, algorithms, theorems, etc., and to organize them into a dependency graph that captures prerequisite and compositional relationships among these elements. This summary serves two purposes: it provides the generation and verification agents with a compact, explicitly structured view of the chapter's content, and it enables the pipeline to encourage problems that draw on multiple interrelated concepts rather than testing isolated facts. Additional details has been provided in  Appendix~\ref{app:knowledge_structuring}.

\paragraph{Bloom's taxonomy and difficulty control.}
To ensure diversity in the depth of reasoning required by generated problems, we incorporate Bloom's taxonomy of cognitive skills~\citep{anderson2001taxonomy} into the generation prompts. Bloom's taxonomy organizes cognitive demands into progressively more complex levels, from basic recall to higher-order skills such as application, analysis, evaluation, and creation. We pair each Bloom's level with a difficulty setting (easy, medium, hard), but restrict combinations to meaningful ones: lower Bloom's levels are paired only with easy difficulty, while higher Bloom's levels are paired with medium or hard difficulty. A description of each Bloom's level and difficulty setting is included in the generation prompt to guide the designer agent. In preliminary experiments, problems generated at easy and medium difficulty were overwhelmingly straightforward (for instance, recalling definitions) and provided little discriminative signal across subject models. We therefore restrict the final benchmark to hard-difficulty problems aligned with higher-order cognitive levels, yielding four target categories: \emph{Hard--Apply}, \emph{Hard--Analyze}, \emph{Hard--Evaluate}, and \emph{Hard--Create}. The full set of valid (Bloom's level, difficulty) pairings is listed in Appendix~\ref{app:bloom_pairings}.

\paragraph{Solution-graph-driven problem generation.}
A key design choice in our pipeline is how problems are
constructed. A natural approach is to prompt the designer agent to
directly generate a question from the chapter text. However, we found
that this strategy often produces problems that are
superficially related to the chapter material. Instead, we adopt a \emph{solution-graph-driven} generation strategy,
inspired by recent work that constructs computational graphs before
formulating problem
statements~\citep{zhou2025gsm,chou2025autocodebench}. Rather than
generating a question directly, the designer agent first constructs a
multi-step \emph{solution trace}, a directed acyclic graph
 in which each node represents a partial solution or
conclusion (e.g., a computed quantity, a verified condition, or an
evaluated claim), and each edge represents an operation that applies a specific concept, definition, or algorithm from the chapter to
transform one intermediate result into the next. The agent then generates a self-contained
question whose solution follows the constructed graph. By building the
reasoning structure first, the pipeline ensures that each problem is
anchored in specific chapter concepts and requires the solver to
execute a concrete chain of operations.
Examples of generated problems and solution graphs are provided in
Appendix~\ref{app:solution_graph_examples}.
 
This approach offers two additional advantages. First, it increases the reliability of ground-truth annotations: because the solution trace is constructed \emph{before} the question, the verifier agent need only check whether each step in the trace is logically valid and whether the answer follows from it. This verification task is substantially easier and more reliable than solving the problem from scratch~\citep{cobbe2021training,lightman2023let}. Second, the  graph structure of the solution trace provides a mechanism for controlling problem difficulty: increasing the depth or width of the graph yields more complex problems. In practice, we found that the default generation process already produces challenging problems for current models, and we did not extensively explore this direction.
To encourage diversity and challenge, we provide a small number of exemplar problems from an existing established benchmark, e.g., Humanity's Last Exam (HLE)~\citep{phan2025lastexam},  as few-shot demonstrations. The resulting candidate, consisting of the problem statement, answer options, correct answer label, and solution trace, is then passed to the refinement stages.

\paragraph{Iterative refinement and verification.}
Each seed problem undergoes a sequence of refinement and verification stages before it is added to the benchmark. These stages are designed to address failure modes that we observed during development, and they alternate between the designer agent (for repair) and the verifier agent (for assessment):
\begin{enumerate}
    \item \textbf{Self-containment repair.} The designer agent inspects the candidate for undefined notation, implicit assumptions, or missing information that would make the problem ambiguous or unsolvable without access to the source chapter. Identified gaps are repaired while avoiding unnecessary elaboration of standard domain knowledge.

    \item \textbf{Trace-aware integrity check.} The verifier agent receives the candidate alongside the chapter text, the structured knowledge summary, and the solution trace. It evaluates whether (a)~the solution trace is logically correct, (b)~the correct answer follows from the trace, (c)~each distractor is incorrect, and (d)~the problem is grounded in the chapter material. When any condition is violated, the verifier produces a structured diagnostic and the designer agent applies a targeted repair.

    \item \textbf{Conciseness pass.} The designer agent removes redundant wording, such as restating definitions that are not necessary for solving the problem, to make each question concise and focused without altering its meaning or cognitive demand.

    \item \textbf{Source-reference removal.} The designer agent eliminates any explicit references to the originating chapter, section numbers, or textbook structure (e.g., ``according to the chapter''), ensuring that the final question is self-contained and free from visible source scaffolding.

    \item \textbf{Soundness check.} This step verifies that the question is coherent, grammatically clear, complete, and free of contradictions. Unlike the trace-aware integrity check this stage is concerned with the readability and quality of the candidate.

    \item \textbf{Final verification.} The verifier agent performs an end-to-end assessment of the fully refined candidate across four dimensions: output format validity, multiple-choice integrity (unique correct answer, plausible distractors), Bloom's taxonomy alignment, and overall constraint compliance. The verifier returns a structured accept/reject decision with a detailed diagnostic.

    \item \textbf{Verification-driven repair.} When a candidate fails final verification, the pipeline does not discard it immediately. Instead, it routes the verifier's diagnostic back to the designer agent for targeted repair. For formatting errors, only the output structure is modified. For substantive issues, the designer agent may revise the question, options, or correct answer while preserving the underlying concept and Bloom's level. The repaired candidate re-enters the pipeline at step~1 and iterates until it passes or a maximum retry count is reached. Detailed verification statistics is provided in Appendix~\ref{app:verification_stats}.
\end{enumerate}

\paragraph{Deduplication.}
To improve diversity within each chapter, the pipeline employs two  deduplication mechanisms. First, during generation, all previously accepted questions from the same chapter are included in the prompt with an explicit instruction to avoid near-duplicates. Second, after generation for a chapter is complete, an embedding-based filter removes any remaining pair of questions whose cosine similarity exceeds a predefined threshold.

In summary, the task generation pipeline operates as a generate-then-verify process: candidate problems are grounded in chapter content via solution traces, iteratively refined through alternating repair and verification stages, and deduplicated to ensure diversity. Full details of the pipeline and prompts are available in Appendices~\ref{app:task_generation} and~\ref{app:pipeline_prompts} respectively.

\section{Experiments}
To evaluate \framework we perform several studies that aim to address the following research questions:

\begin{itemize}
    \item[\textbf{RQ1}] \textit{Correctness}: Are the problems generated by \framework correct, unambiguous, and of high quality for reliable evaluation? (\S\ref{sec:verification})
    \item[\textbf{RQ2}] \textit{Coverage and Discriminative Power}: Does \framework produce benchmarks with broader coverage, appropriate difficulty, and strong discriminative power compared to existing benchmarks? (\S\ref{sec:coverage})
    \item[\textbf{RQ3}] \textit{Fine-Grained Profiling}: Does evaluation with \framework-generated benchmarks reveal meaningful differences across models that are not captured by aggregate scores? (\S\ref{sec:profiling})
    \item[\textbf{RQ4}] \textit{Contamination Resistance}: Do \framework-generated benchmarks mitigate the inflated performance associated with data contamination in prior benchmarks? (\S\ref{sec:contamination})
\end{itemize}

\subsection{Experimental Setup}

\paragraph{Benchmark generation.} We generate benchmarks in two domains: Machine Learning (ML) and Finance.
For ML, we ground problem generation in the following textbooks: \textit{Foundations of Machine Learning}~\citep{mohri2018foundations}, \emph{Probabilistic Machine Learning: An introduction}~\citep{murphy2022probabilistic}, and \textit{Understanding Deep Learning}~\citep{prince2023understanding}.
For Finance, we generate two specialized benchmarks: a Corporate Finance benchmark grounded in \emph{Corporate Finance} by Ivo Welch~\citep{welch2014corporate} and a Personal Finance benchmark based on the material in \textit{Strategic Financial Planning over the Lifecycle: A Conceptual Approach to Personal Risk Management}~\citep{charupat2012strategic}. All reference textbooks are selected by a domain expert to have a comprehensive coverage of all important topics in the domain.
In each domain, we generate problems at four target categories corresponding to hard problems and higher-order Bloom cognitive levels: \emph{Apply}, \emph{Analyze}, \emph{Evaluate}, and \emph{Create}. Details of benchmark generation cost is provided in Appendix~\ref{app:benchmark_gen_cost}.

\paragraph{Subject models.} We evaluate a diverse set of recent foundation models spanning commercial and open-source families. Commercial models includes \geminipro~\citep{google2026gemini31propreview}, \claudeopus~\citep{anthropic2026claude46}, \gptlarge~\citep{openai2026gpt54}, \geminiflashlite, \claudehaiku, \gptmini, and open-source models include \qwenmed, \qwenlarge~\citep{yang2025qwen3}, \gemmamed, \gemmalarge~\citep{gemmateam2025gemma3}, and \deepseekmed, \deepseeklarge~\citep{deepseekai2025deepseekr1}.
Models were selected to span the performance spectrum from strong frontier to smaller open-source models, enabling us to assess the discriminative power of the benchmarks across a wide range of competencies. Details of subject model inference cost are provided in Appendix~\ref{app:inference_cost}.

\paragraph{Existing benchmarks for comparison.} In each domain we selected prior benchmarks that are suitable for evaluating competencies of models. In Finance, these benchmarks include xFinBench~\citep{zhang2025xfinbench}, FinanceMath~\citep{zhao2024financemath}, and BizBench~\citep{krumdick2024bizbench}. In ML, it is hard to find such benchmarks. Therefore, we selected AI and ML subsets of MMLU-Pro~\citep{wang2024mmlu} and Humanity's Last Exam (HLE)~\citep{phan2025lastexam}. For the competency coverage study, we map each benchmark's problems to the corresponding taxonomy competencies.

All models are evaluated under a standard zero-shot multiple-choice protocol. For each problem, the model is presented with the problem statement and answer options, and must select the correct option. Accuracy is the primary performance metric. We set the temperature to zero for all subject models (when possible) for reproducibility.

\subsection{Manual Problem-Solution Inspection (RQ1)}
\label{sec:verification}

We conduct expert inspection of the generated benchmarks to assess
correctness and quality. For each benchmark, a domain expert solves a
sample of problems independently, compares their answer to the generated
ground truth, and inspects the \solutionrationale for soundness. Our domain experts are a PhD graduate in ML and a PhD student in Finance. Expert inspection required 58 hours for ML and 60 hours for Finance, averaging approximately 24 minutes per problem. We
inspect two samples per benchmark: a uniform random sample of 10\% of
problems, and an \emph{incorrectly-solved} random sample consisting of problems
that at least one of our two strongest models (\geminipro or
\claudeopus) answered incorrectly. The rationale for this second sample is that if a problem has an incorrect ground-truth label, strong models are more likely to
select the actual correct answer and thus appear to have solved the problem incorrectly. This strategy follows the principle that model predictions 
can be used to flag likely mislabeled examples for targeted human 
review~\citep{nahum2025llms}.
\begin{table}[b]
\centering
\small
\caption{Expert inspection results. \textit{Random Sample}: incorrect
ground-truth solutions in a uniform subset. \textit{Incorrectly-Solved
Sample}: incorrect ground-truth solutions among problems that at least
one frontier model answered incorrectly. Counts shown as
incorrect\,/\,total inspected.}
\label{tab:expert_inspection}
\begin{tabular}{@{}lccc@{}}
\toprule
\textbf{Domain} & \textbf{Size} & \textbf{Random Sample} & \textbf{Incorrectly-Solved Sample} \\
\midrule
Machine Learning      & 970 & 3 / 100 & 1 / 15 \\
Corporate Finance     & 685 & 0 / 80  & 0 / 47 \\
Personal Finance      & 199 & 0 / 22  & 1 / 26 \\
\bottomrule
\end{tabular}
\end{table}

Table~\ref{tab:expert_inspection} shows the results of expert inspection. Error rates are low across all benchmarks: 3\% in ML and 0\% in both
Finance benchmarks on the random sample. These rates are lower than documented error
rates in human-curated benchmarks, such as 5\% in
GSM8K~\citep{vendrow2025large} and 6.5\% in
MMLU~\citep{gema2025we}, suggesting that the solution-graph-driven
generation strategy produces reliable ground-truth labels. In addition, the low error rates on the incorrectly-solved samples increase our confidence in the overall correctness of the benchmarks.

The expert review also found all problems relevant to their source chapters, with
clear problem statements, plausible distractors, and no duplicates.
Additional details on the inspection protocol and qualitative findings
are provided in Appendix~\ref{app:expert_inspection}.

\subsection{Coverage, Difficulty, and Discriminative Power Comparison (RQ2)}
\label{sec:coverage}
\begin{figure}[t]
  \centering
  \includegraphics[width=0.95\linewidth]{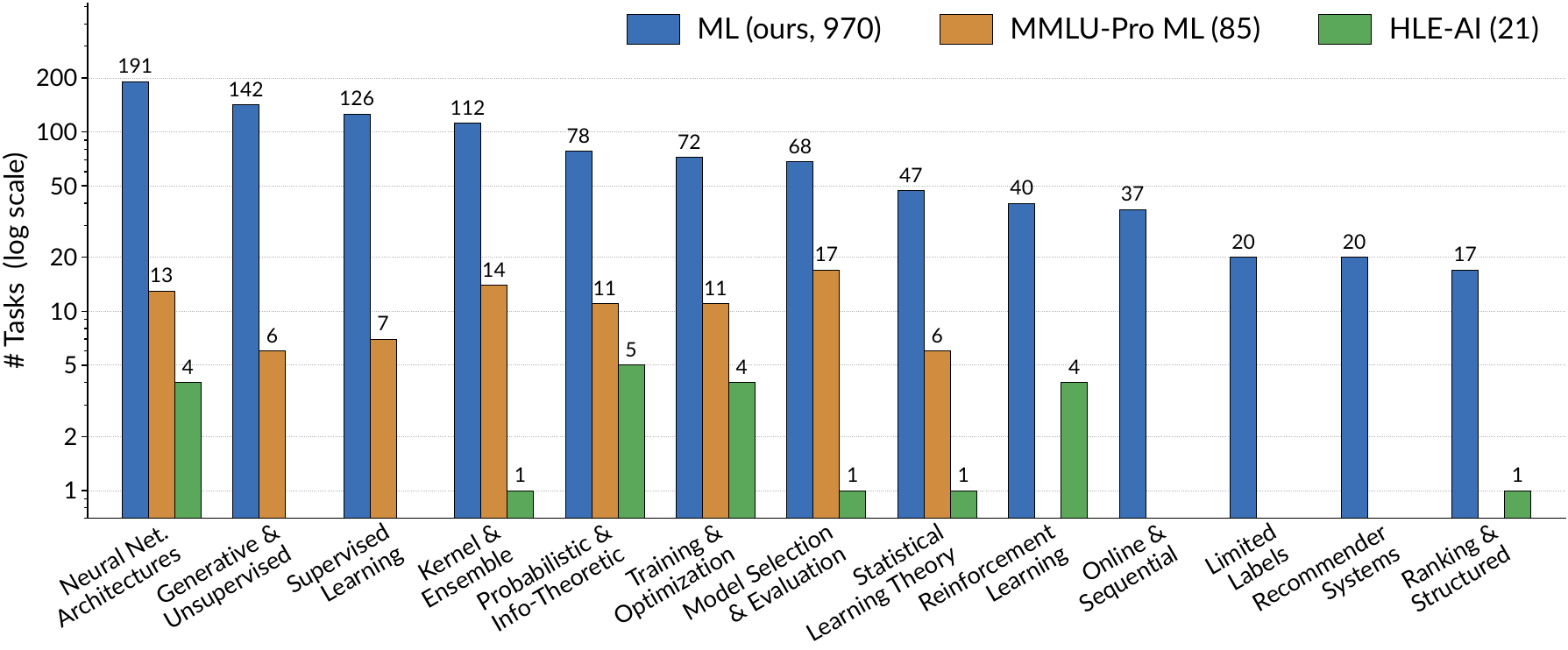}
  \caption{Task counts per area for each dataset in the ML domain (log scale on the $y$-axis.)}
  \label{fig:area_histogram_ml}
\end{figure}

We compare coverage of the competency space between \framework-generated benchmarks and existing benchmarks. By design, \framework-generated benchmarks should provide a broad coverage across competencies. To study existing benchmarks, we map each problem to area and competency via a two-stage automated classification with \texttt{GPT-5-mini}. In the first stage, the model receives the description of all areas and is prompted to assign the problem to the most relevant area. After problem area is detected, a similar process is used in the second stage to assign the problem to a competency within that area. We validate the reliability of this classification by having domain experts review a random sample of the mapped problems. To quantify coverage, we report the \emph{normalized entropy} of the problem distribution across competencies for each benchmark in Table~\ref{tab:rq2}. Our benchmarks achieve normalized entropy near 1.0 in all three domains, indicating near-uniform coverage across competencies. In contrast, existing benchmarks exhibit substantially lower entropy, reflecting concentration in a few areas with sparse or no coverage in other areas. Figure~\ref{fig:area_histogram_ml} illustrates this for Machine Learning: while our benchmark covers all areas with many tasks, MMLU-Pro and HLE leave several areas with few or no tasks. Similar plots for the Finance benchmarks are provided in Appendix~\ref{app:task_area_stats}.

Beyond coverage, we compare our benchmarks against prior benchmarks in terms of difficulty of questions and their discriminative power. Given a set of subject model, following \citet{li2024autobencher} we measure \textit{difficulty} as the error rate of the best-performing model and \textit{separability} as the mean absolute deviation of model accuracies, which captures how effectively a benchmark distinguishes models of different capability levels. Results are reported in Table~\ref{tab:rq2}. In all domains our benchmarks provide a good tradeoff between separability and difficulty. Sepcifically, in ML and Corporate Finance our benchmarks achieve the highest separability and second best difficulty, while in Personal Finance the benchmark achieves highest difficulty and second best separability.
\begin{table}[t]
\centering
\small
\caption{Normalized entropy ($H/\log N$) of the problem distribution across competencies, alongside Difficulty and Separability per benchmark. Higher normalized entropy indicates broader coverage of the competency space within each area. Numbers in parentheses indicate the number of problems. Within each section, the \best{best} value per column is bolded and the \second{second-best} is underlined.}
\label{tab:rq2}
\setlength{\tabcolsep}{8pt}
\renewcommand{\arraystretch}{1.15}
\resizebox{\columnwidth}{!}{%
\begin{tabular}{@{}llccc@{}}
\toprule
\textbf{Domain} & \textbf{Benchmark} & \textbf{Normalized entropy}\,$\uparrow$ & \textbf{Difficulty}\,$\uparrow$ & \textbf{Separability}\,$\uparrow$ \\
\midrule
\multirow{3}{*}{Machine Learning}
 & HLE (21)              & 0.6754           & \best{0.762}    & 0.056            \\
 & MMLU-Pro (85)         & 0.8094           & 0.200           & \second{0.067}   \\
 & \textbf{Ours (970)}   & \textbf{0.9791}  & \second{0.283}  & \best{0.108}     \\
\cmidrule(l){2-5}
\multirow{4}{*}{Personal Finance}
 & xFinBench (441)       & 0.6883           & 0.244           & 0.046            \\
 & FinanceMath (94)      & 0.6737           & \second{0.440}  & \best{0.077}     \\
 & BizBench (357)        & 0.7418           & 0.159           & 0.033            \\
 & \textbf{Ours (199)}   & \textbf{0.9605}  & \best{0.501}    & \second{0.059}   \\
\cmidrule(l){2-5}
\multirow{4}{*}{Corporate Finance}
 & xFinBench (537)       & 0.8448           & 0.244           & 0.046            \\
 & FinanceMath (106)     & 0.6013           & \best{0.440}    & \second{0.077}   \\
 & BizBench (387)        & 0.6133           & 0.159           & 0.033            \\
 & \textbf{Ours (685)}   & \textbf{0.9969}  & \second{0.321}  & \best{0.091}     \\
\bottomrule
\end{tabular}}
\end{table}

\subsection{Fine-Grained Model Profiling (RQ3)}
\label{sec:profiling}

\begin{figure}[t]
  \centering
  \includegraphics[width=\linewidth]{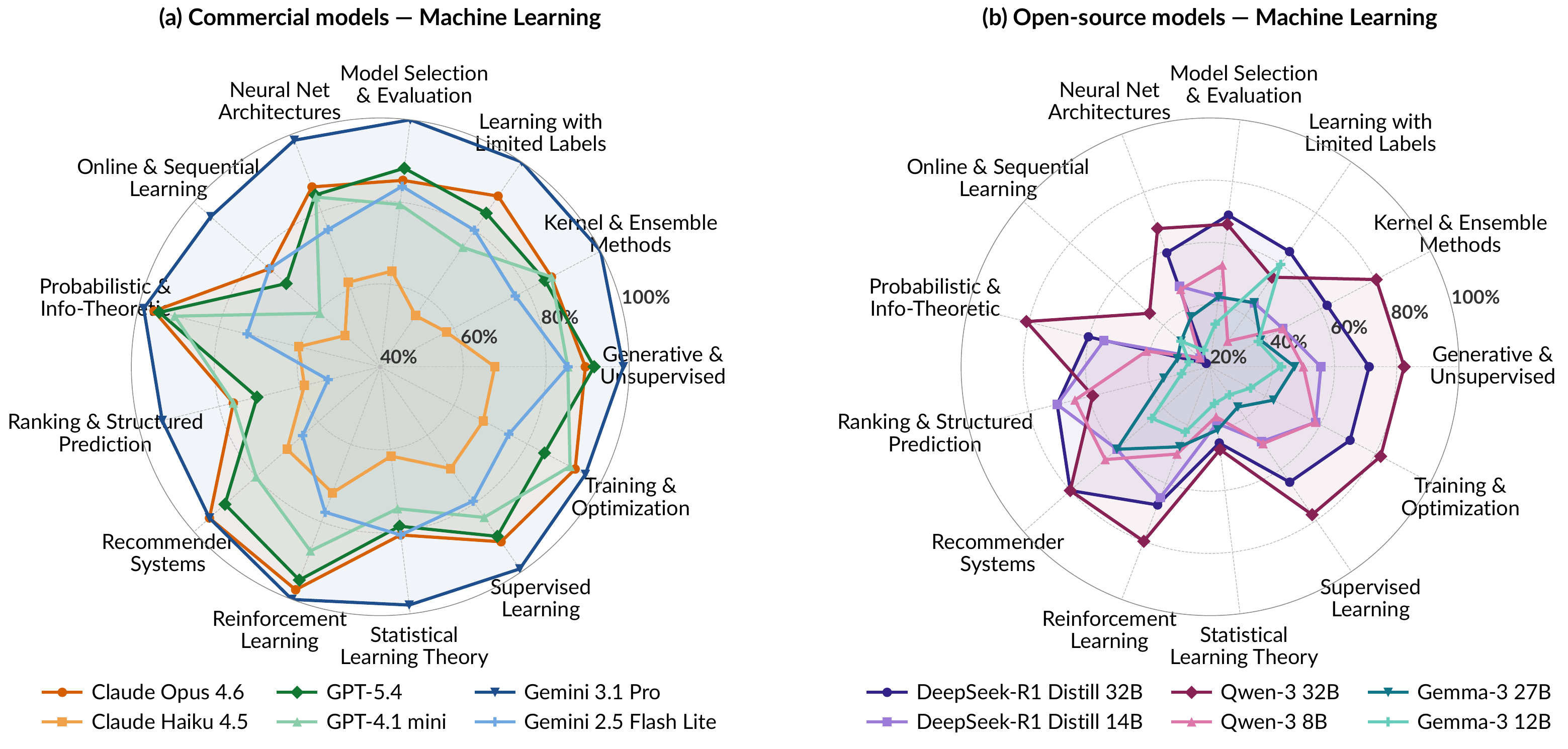}
  \caption{Area-wise performance of closed and open models on the Machine Learning benchmark.}
  \label{fig:spider_chart}
\end{figure}

To produce a fine-grained profile, for each model we compute accuracy for each area and competency in the taxonomy. Figure~\ref{fig:spider_chart} shows area-level model profiles in ML. These profiles reveal areas of strength and weakness for each model. For example, \geminipro and \claudeopus perform similarly on \emph{Recommender Systems} but diverge significantly on \emph{Online \& Sequential Learning}. Among open-source models, \texttt{DeepSeek-R1~Distill~32B} performs well across most areas yet ranks worst on \emph{Online \& Sequential Learning}.
Such findings help model developers target specific areas for improvement, and practitioners to select the model best suited for their application. Existing benchmarks cannot provide such details due to missing metadata or insufficient task coverage across areas of a domain. Similar results for the Finance benchmarks are provided in Appendix~\ref{app:area_level_performance}, and competency-level numbers for all three domains are reported in Appendices~\ref{app:competency_level_performance_ml} to~\ref{app:competency_level_performance_pf}.

\begin{figure}[t]
  \centering
  \includegraphics[width=\linewidth]{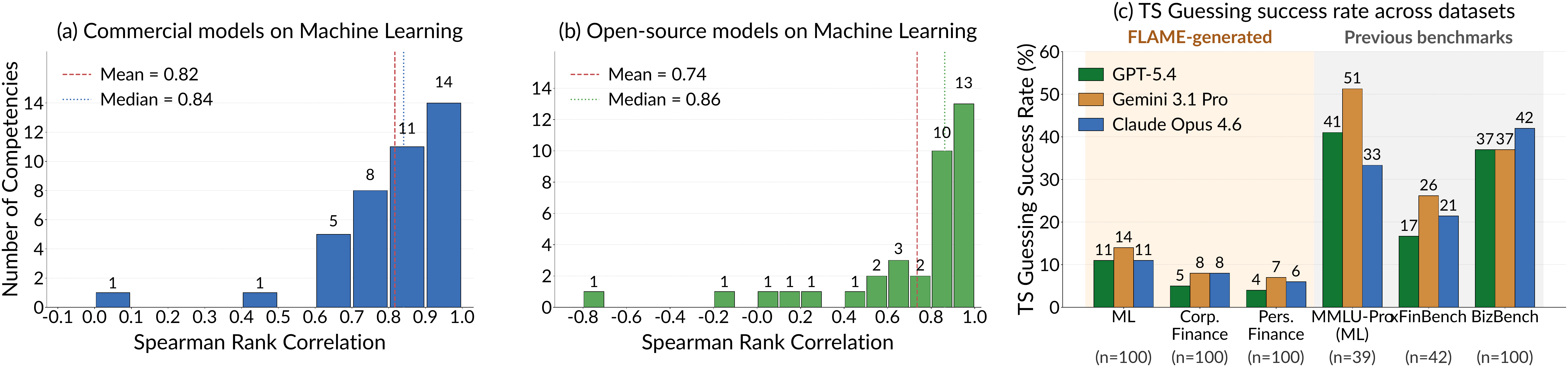}
  \caption{\textbf{(a, b)} Distribution of pairwise Spearman rank correlations across competencies for the ML domain. \textbf{(c)} TS-Guessing success rates across \framework-generated datasets (ML, Corporate Finance, Personal Finance) and existing benchmarks. Lower rates indicate less contamination. Subset sizes after filtering are shown in parentheses.}
  \label{fig:spearman_correlation_histogram}
\end{figure}

To assess whether competency-level evaluation adds information beyond aggregate scores, we compute the Spearman rank correlation between the vector of average model accuracies (over all competencies) and the vector of model accuracies on each individual competency. Figure~\ref{fig:spearman_correlation_histogram}(a,b) shows the histogram of rank correlations on the ML benchmark. Full results are available in the Appendix~\ref{app:additional_spearman_correlation_res}.
The distribution of correlations shows many values less than $1$, which indicate that on those competencies average performance scores mask meaningful differences in performance. This shows that competency-level performance scores provide information that is not captured by average accuracy. 
We also provide a Bloom’s-level breakdown of model performance in Appendix~\ref{app:blooms_level_performance}.

\subsection{Contamination Resistance (RQ4)}
\label{sec:contamination}

Public benchmarks are susceptible to data contamination, which undermines the validity of evaluation. Because \framework generates novel benchmark problems, even if the underlying source material has been included in a model's pretraining data, benchmark problems should be less susceptible to data contamination. To test this hypothesis, we apply the Testset Slot Guessing (TS-Guessing) protocol of \citet{deng2024investigating}. The premise is that a model which has memorized a benchmark item during training should be able to reconstruct components of that item that are not inferrable from the problem's semantic content alone. Because the exact wording of a distractor is not deducible from the question stem or the correct answer, successful reconstruction is strong evidence of memorization.

We apply TS-Guessing to our benchmarks and three existing benchmarks, MMLU-Pro ML subset, xFinBench, and BizBench, which contain a reasonable number of MCQ problems. For each dataset, we restrict the problems to those with short answer options ($\leq$5 words), numerical, or mathematical expressions, excluding problems with long free-text options to ensure that exact-match comparison is meaningful. For each problem, we randomly select one of the incorrect options as the masking candidate, present the model with the problem statement and all options preceding the candidate, and prompt it to guess the masked option. We evaluate three frontier models: \geminipro, \claudeopus, and \gptlarge.
Figure~\ref{fig:spearman_correlation_histogram}(c) reports the TS-Guessing success rate for each (dataset, model) pair. Across all three models, the \framework-generated datasets exhibit substantially lower guessing rates than the existing benchmarks. While our benchmark problems are completely novel, we conjecture that in some problems there's a pattern that helps LLMs guess the next option correctly. This could explain the small percentage of correct guessing in our benchmarks. On the prior benchmarks the success rates are much higher, which could indicate that these datasets having been exposed during pretraining. These results show that \framework's dynamic generation of novel problems provides robustness to data contamination, which makes evaluation more reliable compared to previous benchmarks.

\section{Conclusion}

We presented \framework, a framework for automated generation of fine-grained evaluation benchmarks grounded in external knowledge sources. \framework addresses three limitations of existing benchmarks: incomplete coverage of the competency space, lack of metadata for fine-grained analysis, and susceptibility to data contamination. The framework generates problems via a solution-graph-driven multi-agent pipeline, which significantly improves error rate of ground truth solutions. Using \framework, we generated benchmarks in three domains where previous benchmarks are scarce or lack enough coverage of domain knowledge: Machine Learning, Corporate Finance, and Personal Finance, producing 1,854 problems from 84 chapters across five textbooks at a cost of \$515. Our experiments demonstrate that these benchmarks achieve near-uniform coverage of the competency space, with normalized entropy over 0.96 in all three domains. 

\framework in its current form has a few limitations. For instance, to simplify evaluation, we generated problems in MCQ format, which may not capture all forms of reasoning. The pipeline depends on frontier LLMs for both generation and verification; hence, the quality of the generated problems is bounded by the capabilities of these models. While the solution-graph-driven strategy reduces ground-truth errors, it does not eliminate them entirely, and expert validation remains necessary. Additionally, benchmarks grounded in textbooks inherit the perspectives and potential biases of those sources, and strong benchmark performance should not be treated as a substitute for domain-specific validation in high-stakes deployment settings.



{
\small
\bibliographystyle{apalike}
\bibliography{main}
}

\newpage
\appendix

\section{Related Work}
\label{sec:related_work}

\subsection{Foundation Model Evaluation Frameworks}

As foundation models have grown in scale and capability, the need for principled evaluation frameworks has become increasingly apparent. HELM \citep{liang2023holistic} represents a
landmark effort toward holistic evaluation, introducing a two-dimensional taxonomy over
scenarios (use cases) and metrics (desiderata) and benchmarking 30 language models across
42 scenarios on 7 metrics including accuracy, calibration, robustness, fairness, and
efficiency. HELM's key insight is that evaluation should be multi-metric and transparent,
exposing trade-offs rather than collapsing performance into a single score. However, HELM
operates by aggregating existing benchmark datasets under a unified evaluation protocol; it
does not address the generation of new evaluation items or the systematic coverage of
capability spaces within individual domains.

Several recent works have highlighted fundamental methodological concerns with how
benchmarks are constructed and interpreted.
\citet{bean2025measuring} conduct a systematic review of 445 LLM benchmarks from
ICML, ICLR, NeurIPS, ACL, NAACL, and EMNLP, identifying recurring patterns---in the
choice of phenomena, tasks, and scoring metrics---that undermine the construct validity of
resulting claims. They provide eight recommendations and an operational checklist for
benchmark developers. In a complementary line of work, \citet{alaa2025position} argue that
LLM benchmarks should be empirically evaluated for construct validity, borrowing from the
psychometrics literature. Applying this framework to medical benchmarks, they demonstrate
significant gaps between benchmark performance and real-world clinical competency, even for
top-scoring models. These findings motivate our emphasis on grounding benchmark generation
in structured external knowledge sources with well-defined conceptual taxonomies, which
provides a principled basis for mapping evaluation items to the specific capabilities they
are designed to assess.

\subsection{Data Contamination and Dynamic Benchmarks}

A well-documented threat to the validity of static benchmarks is data contamination–the
leakage of benchmark items into model training corpora.
\citet{deng2024investigating} propose both retrieval-based and model-probing methods to
detect contamination, finding that on the MMLU benchmark, GPT-4 could guess missing answer
options with a 57\% exact match rate, suggesting substantial memorization of test data.
\citet{zhang2024careful} construct GSM1k, a held-out benchmark carefully matched in
difficulty to GSM8k, and observe accuracy drops of up to 13\% for certain model families,
with systematic overfitting across model sizes. \citet{golchin2023time} introduce a guided
instruction protocol for tracing contamination at the instance level. More recently,
\citet{zhao2025mmlu} build a contamination-free variant of MMLU by collecting novel
questions and applying rewriting strategies, confirming that evaluated LLMs exhibit clear
data leakage on the original MMLU. A comprehensive survey by \citet{cheng2025survey}
taxonomizes contamination along both phase-based (pretraining, fine-tuning, cross-modal)
and benchmark-based dimensions, cataloguing detection and mitigation methods.

These findings have motivated the development of dynamic benchmarks that resist
contamination through temporal freshness or procedural generation. LiveBench
\citep{white2024livebench} releases new questions monthly sourced from recent arXiv papers,
news articles, and datasets, scoring answers automatically against objective ground-truth
values without relying on LLM judges. DyVal \citep{zhu2023dyval} uses directed acyclic
graphs to dynamically generate evaluation samples with controllable complexity. OKBench
\citep{li2025okbench} proposes fully automated, on-demand benchmark generation sourced from Wikipedia to address staleness.

Our framework shares the goal of mitigating contamination but takes a fundamentally
different approach: rather than refreshing questions from ephemeral sources or
procedurally generating synthetic items, \framework grounds generation in authoritative,
well-structured domain references such as textbooks. This provides not only novelty of
items but also a principled taxonomic structure that enables systematic coverage of the
capability space---a property that purely dynamic benchmarks do not guarantee.

\subsection{Automated Benchmark and Question Generation}

Automated question generation has a long history in NLP and education.
\citet{scaria2024automated} examine the ability of five LLMs to generate educational questions
at different cognitive levels as defined by Bloom's taxonomy, finding that LLMs can produce
relevant, high-quality questions when prompted with adequate contextual information,
although quality varies substantially across models and cognitive levels.
\citet{hou2024compound} introduce CQ-Syn, a framework that leverages LLMs to generate and
refine compound questions according to structured guidelines followed by human review,
constructing a benchmark of 1,500 questions spanning language understanding, reasoning, and
knowledge.

In the domain of model evaluation specifically, the most related line of work uses LLMs to
generate evaluation items, sometimes grounded in external sources. CTIBench
\citep{alam2024ctibench} constructs multiple-choice questions for evaluating LLMs in cyber
threat intelligence (CTI) by drawing on authoritative standards and frameworks within the CTI
domain. \citet{zhang2025xfinbench} derive over 4,000 examples from graduate-level finance textbooks to evaluate five core financial capabilities, though the benchmark was curated with a
relatively narrow scope and limited taxonomic depth.

\framework differs from these prior efforts in several respects. First, our framework is
designed as a general-purpose benchmark generation pipeline rather than a one-off dataset
construction effort: it systematically derives evaluation items from the full conceptual
structure of reference sources, enabling broad and verifiable coverage. Second, each
generated item is annotated with rich, structured metadata that supports fine-grained
diagnostic evaluation at the level of individual concepts and skills. Third, the generation
process is automated and parameterized, allowing benchmarks to be reproduced, extended, and
regenerated as needed.

\subsection{Finance Benchmarks for Foundation Models}

The financial domain has seen growing interest in LLM evaluation, with several benchmarks
targeting different aspects of financial competency. FinQA \citep{chen2021finqa} and its
multi-turn extension ConvFinQA \citep{chen2022convfinqa} focus on numerical reasoning over
financial reports and tables. FinanceBench \citep{islam2023financebench} provides over
10,000 question-answer-evidence triplets for open-book financial QA grounded in public
company filings, revealing that even GPT-4-Turbo with retrieval incorrectly answered or
refused to answer 81\% of questions. FinanceQA \citep{mateega2025financeqa} targets professional financial analysis tasks mimicking on-the-job analyses at financial institutions, finding that current LLMs fail approximately 60\% of such tasks.

The most comprehensive prior effort is FinBen \citep{xie2024finben}, which organizes evaluation across eight financial task categories---including
information extraction, textual analysis, question answering, text generation, risk
management, forecasting, and decision-making---using 36 datasets and maintaining an open
leaderboard. While FinBen provides breadth across task types, it assembles existing datasets
rather than systematically generating items to cover the conceptual space of financial
knowledge.

Our work complements and extends these efforts along a different axis. Rather than
aggregating task-specific datasets, \framework generates a benchmark grounded in the
structured knowledge of finance textbooks, enabling evaluation along the
dimension of domain knowledge coverage and conceptual granularity. The resulting benchmark
provides a more complete map of where models succeed and fail across the space of financial
concepts and skills, going beyond aggregate task-level scores to fine-grained diagnostic
profiles.

\section{Source Preprocessing Details}
\label{app:preprocessing}

The source preprocessing pipeline consists of three steps: document
conversion, content filtering, and quality verification.

\paragraph{Document conversion.} We convert PDF-formatted textbooks to
machine-readable plain text using the Docling
toolkit~\citep{Docling} with \texttt{olmOCR-2-7B-0225}~\citep{olmocr2}
as the OCR backbone. Docling processes each page independently,
producing structured text output that preserves chapter and section
boundaries. For the Finance domain, this step was applied to two textbooks: \textit{Corporate Finance} \citep{welch2022corporate},
and \textit{Strategic Financial
Planning over the Lifecycle} \citep{charupat2012strategic}. And for Machine Learning domain the step was applied to three books:
Foundations of Machine Learning \citep{mohri2018foundations}, Probabilistic Machine Learning: An Introduction \citep{murphy2022probabilistic}, and Understanding Deep Learning \citep{prince2023understanding}.

\paragraph{Content filtering.} Once the full text of each book is
extracted, we segment the content by chapter and store each chapter as a
separate file for downstream consumption by the generation agents. We
automatically discard ancillary sections including prefaces, tables of
contents, lists of figures, glossaries, indices, and end-of-chapter
exercises and problem sets. The exclusion of exercise sections is
deliberate: retaining them would risk generating problems that closely
mirror existing textbook exercises, reducing both the novelty and the
diversity of the resulting benchmark. We also discard introductory chapters, as they tend to produce evaluation questions that are overly generic or comparatively easy and often fail to target a well-defined competency. This process resulted in 31 usable chapters from two Finance textbooks and 53 usable chapters from three Machine Learning textbooks.

\paragraph{Quality verification.} A human annotator reviewed the
extracted and filtered text for each chapter to verify that (i) the OCR
output is faithful to the source material, (ii) mathematical notation,
tables, and figures are rendered correctly or flagged for manual
correction, and (iii) the chapter segmentation aligns with the original
textbook structure. 

\section{Task Generation Details}
\label{app:task_generation}

This section provides the full implementation details of the task generation pipeline described in \S\ref{sec:task_generation}, including the specific models, frameworks, prompts, and hyperparameters used.

\subsection{Agent Architecture and Models}

The pipeline is implemented using the AutoGen multi-agent framework~\citep{microsoft2026autogen}, which provides the infrastructure for instantiating, coordinating, and managing message-passing between the designer and verifier agents.

The \textbf{designer agent} is instantiated with \geminipro~\citep{google2026gemini31propreview}. This model is responsible for all generation and repair steps: knowledge structuring, seed task generation, self-containment repair, conciseness editing, source-reference removal, soundness repair, and verification-driven repair.

The \textbf{verifier agent} is instantiated with \claudeopus~\citep{anthropic2026claude46}. This model is responsible for all assessment steps: trace-aware integrity check and repair, and final verification. Using a different model family for verification than for generation reduces the risk of shared failure modes between the two agents.

Both models are frontier LLMs chosen for their strong instruction-following, reasoning, and long-context capabilities, which are essential for processing full chapter texts alongside structured prompts. The framework is modular and the specific model choices can be substituted without modifying the pipeline architecture.

\subsection{Knowledge Structuring}
\label{app:knowledge_structuring}

Given the full text of a chapter, the designer agent is prompted to extract a structured summary organized into several categories. \emph{Core concepts and definitions} captures the fundamental terms introduced in the chapter (e.g., ``net present value,'' ``efficient frontier''). \emph{Rules and procedures} covers operational rules, algorithms, step-by-step procedures, and computational methods (e.g., the procedure for computing a bond's yield to maturity). \emph{Derived relationships} includes theorems, lemmas, corollaries, and formally derived results (e.g., the put--call parity relationship). \emph{Constraints and caveats} records boundary conditions, assumptions, exceptions, edge cases, and common misconceptions (e.g., the assumption of no-arbitrage in derivative pricing). Finally, the agent constructs a \emph{dependency graph}: a directed graph capturing prerequisite relationships among the extracted knowledge elements (e.g., understanding the risk-free rate is required before deriving the capital asset pricing model).

The resulting summary is stored as a structured JSON object and incorporated into all subsequent generation and verification prompts for that chapter. Detailed prompt has been provided in Appendix~\ref{app:kw_struct_prompt}.

\subsection{Valid Bloom's Taxonomy--Difficulty Pairings}
\label{app:bloom_pairings}

Table~\ref{tab:bloom_pairings} lists all Bloom's taxonomy levels, their definitions as provided to the designer agent, and the difficulty levels with which each is paired. We identified the difficulty-bloom level pairings with the assistance of GPT-5.2~\citep{openai2025gpt52}.

\begin{table}[h]
\centering
\small
\caption{Valid Bloom's taxonomy--difficulty pairings. Only cognitively coherent combinations are used for generation. The final benchmark is restricted to \emph{Hard} difficulty at higher-order Bloom's levels (marked with $\star$).}
\label{tab:bloom_pairings}
\begin{tabular}{lp{6.5cm}l}
\toprule
\textbf{Bloom's Level} & \textbf{Description (provided to agent)} & \textbf{Difficulty} \\
\midrule
Remember & Recall facts, terms, and basic concepts & Easy \\
Understand & Explain ideas or concepts in own words & Easy, Medium \\
Apply & Use knowledge or methods in new but familiar situations. Example verbs: calculate, demonstrate, use, implement. & Easy, Medium, Hard$^\star$ \\
Analyze & Break information into parts and examine relationships or patterns. Example verbs: differentiate, compare, examine, infer. & Medium, Hard$^\star$ \\
Evaluate & Make judgments based on criteria and standards. Example verbs: justify, critique, assess, argue. & Medium, Hard$^\star$ \\
Create & Combine elements to form a new pattern, structure, or product. Example verbs: design, compose, formulate, generate. & Medium, Hard$^\star$ \\
\bottomrule
\end{tabular}
\end{table}

\noindent In preliminary experiments, problems generated at Easy and Medium difficulty did not provide sufficient discriminative signal across models (see \S\ref{sec:task_generation}). The final benchmark therefore uses only the four Hard$^\star$ categories.


\subsection{Seed Task Generation}
\label{app:seed_generation}

The seed task generation prompt instructs the designer agent to produce a candidate MCQ in two sequential phases. In the first phase, the agent constructs a multi-step solution trace---a reasoning graph specifying the concepts drawn from the chapter, the intermediate computational or logical steps, the dependencies between steps, and any constraints or caveats that must be accounted for. The trace is represented as a structured list of steps, each with an explicit input--output specification.

In the second phase, the agent generates a self-contained multiple-choice question with five options (A--E) based on the solution trace. Options A--D are candidate answers (one correct, three distractors) and option E is always ``None of the above.'' The prompt explicitly requires that the question be solvable without access to the source chapter, that the correct answer be uniquely determined by the solution trace, that distractors correspond to plausible errors (e.g., omitting a step, applying a formula incorrectly, or misidentifying a constraint) rather than arbitrary values, and that the problem align with the target Bloom's level and difficulty.

To encourage diversity and appropriate challenge level, the prompt includes three exemplar problems drawn from the AI/Computer Science subset of the Humanity's Last Exam  benchmark~\citep{phan2025lastexam}. 
These exemplars are domain-external (not from Finance or ML) and serve only as stylistic and structural references for the desired level of complexity. Detailed prompt is provided in Appendix~\ref{app:seed_prompt}.

\subsection{Refinement Stage Prompts}
\label{app:refinement_prompts}

Below we describe the prompts used in each refinement stage.

\subsubsection{Self-Containment Repair}

The designer agent is instructed to inspect the candidate for undefined variables, notation, or abbreviations; implicit assumptions that are not stated in the problem; and missing quantitative data needed to compute the answer. The agent repairs any identified issues while explicitly avoiding unnecessary elaboration of standard domain knowledge that a competent practitioner would be expected to know. The prompt emphasizes that the subject model is expected to possess the knowledge presented in the chapter (algorithms, methods, formulas, theorems), and therefore such information should not be restated in the problem. Detailed prompt is provided in Appendix~\ref{app:self_contain_prompt}.

\subsubsection{Trace-Aware Integrity Check}

The verifier agent receives the 
candidate question, solution trace, and complete solution, and evaluates whether each step in the solution trace is logically and mathematically correct, whether the labeled correct answer is the unique output of the trace, whether each distractor is definitively incorrect, 
and whether the problem aligns with the target Bloom's taxonomy level. The verifier's role is not to solve the problem independently from scratch but rather to validate and, if necessary, repair the provided solution trace so that it is internally consistent, grounded only in the chapter material, and matches exactly one answer option.

When any condition is violated, the verifier generates a repaired version of the candidate, including corrected solution trace, answer options, and correct answer label, following a minimal-edit policy that preserves the problem's underlying concept and difficulty. The prompt is provided in Appendix~\ref{app:trace_integrity_prompt}.

\subsubsection{Conciseness Pass}

The designer agent is instructed to remove redundant or non-essential wording that does not affect the semantics of the question, such as restated definitions not required for solving the problem (e.g., ``a bond is a fixed-income security that pays periodic interest'' when this definition is not needed for the solution) or unnecessary qualifiers and verbose preambles. The agent must preserve the original meaning, cognitive demand, and all information necessary for solving the problem. No new information may be added, and technical content must not be rephrased. Detailed prompt is provided in Appendix~\ref{app:conciseness_prompt}.

\subsubsection{Source-Reference Removal}

The designer agent removes all explicit references to the source material. This includes phrases such as ``according to the chapter,'' ``as discussed in Section X,'' or ``the textbook states'', references to specific section, chapter, or page numbers, and any wording that reveals the textbook origin of the problem. The agent must preserve the original meaning, intent, and difficulty of the question while removing only source-identifying text. Detailed prompt is provided in Appendix~\ref{app:source_ref_remove_prompt}.

\subsubsection{Soundness Check}

The designer agent evaluates and repairs the candidate along four dimensions of soundness: logical coherence (no internal contradictions in the problem statement), grammatical clarity (the question reads naturally and unambiguously), completeness (all necessary information is present), and professional quality (the item meets the standard expected of a well-written assessment question). Unlike the trace-aware integrity check, which examines the consistency between the reasoning trace and the answer options, this stage is concerned with the surface-level quality of the item as written text. If the question is already sound, it is returned unchanged.
Detailed prompt is provided in Appendix~\ref{app:soundness_check_prompt}.

\subsection{Final Verification}
\label{app:final_verification}

The verifier agent performs an end-to-end assessment of the candidate task across four dimensions. \emph{Format validity} checks that the output is valid JSON conforming to the expected schema. \emph{MCQ integrity} verifies that exactly five options are present, that exactly one is correct, that all distractors are plausible yet unambiguously incorrect, and that option E is labeled ``None of the above.'' \emph{Bloom's alignment} checks whether the cognitive demand of the problem genuinely matches the target Bloom's level, using operational definitions for each level. \emph{Constraint compliance} verifies that the problem satisfies all constraints imposed during generation, including self-containment, absence of source references, proper LaTeX escaping, and avoidance of vague absolutes.

The verifier returns a structured JSON report containing a per-dimension assessment (Yes/No) and an overall accept/reject verdict. The overall verdict is ``Pass'' only if all four dimensions receive a ``Yes'' assessment. Detailed prompt is provided in Appendix~\ref{app:final_ver_prompt}.

\subsection{Verification-Driven Repair}
\label{app:repair}

When a candidate fails final verification, the pipeline routes the verifier's diagnostic to the designer agent for targeted repair. Two repair paths are available depending on the nature of the failure. If the failure is purely structural (invalid JSON while the MCQ content is otherwise sound), the designer agent is instructed to preserve the question content exactly and modify only the output format so that it becomes syntactically valid JSON---no changes to meaning, wording, or correctness are permitted. If the failure involves substantive MCQ issues (incorrect answer, ambiguous question, misaligned Bloom's level, or constraint violations), the designer agent receives the full context---previous candidate, verifier feedback, chapter text, knowledge summary, solution trace, and all previously accepted questions for anti-duplication---and may revise the question, options, and correct answer while preserving the underlying concept and Bloom's level.

The repaired candidate re-enters the pipeline at the self-containment repair stage and proceeds through all subsequent stages. This loop continues until the candidate passes final verification or a maximum of 3 retry attempts is reached, after which the candidate is discarded. The prompts for Format-only and Content-level repair have been provided in Appendices~\ref{app:format_repair_prompt}, and~\ref{app:content_repair_prompt}, respectively.

\subsection{Deduplication Details}
\label{app:dedup}

\paragraph{Prompt-level anti-duplication.}
Whenever one or more questions from the same chapter have already been accepted, the generation prompt is augmented with these accepted items and an explicit instruction to avoid near-duplicates. Duplication is defined broadly in the prompt: two questions are considered duplicates not only when they are lexically similar, but also when they test the same underlying concept or sub-skill, rely on the same primary method of solution, or follow the same reasoning pattern or solution structure even if the wording, numbers, scenario, or entities differ. The instruction further specifies that the agent must not generate questions that could be solved using the same mental steps as any prior question. The anti-duplication prompt is provided in Appendix~\ref{app:anti_duplication_prompt}.

\paragraph{Embedding-based chapter-level deduplication.}
After all candidates for a chapter have been generated and accepted, the pipeline applies a post-hoc deduplication filter. For each accepted question, a deduplication representation is constructed by concatenating the question stem with the text of its correct answer. These representations are embedded using OpenAI's \texttt{text-embedding-3-small} model, and pairwise cosine similarities are computed via a greedy sequential procedure: each candidate is compared against all previously retained items, and any candidate whose maximum cosine similarity to a retained item exceeds a threshold of 0.90 is discarded. The embedding vectors are stored to avoid redundant API calls across runs.

\begin{table*}[h]
\centering
\small
\resizebox{\textwidth}{!}{%
\begin{tabular}{lrrrrrrr}
\toprule
\textbf{Domain} 
& \textbf{Seed} 
& \textbf{Pass@1} 
& \textbf{Repaired} 
& \textbf{Discarded} 
& \textbf{Dedup. removed} 
& \textbf{Final size} 
& \textbf{Cost / Time} \\
\midrule
Personal Finance 
& 220 
& 208 
& 5 
& 0 
& 14 
& 199 
& \$48.67 / $\sim2$ days \\
Corporate Finance 
& 800 
& 772 
& 18 
& 1 
& 104 
& 685 
& \$181.96 / $\sim2$ days \\
ML/DL 
& 1,060 
& 910 
& 93 
& 7 
& 33 
& 970 
& \$284.13 / $\sim4$ days \\
\midrule
\textbf{Total}
& \textbf{2,080}
& \textbf{1,890}
& \textbf{116}
& \textbf{8}
& \textbf{151}
& \textbf{1,854}
& \textbf{\$514.76 / $\sim8$ days} \\
\bottomrule
\end{tabular}%
}
\caption{
Generation, verification, repair, deduplication, and API cost statistics for each benchmark domain. Here, \textit{Seed}: total number of generated candidates; \textit{Pass@1} and \textit{Repaired}: candidates accepted with and without repair, respectively; \textit{Discarded}: candidates rejected after retry exhaustion; and \textit{Dedup. removed}: candidates removed by embedding-based deduplication. \textit{Costs}: tracked designer and verifier generation calls only.}
\label{tab:generation_statistics}
\end{table*}
\subsection{Pipeline Statistics}
\label{app:pipeline_stats}

Table~\ref{tab:generation_statistics} reports the generation, verification, repair, deduplication, and cost statistics for each benchmark domain. 
For Personal Finance, the pipeline generated 220 seed candidates, of which 208 passed final verification on the first attempt and 5 required at least one repair iteration before acceptance. 
No candidates were discarded after exhausting the retry budget. 
Embedding-based deduplication removed 14 candidates, yielding a final benchmark of 199 accepted questions. 
The total tracked API cost for this domain was \$48.67, and the estimated active wall-clock time was $\sim2$ days.

For Corporate Finance, the pipeline generated 800 seed candidates. 
Among these, 772 passed final verification on the first attempt, while 18 required one or more repair iterations. 
Only 1 candidate was discarded after exhausting the retry budget. 
The embedding-based deduplication stage removed 104 candidates, resulting in a final benchmark size of 685 accepted questions. 
The total tracked API cost was \$181.96, with an estimated active wall-clock time of $\sim2$ days.

For Machine Learning, the pipeline generated 1,060 seed candidates. 
A total of 910 candidates passed final verification on the first attempt, and 93 candidates passed after at least one repair iteration. 
The pipeline discarded 7 candidates after exhausting the retry budget. 
After verification, embedding-based deduplication removed 33 candidates, producing a final benchmark of 970 accepted questions. 
The total tracked API cost for ML/DL was \$284.13, and the estimated wall-clock time was $\sim4$ days.

\section{Manual Inspection Details}
\label{app:expert_inspection}

Each expert evaluates sampled problems along several
dimensions: difficulty, relevance to the source chapter, clarity of the
problem statement and answer options, presence of redundant information,
and duplication.

Among the three ML errors identified in the random sample, all shared a
common pattern: a missing or incorrectly applied constant,
normalization, or scaling factor when invoking a theorem or bound.

Across all three benchmarks, problems spanned the difficulty spectrum
from easy to hard, with none flagged as trivial or unreasonably
difficult. This is notable given that the pipeline targets
hard-difficulty generation, suggesting that actual difficulty naturally
varies even under a hard-difficulty constraint. Given that the pipeline
generates a fixed number of problems per chapter, problem diversity
correlates with chapter length, i.e., shorter chapters resulted in less topical
variation, while longer chapters yielded a broader range of problems.
Problem statements and answer options were consistently rated as clear,
with distractors reflecting plausible reasoning errors and no redundant
information identified. Finally, inspection of the full problem set on a
random 10\% of chapters revealed no duplicate or near-duplicate questions
within any chapter.

\section{Additional Results}
\begin{figure}[h]
  \centering
  \includegraphics[width=\linewidth]{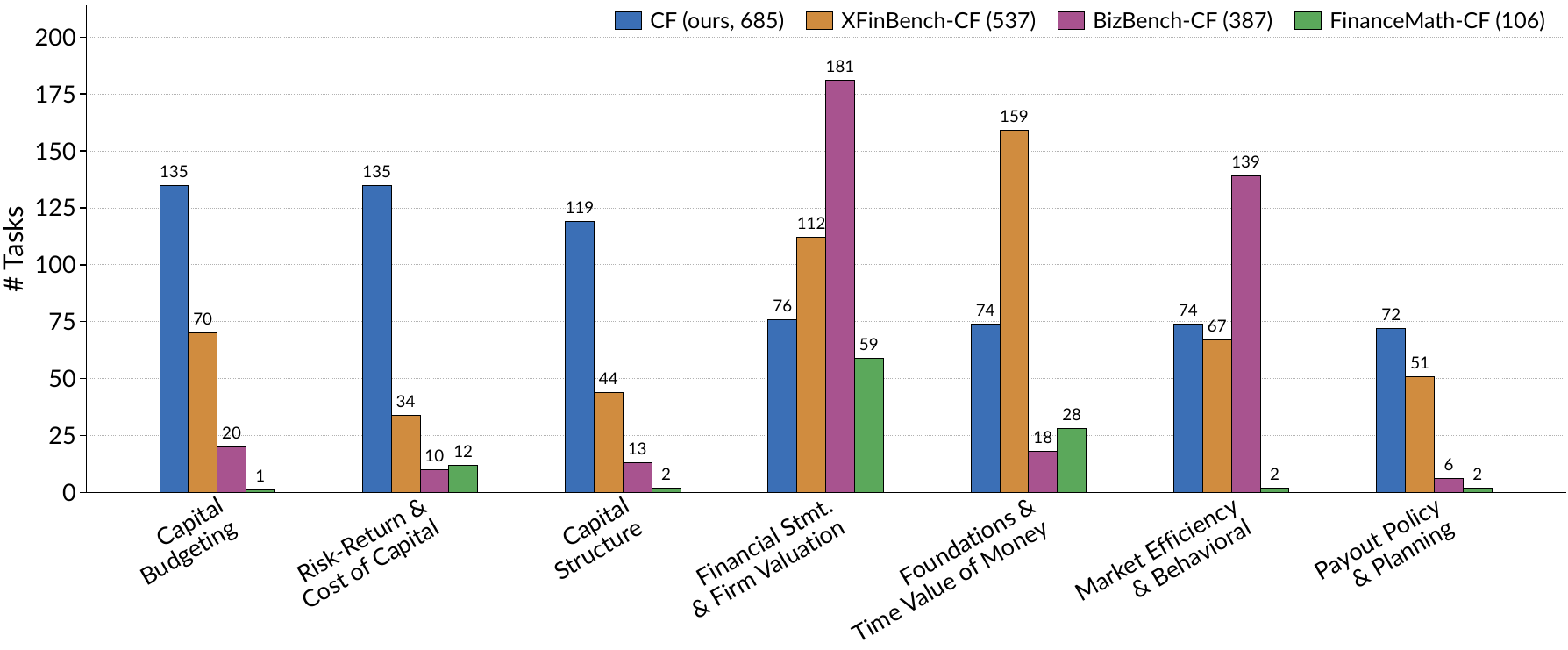}
  \caption{Task counts per area for corporate finance dataset (log scale on the $y$-axis)}
  \label{fig:cp_area_histogram}
\end{figure}

\subsection{Additional Area Coverage Statistics (RQ2)}
\label{app:task_area_stats}
\begin{figure}[h]
  \centering
  \includegraphics[width=\linewidth]{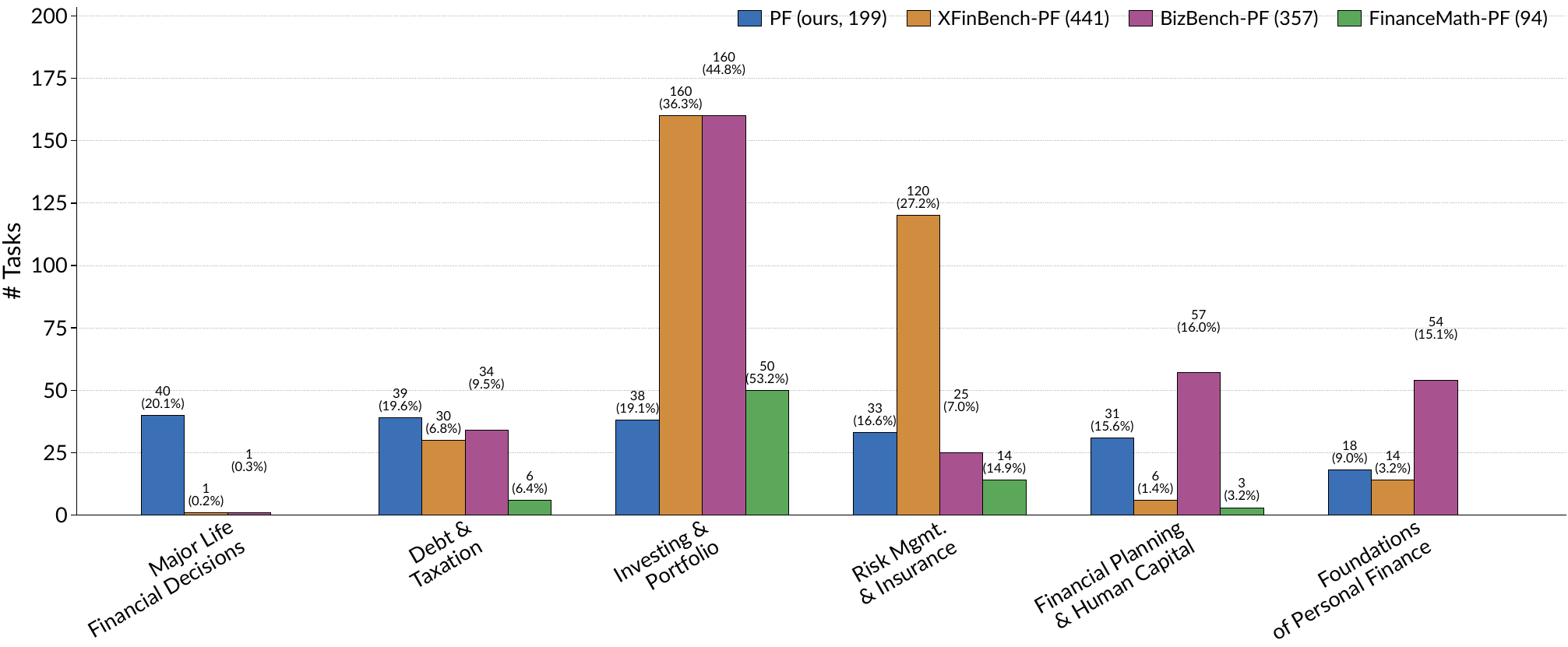}
  \caption{Task counts per area for personal finance dataset (log scale on the $y$-axis)}
  \label{fig:pf_area_histogram}
\end{figure}

Figure~\ref{fig:cp_area_histogram} and \ref{fig:pf_area_histogram} shows the distribution of task counts across domain areas for the Corporate Finance and Personal Finance benchmarks, respectively. 
In Corporate Finance, our benchmark is both larger and more balanced than prior datasets, with substantial coverage across all seven areas, whereas comparison benchmarks are more concentrated in a few areas such as \textit{Financial Statement \& Firm Valuation}, \textit{Foundations \& Time Value of Money}, or \textit{Market Efficiency \& Behavioral Finance}, etc. 

In Personal Finance, our benchmark similarly exhibits relatively uniform coverage across major areas, while prior benchmarks place disproportionate emphasis on a limited subset of topics, particularly \textit{Debt \& Taxation}, \textit{Investing \& Portfolio}, and \textit{Risk Management \& Insurance}, etc. Overall, the figure highlights the broader and more even area-level coverage of our benchmarks relative to existing alternatives.

\subsection{Fine-Grained Model Profiling (RQ3)}
\label{app:competency_level_profiling}
In this section, we provide additional details on area, competency, and bloom's level accuracy scores of closed and open-source subject models across Machine Learning, Corporate Finance, and Personal Finance domain.

\begin{figure}[t]
  \centering
  \includegraphics[width=\linewidth]{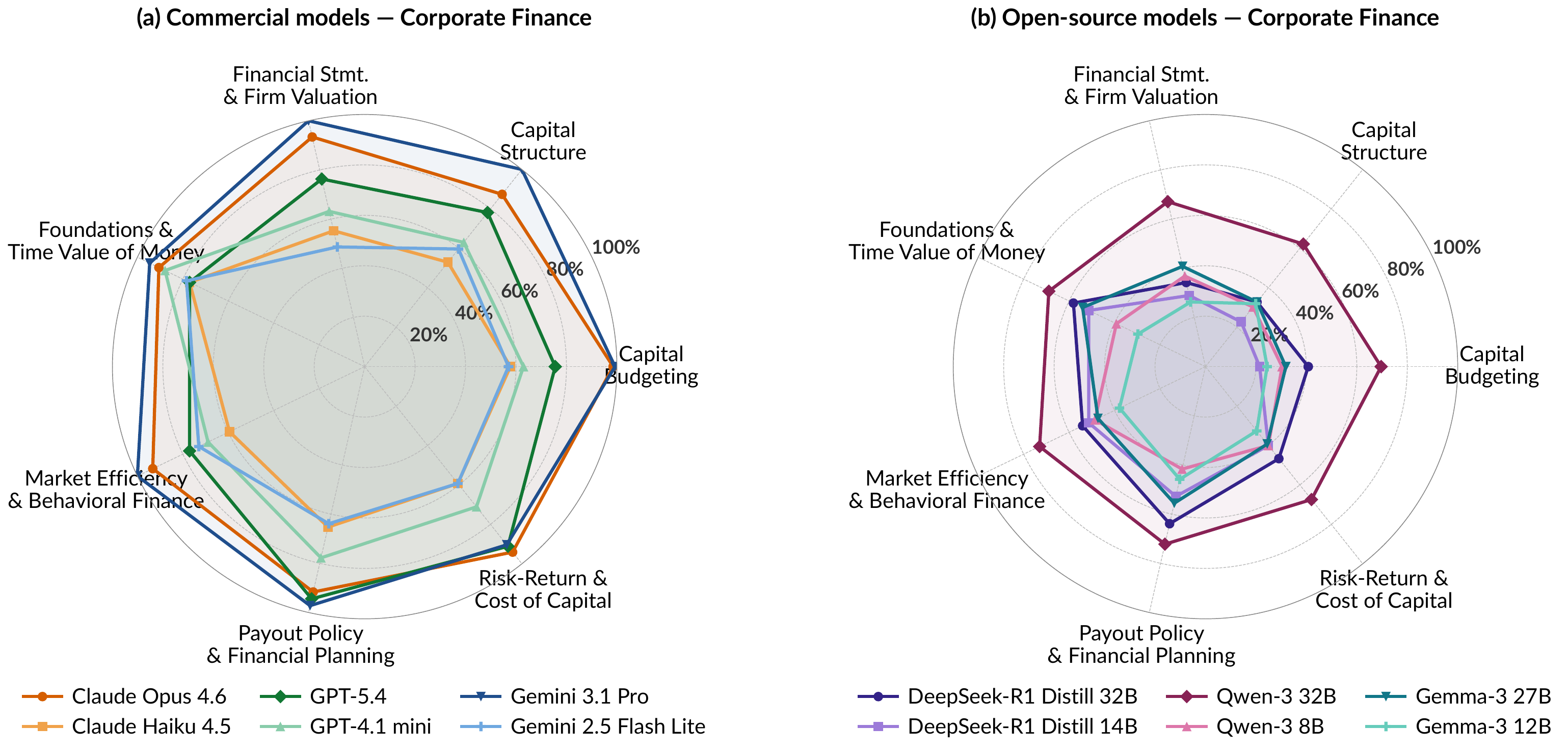}
  \caption{Area-wise performance of closed and open models on the Corporate Finance benchmark.}
  \label{fig:corporate_finance_spider}
\end{figure}

\begin{figure}[h]
  \centering
  \includegraphics[width=\linewidth]{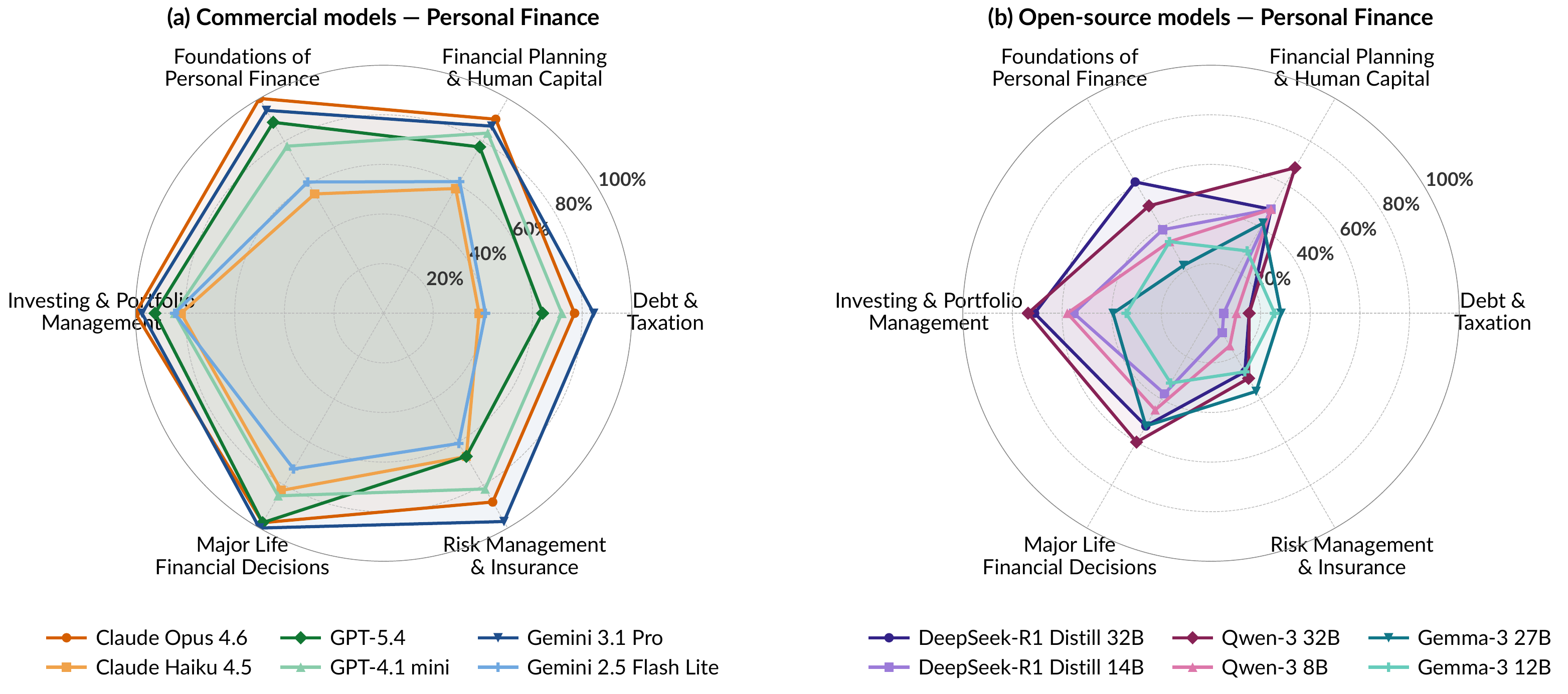}
  \caption{Area-wise performance of closed and open models on the Personal Finance benchmark.}
  \label{fig:personal_finance_spider}
\end{figure}
\subsubsection{Area-level Performance in Finance Benchmarks:}
\label{app:area_level_performance}
Figure~\ref{fig:corporate_finance_spider} and \ref{fig:personal_finance_spider} presents area-wise performance for commercial and open-source models in the Corporate Finance and Personal Finance domains. In Corporate Finance, commercial models exhibit large and relatively uniform accuracy, indicating both strong overall performance and broad coverage across areas. The strongest commercial model \geminipro remain close to the outer boundary on most axes, particularly in \textit{Capital Budgeting}, \textit{Capital Structure}, \textit{Financial Statement Analysis \& Firm Valuation}, and \textit{Risk-Return \& Cost of Capital}. In contrast, open-source models occupy a substantially smaller region and display more irregular profiles, suggesting both lower overall accuracy and greater variation across areas.

A similar pattern is observed in Personal Finance. Commercial models again form larger and more balanced accuracy, reflecting consistently strong performance across most areas, especially \textit{Foundations of Personal Finance}, \textit{Investing \& Portfolio Management}, and \textit{Major Life Financial Decisions}. By comparison, open-source models show markedly smaller and less uniform profiles, indicating weaker overall performance and more uneven area-level coverage. Their relative strengths appear in \textit{Investing \& Portfolio Management} and \textit{Financial Planning \& Human Capital}, while \textit{Debt \& Taxation} and \textit{Risk Management \& Insurance} remain comparatively more challenging. 

\begin{table}[h]
\centering
\small
\setlength{\tabcolsep}{4pt}
\resizebox{\textwidth}{!}{%
\begin{tabular}{lrrrrrr}
\toprule
\textbf{Competency} & \textbf{\geminiflashlite} & \textbf{\gptmini} & \textbf{\claudehaiku} & \textbf{\geminipro} & \textbf{\gptlarge} & \textbf{\claudeopus} \\
\midrule
Clustering & 1.00 & 1.00 & 0.47 & 0.94 & 0.88 & 0.71 \\
Diffusion Models & 0.87 & 0.87 & 0.60 & 1.00 & 0.87 & 0.93 \\
Dimensionality Reduction \& Representation Learning & 0.76 & 0.73 & 0.64 & 1.00 & 0.94 & 0.82 \\
Generative Adversarial Networks & 0.90 & 0.90 & 0.75 & 1.00 & 0.95 & 1.00 \\
Normalizing Flows & 0.88 & 0.94 & 0.71 & 1.00 & 0.94 & 1.00 \\
Unsupervised Learning: Taxonomy \& Evaluation & 0.75 & 0.75 & 0.80 & 1.00 & 0.85 & 0.95 \\
Variational Autoencoders & 0.90 & 0.90 & 0.75 & 0.95 & 0.95 & 0.90 \\
Decision Trees \& Ensemble Methods & 0.83 & 0.83 & 0.66 & 1.00 & 0.89 & 0.89 \\
Instance-Based \& Exemplar Methods & 0.60 & 0.90 & 0.45 & 1.00 & 0.95 & 0.85 \\
Kernel Methods \& Gaussian Processes & 0.79 & 0.92 & 0.51 & 1.00 & 0.77 & 0.82 \\
Support Vector Machines & 0.78 & 0.78 & 0.72 & 1.00 & 0.83 & 0.94 \\
Transfer, Semi-supervised, and Meta-Learning & 0.80 & 0.75 & 0.55 & 1.00 & 0.85 & 0.90 \\
Decision Theory \& Probabilistic Evaluation & 0.84 & 0.84 & 0.58 & 1.00 & 0.95 & 0.79 \\
Model Selection \& Cross-Validation & 0.83 & 0.90 & 0.76 & 1.00 & 0.93 & 0.90 \\
Regularization Techniques & 0.85 & 0.60 & 0.50 & 1.00 & 0.75 & 0.85 \\
Convolutional Neural Networks & 0.84 & 0.87 & 0.61 & 1.00 & 0.95 & 0.84 \\
Graph Neural Networks \& Graph Embeddings & 0.84 & 0.87 & 0.68 & 1.00 & 0.87 & 0.92 \\
Recurrent Neural Networks \& Sequence Modeling & 0.78 & 1.00 & 0.67 & 1.00 & 0.89 & 0.94 \\
Residual Networks \& Normalization & 0.75 & 0.90 & 0.80 & 1.00 & 0.85 & 1.00 \\
Shallow \& Deep Neural Networks & 0.68 & 0.72 & 0.58 & 0.96 & 0.68 & 0.75 \\
Transformers \& Attention Mechanisms & 0.60 & 0.85 & 0.40 & 0.95 & 1.00 & 0.90 \\
Learning Automata \& Formal Languages & 0.68 & 0.47 & 0.42 & 0.95 & 0.58 & 0.79 \\
Online Learning & 0.83 & 0.72 & 0.61 & 0.94 & 0.83 & 0.72 \\
Bayesian Learning \& Statistical Inference & 0.70 & 0.90 & 0.75 & 1.00 & 1.00 & 0.95 \\
Information Theory for Machine Learning & 0.60 & 0.95 & 0.55 & 1.00 & 0.95 & 1.00 \\
Maximum Entropy Models \& Density Estimation & 0.94 & 0.78 & 0.50 & 0.94 & 0.83 & 0.94 \\
Probabilistic Graphical Models \& Mixture Models & 0.70 & 1.00 & 0.60 & 1.00 & 1.00 & 0.95 \\
Ranking \& Learning to Rank & 0.53 & 0.76 & 0.59 & 0.94 & 0.71 & 0.76 \\
Recommender Systems & 0.65 & 0.80 & 0.70 & 0.95 & 0.90 & 0.95 \\
Reinforcement Learning & 0.78 & 0.88 & 0.72 & 1.00 & 0.95 & 0.97 \\
Algorithmic Stability \& Generalization & 0.64 & 0.91 & 0.45 & 0.91 & 0.73 & 0.73 \\
Complexity Measures: Rademacher Complexity \& VC-Dimension & 0.69 & 0.62 & 0.56 & 1.00 & 0.75 & 0.81 \\
PAC Learning Framework & 1.00 & 0.75 & 0.75 & 1.00 & 0.85 & 0.85 \\
Classification \& Logistic Regression & 0.71 & 0.73 & 0.76 & 1.00 & 0.90 & 0.86 \\
Generalized Linear Models & 0.83 & 0.83 & 0.78 & 1.00 & 0.94 & 1.00 \\
Generative Classification Models & 0.90 & 1.00 & 0.70 & 1.00 & 0.80 & 0.95 \\
Regression & 0.82 & 0.90 & 0.59 & 0.97 & 0.92 & 0.92 \\
Backpropagation \& Weight Initialization & 0.63 & 0.84 & 0.42 & 0.89 & 0.58 & 0.84 \\
Loss Functions & 0.94 & 1.00 & 0.78 & 1.00 & 1.00 & 1.00 \\
Optimization Algorithms & 0.71 & 0.91 & 0.77 & 0.97 & 0.91 & 0.94 \\
\bottomrule
\end{tabular}
}
\caption{Competency-level accuracy for Machine Learning: Closed Models.}
\label{tab:machine_learning_competency_accuracy_closed_models}
\end{table}

\subsubsection{Competency-level Performance in Machine Learning Benchmark:}
\label{app:competency_level_performance_ml}
Tables~\ref{tab:machine_learning_competency_accuracy_closed_models} and \ref{tab:machine_learning_competency_accuracy_open_models} demonstrate that the Machine Learning benchmark covers the broadest range of competencies among the three domains, while also revealing a clear and systematic performance gap between closed and open models. Among the closed models, \geminipro is the most consistently strong, achieving near-perfect accuracy across the majority of competencies, while \gptlarge and \claudeopus also perform robustly across most areas, particularly in topics such as \textit{Transformers \& Attention Mechanisms}, \textit{Bayesian Learning \& Statistical Inference}, \textit{Probabilistic Graphical Models \& Mixture Models}, \textit{Reinforcement Learning}, and \textit{Loss Functions}.

\begin{table}[h]
\centering
\small
\setlength{\tabcolsep}{3pt}
\resizebox{\textwidth}{!}{%
\begin{tabular}{lrrrrrr}
\toprule
\textbf{Competency} & \textbf{Qwen 8B} & \textbf{Gemma 12B} & \textbf{DeepSeek 14B} & \textbf{Gemma 27B} & \textbf{DeepSeek 32B} & \textbf{Qwen 32B} \\
\midrule
Clustering & 0.47 & 0.47 & 0.53 & 0.53 & 0.88 & 0.71 \\
Diffusion Models & 0.53 & 0.60 & 0.67 & 0.47 & 0.67 & 0.80 \\
Dimensionality Reduction \& Representation Learning & 0.30 & 0.33 & 0.30 & 0.42 & 0.58 & 0.73 \\
Generative Adversarial Networks & 0.70 & 0.40 & 0.60 & 0.55 & 0.90 & 0.95 \\
Normalizing Flows & 0.47 & 0.47 & 0.65 & 0.59 & 0.65 & 0.76 \\
Unsupervised Learning: Taxonomy \& Evaluation & 0.60 & 0.60 & 0.65 & 0.50 & 0.55 & 0.90 \\
Variational Autoencoders & 0.55 & 0.25 & 0.70 & 0.30 & 0.85 & 0.95 \\
Decision Trees \& Ensemble Methods & 0.40 & 0.37 & 0.40 & 0.37 & 0.60 & 0.74 \\
Instance-Based \& Exemplar Methods & 0.55 & 0.25 & 0.65 & 0.30 & 0.65 & 0.75 \\
Kernel Methods \& Gaussian Processes & 0.46 & 0.38 & 0.44 & 0.46 & 0.62 & 0.87 \\
Support Vector Machines & 0.50 & 0.50 & 0.44 & 0.33 & 0.67 & 0.83 \\
Transfer, Semi-supervised, and Meta-Learning & 0.30 & 0.60 & 0.45 & 0.45 & 0.65 & 0.55 \\
Decision Theory \& Probabilistic Evaluation & 0.58 & 0.21 & 0.47 & 0.37 & 0.79 & 0.79 \\
Model Selection \& Cross-Validation & 0.66 & 0.34 & 0.55 & 0.55 & 0.83 & 0.79 \\
Regularization Techniques & 0.30 & 0.45 & 0.20 & 0.30 & 0.40 & 0.35 \\
Convolutional Neural Networks & 0.39 & 0.32 & 0.34 & 0.37 & 0.61 & 0.68 \\
Graph Neural Networks \& Graph Embeddings & 0.58 & 0.29 & 0.61 & 0.45 & 0.66 & 0.79 \\
Recurrent Neural Networks \& Sequence Modeling & 0.72 & 0.06 & 0.72 & 0.22 & 0.78 & 0.67 \\
Residual Networks \& Normalization & 0.30 & 0.25 & 0.40 & 0.50 & 0.60 & 0.80 \\
Shallow \& Deep Neural Networks & 0.42 & 0.26 & 0.46 & 0.33 & 0.51 & 0.61 \\
Transformers \& Attention Mechanisms & 0.45 & 0.25 & 0.40 & 0.35 & 0.50 & 0.50 \\
Learning Automata \& Formal Languages & 0.26 & 0.21 & 0.11 & 0.26 & 0.16 & 0.37 \\
Online Learning & 0.22 & 0.44 & 0.39 & 0.39 & 0.28 & 0.56 \\
Bayesian Learning \& Statistical Inference & 0.35 & 0.40 & 0.35 & 0.35 & 0.65 & 0.80 \\
Information Theory for Machine Learning & 0.50 & 0.20 & 0.65 & 0.40 & 0.55 & 0.85 \\
Maximum Entropy Models \& Density Estimation & 0.28 & 0.33 & 0.56 & 0.33 & 0.44 & 0.72 \\
Probabilistic Graphical Models \& Mixture Models & 0.50 & 0.15 & 0.65 & 0.15 & 0.75 & 0.85 \\
Ranking \& Learning to Rank & 0.65 & 0.29 & 0.71 & 0.35 & 0.71 & 0.59 \\
Recommender Systems & 0.65 & 0.45 & 0.60 & 0.60 & 0.80 & 0.80 \\
Reinforcement Learning & 0.50 & 0.42 & 0.65 & 0.47 & 0.68 & 0.80 \\
Algorithmic Stability \& Generalization & 0.18 & 0.45 & 0.18 & 0.73 & 0.27 & 0.09 \\
Complexity Measures: Rademacher Complexity \& VC-Dimension & 0.31 & 0.25 & 0.31 & 0.06 & 0.50 & 0.50 \\
PAC Learning Framework & 0.50 & 0.30 & 0.55 & 0.50 & 0.50 & 0.65 \\
Classification \& Logistic Regression & 0.47 & 0.33 & 0.51 & 0.33 & 0.63 & 0.71 \\
Generalized Linear Models & 0.83 & 0.39 & 0.72 & 0.39 & 0.78 & 0.89 \\
Generative Classification Models & 0.55 & 0.30 & 0.40 & 0.35 & 0.60 & 0.75 \\
Regression & 0.36 & 0.26 & 0.41 & 0.38 & 0.64 & 0.82 \\
Backpropagation \& Weight Initialization & 0.21 & 0.21 & 0.26 & 0.16 & 0.42 & 0.63 \\
Loss Functions & 0.83 & 0.50 & 0.83 & 0.61 & 1.00 & 1.00 \\
Optimization Algorithms & 0.66 & 0.34 & 0.63 & 0.49 & 0.71 & 0.83 \\
\bottomrule
\end{tabular}
}
\caption{Competency-level accuracy for Machine Learning: Open Models. Abbreviations: Qwen 8B = qwen-3-8b; Gemma 12B = gemma-3-12b-it; DeepSeek 14B = deepseek-r1-distill-qwen-14b; Gemma 27B = gemma-3-27b-it; DeepSeek 32B = deepseek-r1-distill-qwen-32b; Qwen 32B = qwen-3-32b.}
\label{tab:machine_learning_competency_accuracy_open_models}
\end{table}
Nevertheless, even the stronger closed models such as \gptlarge and \claudeopus exhibit moderate performance on several specialized and theoretically demanding competencies, including \textit{Learning Automata \& Formal Languages}, \textit{Ranking \& Learning to Rank}, \textit{Algorithmic Stability \& Generalization}, and \textit{Backpropagation \& Weight Initialization}. 

The open-model results are substantially lower overall and exhibit much greater variation across competencies. Among them, \qwenlarge and \deepseeklarge clearly stand out as the strongest performers, often achieving comparatively strong accuracy on more applied or well-represented topics such as \textit{Generative Adversarial Networks}, \textit{Variational Autoencoders}, \textit{Kernel Methods \& Gaussian Processes}, \textit{Recommender Systems}, and \textit{Loss Functions}. In contrast, the remaining open models show considerably less stable performance and struggle more noticeably on conceptually abstract areas, including \textit{Residual Networks \& Normalization}, \textit{Algorithmic Stability \& Generalization}, and \textit{Complexity Measures: Rademacher Complexity \& VC-Dimension}.

\subsubsection{Competency-level Performance in Corporate Finance Benchmark:}
\label{app:competency_level_performance_cp}
Tables~\ref{tab:corporate_finance_competency_accuracy_closed_models} and \ref{tab:corporate_finance_competency_accuracy_open_models} reveal a clear separation between closed and open models on Corporate Finance in overall accuracy metric. Among closed models, \geminipro and \claudeopus are consistently the strongest, often reaching near-perfect performance across a wide range of competencies, while \gptlarge also performs strongly but with somewhat greater variability on topics such as \textit{Capital Budgeting Applications \& Pitfalls} and \textit{Stock \& Bond Valuation via Annuities \& Perpetuities}.
\begin{table}[h]
\centering
\small
\setlength{\tabcolsep}{3pt}
\resizebox{\textwidth}{!}{%
\begin{tabular}{lrrrrrr}
\toprule
\textbf{Competency} & \textbf{\geminiflashlite} & \textbf{\gptmini} & \textbf{\claudehaiku} & \textbf{\geminipro} & \textbf{\gptlarge} & \textbf{\claudeopus} \\
\midrule
Capital Budgeting Applications and Pitfalls & 0.41 & 0.62 & 0.38 & 1.00 & 0.67 & 0.95 \\
Capital Budgeting Decision Rules & 0.74 & 0.65 & 0.77 & 0.97 & 0.77 & 1.00 \\
Time-Varying Discount Rates and the Yield Curve & 0.56 & 0.59 & 0.65 & 1.00 & 0.85 & 1.00 \\
Uncertainty, Default Risk and Expected Returns & 0.61 & 0.68 & 0.55 & 1.00 & 0.74 & 1.00 \\
Capital Structure in a Perfect Market (Modigliani-Miller) & 0.79 & 0.50 & 0.54 & 1.00 & 0.71 & 0.86 \\
Capital Structure with Market Imperfections & 0.55 & 0.64 & 0.32 & 1.00 & 0.77 & 0.86 \\
Corporate Securities and Claims & 0.71 & 0.71 & 0.65 & 1.00 & 0.77 & 0.84 \\
Taxes and Capital Structure & 0.39 & 0.66 & 0.55 & 1.00 & 0.84 & 0.92 \\
Financial Statements and Economic Cash Flows & 0.49 & 0.41 & 0.41 & 1.00 & 0.64 & 0.87 \\
Firm Valuation via Comparables and Financial Ratios & 0.49 & 0.86 & 0.70 & 1.00 & 0.89 & 1.00 \\
Present Value and Discounting & 0.92 & 0.86 & 0.89 & 1.00 & 0.92 & 0.94 \\
Stock and Bond Valuation via Annuities and Perpetuities & 0.66 & 0.89 & 0.66 & 0.89 & 0.63 & 0.87 \\
Market Efficiency and Behavioral Finance & 0.79 & 0.66 & 0.68 & 1.00 & 0.84 & 0.95 \\
Market Imperfections & 0.67 & 0.72 & 0.50 & 1.00 & 0.69 & 0.92 \\
Equity Payouts: Dividends and Share Repurchases & 0.66 & 0.79 & 0.63 & 0.95 & 0.89 & 0.89 \\
Pro Forma Financial Statement Forecasting & 0.62 & 0.76 & 0.68 & 1.00 & 1.00 & 0.94 \\
Benchmarked Costs of Capital & 0.31 & 0.45 & 0.52 & 0.79 & 0.86 & 0.90 \\
Capital Asset Pricing Model (CAPM) & 0.71 & 0.77 & 0.77 & 0.86 & 0.94 & 0.91 \\
Investment Motives and Asset Classes & 0.62 & 0.82 & 0.62 & 0.95 & 0.85 & 0.97 \\
Investor Choice: Risk-Return Tradeoffs & 0.68 & 0.74 & 0.42 & 1.00 & 1.00 & 0.97 \\
\bottomrule
\end{tabular}
}
\caption{Competency-level accuracy for Corporate Finance: Closed Models.}
\label{tab:corporate_finance_competency_accuracy_closed_models}
\end{table}
In contrast, the lighter closed models, especially \geminiflashlite, \gptmini, and \claudehaiku show noticeably larger drops on more specialized or calculation-heavy topics, including \textit{Benchmarked Costs of Capital}, \textit{Taxes \& Capital Structure}, and \textit{Financial Statements \& Economic Cash Flows}. 

The open-model results are substantially lower overall and much less stable across competencies. 
\begin{table}[h]
\centering
\small
\setlength{\tabcolsep}{4pt}
\resizebox{\textwidth}{!}{%
\begin{tabular}{lrrrrrr}
\toprule
\textbf{Competency} & \textbf{Qwen 8B} & \textbf{Gemma 12B} & \textbf{DeepSeek 14B} & \textbf{Gemma 27B} & \textbf{DeepSeek 32B} & \textbf{Qwen 32B} \\
\midrule
Capital Budgeting Applications and Pitfalls & 0.18 & 0.23 & 0.08 & 0.23 & 0.26 & 0.54 \\
Capital Budgeting Decision Rules & 0.45 & 0.32 & 0.45 & 0.39 & 0.55 & 0.74 \\
Time-Varying Discount Rates and the Yield Curve & 0.26 & 0.21 & 0.21 & 0.29 & 0.38 & 0.68 \\
Uncertainty, Default Risk and Expected Returns & 0.35 & 0.23 & 0.16 & 0.39 & 0.48 & 0.87 \\
Capital Structure in a Perfect Market (Modigliani-Miller) & 0.14 & 0.25 & 0.11 & 0.29 & 0.11 & 0.68 \\
Capital Structure with Market Imperfections & 0.09 & 0.27 & 0.18 & 0.18 & 0.14 & 0.45 \\
Corporate Securities and Claims & 0.39 & 0.42 & 0.35 & 0.35 & 0.52 & 0.68 \\
Taxes and Capital Structure & 0.47 & 0.32 & 0.24 & 0.42 & 0.45 & 0.63 \\
Financial Statements and Economic Cash Flows & 0.18 & 0.26 & 0.15 & 0.36 & 0.15 & 0.49 \\
Firm Valuation via Comparables and Financial Ratios & 0.57 & 0.27 & 0.43 & 0.46 & 0.54 & 0.86 \\
Present Value and Discounting & 0.56 & 0.25 & 0.75 & 0.61 & 0.75 & 0.89 \\
Stock and Bond Valuation via Annuities and Perpetuities & 0.24 & 0.34 & 0.29 & 0.47 & 0.42 & 0.50 \\
Market Efficiency and Behavioral Finance & 0.55 & 0.50 & 0.61 & 0.58 & 0.55 & 0.74 \\
Market Imperfections & 0.42 & 0.25 & 0.42 & 0.36 & 0.53 & 0.72 \\
Equity Payouts: Dividends and Share Repurchases & 0.26 & 0.34 & 0.42 & 0.50 & 0.42 & 0.66 \\
Pro Forma Financial Statement Forecasting & 0.59 & 0.59 & 0.65 & 0.62 & 0.88 & 0.79 \\
Benchmarked Costs of Capital & 0.24 & 0.21 & 0.24 & 0.28 & 0.24 & 0.41 \\
Capital Asset Pricing Model (CAPM) & 0.34 & 0.23 & 0.37 & 0.46 & 0.40 & 0.60 \\
Investment Motives and Asset Classes & 0.60 & 0.47 & 0.55 & 0.62 & 0.68 & 0.80 \\
Investor Choice: Risk-Return Tradeoffs & 0.35 & 0.35 & 0.39 & 0.13 & 0.48 & 0.84 \\
\bottomrule
\end{tabular}
}
\caption{Competency-level accuracy for Corporate Finance: Open Models. Abbreviations: Qwen 8B = qwen-3-8b; Gemma 12B = gemma-3-12b-it; DeepSeek 14B = deepseek-r1-distill-qwen-14b; Gemma 27B = gemma-3-27b-it; DeepSeek 32B = deepseek-r1-distill-qwen-32b; Qwen 32B = qwen-3-32b.}
\label{tab:corporate_finance_competency_accuracy_open_models}
\end{table}
\qwenlarge clearly stands out as the strongest open model, achieving competitive performance on several topics such as uncertainty and expected returns, firm valuation, present value and discounting, and investor choice, whereas the remaining open models struggle to maintain consistent accuracy and often fall below $0.5$ on demanding areas such as \textit{Capital Structure}, \textit{Benchmarked Costs of Capital}, and \textit{Financial Statements \& Economic Cash Flows}. However, some topics appear broadly easier across model families, notably \textit{Present Value \& Discounting}, \textit{Investment Motives \& Asset Classes}, and \textit{Pro Forma Financial Statement Forecasting}.

\subsubsection{Competency-level Performance in Personal Finance Benchmark:}
\label{app:competency_level_performance_pf}
Tables~\ref{tab:personal_finance_competency_accuracy_closed_models} and \ref{tab:personal_finance_competency_accuracy_open_models} show a significant gap between closed and open models on Personal Finance. Similar to the performance in the Corporate Finance benchmark, the \geminipro and \claudeopus are the most consistently strong, frequently reaching near-perfect accuracy across core topics such as \textit{Portfolio Diversification \& Construction}, \textit{Housing Decisions: Buying vs. Renting}, and \textit{Retirement Planning \& Pension Income}, while \gptlarge and \gptmini also perform competitively but exhibit sharper drops on selected areas, such as \textit{Life Insurance \& Mortality Risk} or \textit{Debt Management \& Mortgage Financing}.
\begin{table}[h]
\centering
\small
\setlength{\tabcolsep}{4pt}
\resizebox{\textwidth}{!}{%
\begin{tabular}{lrrrrrr}
\toprule
\textbf{Competency} & \textbf{\geminiflashlite} & \textbf{\gptmini} & \textbf{\claudehaiku} & \textbf{\geminipro} & \textbf{\gptlarge} & \textbf{\claudeopus} \\
\midrule
Debt Management and Mortgage Financing  & 0.37 & 0.84 & 0.37 & 0.68 & 0.53 & 0.84 \\
Personal Income Tax Planning  & 0.45 & 0.60 & 0.40 & 1.00 & 0.75 & 0.70 \\
Consumption Smoothing and Optimal Savings & 0.46 & 0.77 & 0.62 & 1.00 & 0.69 & 0.85 \\
Personal Balance Sheet and Human Capital Valuation & 0.72 & 0.89 & 0.56 & 0.78 & 0.83 & 0.94 \\
Time Value of Money and Interest Rate Mathematics & 0.61 & 0.78 & 0.56 & 0.94 & 0.89 & 1.00 \\
Investment Fundamentals and Asset Classes & 0.79 & 0.79 & 0.89 & 0.95 & 0.89 & 1.00 \\
Portfolio Diversification and Construction & 0.89 & 0.89 & 0.74 & 1.00 & 0.95 & 1.00 \\
Housing Decisions: Buying vs. Renting & 0.80 & 0.90 & 0.80 & 1.00 & 0.95 & 1.00 \\
Retirement Planning and Pension Income  & 0.65 & 0.80 & 0.85 & 1.00 & 1.00 & 0.95 \\
Life Insurance and Mortality Risk & 0.61 & 0.83 & 0.67 & 0.94 & 0.50 & 0.83 \\
Risk Preferences and Insurance Principles & 0.60 & 0.80 & 0.67 & 1.00 & 0.87 & 0.93 \\
\bottomrule
\end{tabular}
}
\caption{Competency-level accuracy for Personal Finance: Closed Models.}
\label{tab:personal_finance_competency_accuracy_closed_models}
\end{table}

\begin{table}[h]
\centering
\small
\setlength{\tabcolsep}{4pt}
\resizebox{\textwidth}{!}{%
\begin{tabular}{lrrrrrr}
\toprule
\textbf{Competency} & \textbf{Qwen 8B} & \textbf{Gemma 12B} & \textbf{DeepSeek 14B} & \textbf{Gemma 27B} & \textbf{DeepSeek 32B} & \textbf{Qwen 32B} \\
\midrule
Debt Management and Mortgage Financing & 0.00 & 0.32 & 0.00 & 0.37 & 0.00 & 0.05 \\
Personal Income Tax Planning & 0.20 & 0.20 & 0.10 & 0.20 & 0.30 & 0.25 \\
Consumption Smoothing and Optimal Savings & 0.46 & 0.31 & 0.46 & 0.38 & 0.46 & 0.69 \\
Personal Balance Sheet and Human Capital Valuation & 0.50 & 0.28 & 0.50 & 0.44 & 0.50 & 0.67 \\
Time Value of Money and Interest Rate Mathematics & 0.33 & 0.33 & 0.39 & 0.22 & 0.61 & 0.50 \\
Investment Fundamentals and Asset Classes & 0.32 & 0.32 & 0.32 & 0.37 & 0.53 & 0.58 \\
Portfolio Diversification and Construction & 0.84 & 0.37 & 0.79 & 0.42 & 0.89 & 0.89 \\
Housing Decisions: Buying vs. Renting & 0.60 & 0.35 & 0.40 & 0.55 & 0.65 & 0.65 \\
Retirement Planning and Pension Income & 0.30 & 0.30 & 0.35 & 0.50 & 0.40 & 0.55 \\
Life Insurance and Mortality Risk & 0.00 & 0.22 & 0.00 & 0.28 & 0.22 & 0.06 \\
Risk Preferences and Insurance Principles & 0.33 & 0.33 & 0.20 & 0.47 & 0.33 & 0.60 \\
\bottomrule
\end{tabular}
}
\caption{Competency-level accuracy for Personal Finance: Open Models. Abbreviations: Qwen 8B = qwen-3-8b; Gemma 12B = gemma-3-12b-it; DeepSeek 14B = deepseek-r1-distill-qwen-14b; Gemma 27B = gemma-3-27b-it; DeepSeek 32B = deepseek-r1-distill-qwen-32b; Qwen 32B = qwen-3-32b.}
\label{tab:personal_finance_competency_accuracy_open_models}
\end{table}

In contrast, the open models are substantially weaker and exhibit much greater variability across competencies. Although \qwenlarge and \deepseeklarge achieve moderately strong performance on topics such as \textit{Portfolio Diversification \& Construction} and \textit{Housing Decisions: Buying vs. Renting}, most open models struggle markedly on several competencies, including \textit{Debt Management \& Mortgage Financing}, \textit{Personal Income Tax Planning}, and \textit{Retirement Planning \& Pension Income}. Their performance is particularly poor on \textit{Life Insurance \& Mortality Risk}. Overall, these results indicate that open models are not only less accurate than closed models, but also considerably less robust on jurisdiction-specific and insurance-related reasoning tasks.

\subsubsection{Model Performance Across Bloom’s Taxonomy Levels and Benchmarks:}
\label{app:blooms_level_performance}
Tables~\ref{tab:bloom_level_accuracy_closed_models} and \ref{tab:bloom_level_accuracy_open_models} show the performance across of different models across Bloom’s Taxonomy levels and benchmarks. Among the commercial models, \geminipro and \claudeopus are consistently the strongest across all three benchmarks, often achieving near-perfect accuracy across Bloom levels, whereas \gptlarge and \gptmini remain competitive but show greater variation across domains and categories. Machine Learning appears to be the most accessible domain for stronger models, with several models maintaining high accuracy across all four Bloom levels, whereas Personal Finance, and especially Corporate Finance, shows greater variation in performance across levels. Notably, accuracy does not consistently decrease for higher-order categories such as \textit{Evaluate} and \textit{Create}; in many cases, models perform comparably to, or even better than, their performance on \textit{Apply} and \textit{Analyze}. 

\begin{table}[h]
\centering
\small
\setlength{\tabcolsep}{4pt}
\resizebox{\textwidth}{!}{%
\begin{tabular}{lrrrrrrrrrrrr}
\toprule
\textbf{Model} & \multicolumn{4}{c}{\textbf{Machine Learning}} & \multicolumn{4}{c}{\textbf{Personal Finance}} & \multicolumn{4}{c}{\textbf{Corporate Finance}} \\
\cmidrule(lr){2-5}\cmidrule(lr){6-9}\cmidrule(lr){10-13}
 & \textbf{Apply} & \textbf{Analyze} & \textbf{Evaluate} & \textbf{Create} & \textbf{Apply} & \textbf{Analyze} & \textbf{Evaluate} & \textbf{Create} & \textbf{Apply} & \textbf{Analyze} & \textbf{Evaluate} & \textbf{Create} \\
\midrule
\geminiflashlite & 0.72 & 0.76 & 0.81 & 0.83 & 0.55 & 0.65 & 0.75 & 0.59 & 0.49 & 0.63 & 0.79 & 0.56 \\
\gptmini & 0.84 & 0.85 & 0.84 & 0.82 & 0.86 & 0.75 & 0.88 & 0.74 & 0.66 & 0.68 & 0.74 & 0.69 \\
\claudehaiku & 0.61 & 0.61 & 0.68 & 0.64 & 0.57 & 0.62 & 0.77 & 0.63 & 0.53 & 0.57 & 0.65 & 0.65 \\
\geminipro & 0.98 & 1.00 & 0.98 & 0.99 & 0.82 & 0.96 & 1.00 & 0.96 & 0.97 & 0.99 & 0.93 & 0.99 \\
\gptlarge & 0.84 & 0.86 & 0.87 & 0.91 & 0.73 & 0.85 & 0.81 & 0.85 & 0.84 & 0.78 & 0.82 & 0.82 \\
\claudeopus & 0.87 & 0.90 & 0.90 & 0.88 & 0.86 & 0.90 & 0.96 & 0.93 & 0.94 & 0.89 & 0.95 & 0.95 \\
\bottomrule
\end{tabular}
}
\caption{Bloom-level weighted accuracy by domain for Commercial Models.}
\label{tab:bloom_level_accuracy_closed_models}
\end{table}
A similar but weaker trend can be observed for open-source models. Notably, the \qwenlarge and \deepseeklarge clearly outperform the other open models and exhibit relatively stable behavior across Bloom levels, particularly in Machine Learning and Corporate Finance, whereas smaller models such as \gemmamed and \qwenmed remain substantially weaker overall. 
\begin{table}[h]
\centering
\small
\setlength{\tabcolsep}{4pt}
\resizebox{\textwidth}{!}{%
\begin{tabular}{lrrrrrrrrrrrr}
\toprule
\textbf{Model} & \multicolumn{4}{c}{\textbf{Machine Learning}} & \multicolumn{4}{c}{\textbf{Personal Finance}} & \multicolumn{4}{c}{\textbf{Corporate Finance}} \\
\cmidrule(lr){2-5}\cmidrule(lr){6-9}\cmidrule(lr){10-13}
 & \textbf{Apply} & \textbf{Analyze} & \textbf{Evaluate} & \textbf{Create} & \textbf{Apply} & \textbf{Analyze} & \textbf{Evaluate} & \textbf{Create} & \textbf{Apply} & \textbf{Analyze} & \textbf{Evaluate} & \textbf{Create} \\
\midrule
Qwen 8B & 0.44 & 0.47 & 0.52 & 0.47 & 0.24 & 0.42 & 0.40 & 0.33 & 0.32 & 0.39 & 0.43 & 0.35 \\
Gemma 12B & 0.26 & 0.33 & 0.40 & 0.37 & 0.33 & 0.33 & 0.35 & 0.20 & 0.23 & 0.32 & 0.46 & 0.26 \\
DeepSeek 14B & 0.46 & 0.48 & 0.51 & 0.55 & 0.22 & 0.31 & 0.40 & 0.33 & 0.35 & 0.30 & 0.43 & 0.36 \\
Gemma 27B & 0.31 & 0.43 & 0.47 & 0.38 & 0.39 & 0.37 & 0.42 & 0.35 & 0.32 & 0.41 & 0.58 & 0.34 \\
DeepSeek 32B & 0.60 & 0.63 & 0.62 & 0.66 & 0.35 & 0.48 & 0.52 & 0.43 & 0.42 & 0.47 & 0.50 & 0.44 \\
Qwen 32B & 0.71 & 0.76 & 0.71 & 0.75 & 0.35 & 0.50 & 0.58 & 0.54 & 0.68 & 0.68 & 0.71 & 0.65 \\
\bottomrule
\end{tabular}
}
\caption{Bloom-level weighted accuracy by domain for Open-Source Models. Abbreviations: Qwen 8B = qwen-3-8b; Gemma 12B = gemma-3-12b-it; DeepSeek 14B = deepseek-r1-distill-qwen-14b; Gemma 27B = gemma-3-27b-it; DeepSeek 32B = deepseek-r1-distill-qwen-32b; Qwen 32B = qwen-3-32b.}
\label{tab:bloom_level_accuracy_open_models}
\end{table}

\subsubsection{Additional Spearman Rank Correlation Results across Finance Benchmark:}
\label{app:additional_spearman_correlation_res}
\begin{figure}[h]
  \centering
  \includegraphics[width=\linewidth]{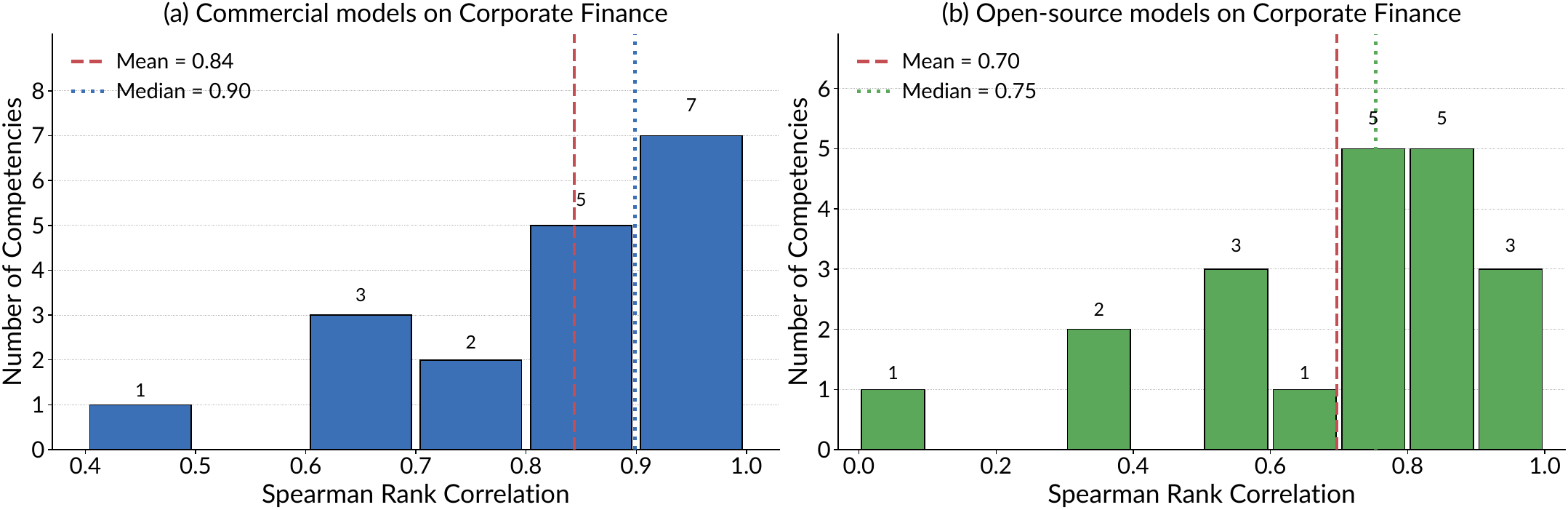}
  \caption{Distribution of pairwise Spearman rank correlations across competencies for Corporate Finance.}
  \label{fig:corporate_finance_spearman_hist}
\end{figure}

\begin{figure}[h]
  \centering
  \includegraphics[width=\linewidth]{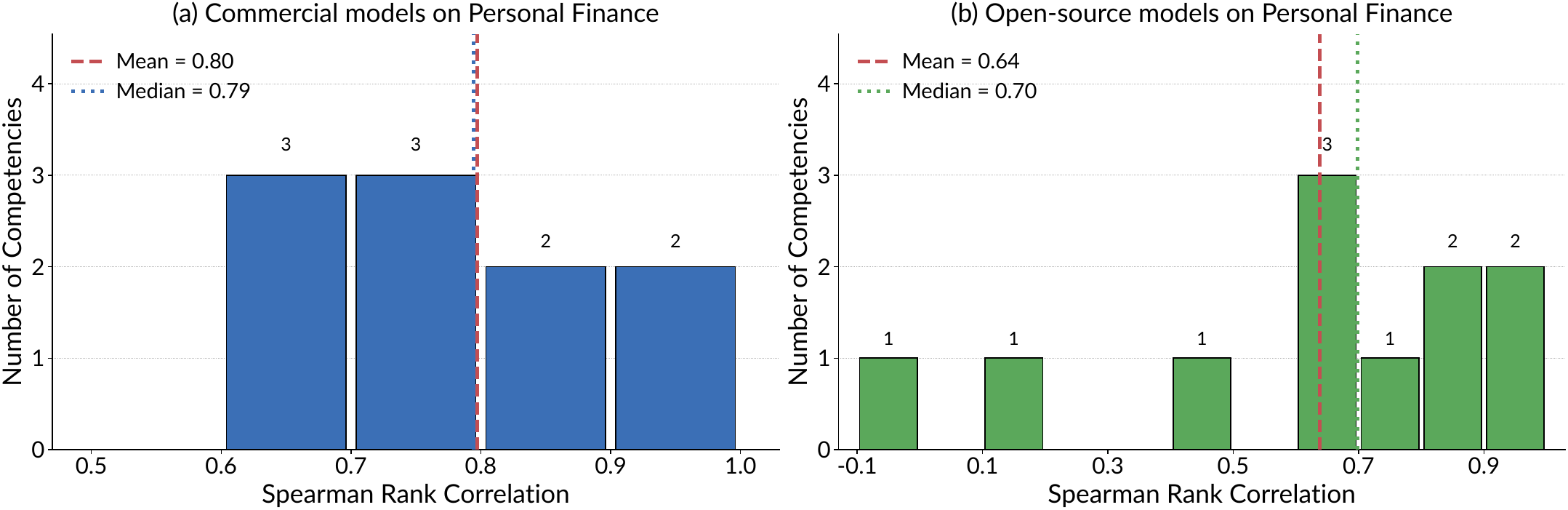}
  \caption{Distribution of pairwise Spearman rank correlations across competencies for Personal Finance.}
  \label{fig:personal_finance_spearman_hist}
\end{figure}

To assess whether competency-level evaluation provides information beyond aggregate scores, we further compute the Spearman rank correlation between the vector of average model accuracies and the vector of model accuracies for each individual competency for the Personal Finance and Corporate Finance benchmarks. Figures~\ref{fig:corporate_finance_spearman_hist} and \ref{fig:personal_finance_spearman_hist} show the histogram of pairwise model correlations for the Corporate Finance, and Personal Finance benchmarks, respectively. 

For Corporate Finance, the correlation distribution remains clearly below 1 for many competencies, with mean/median correlations of 0.84/0.90 for commercial models and 0.70/0.75 for open-source models. For Personal Finance, the pattern is similar, with mean/median values of 0.80/0.79 for commercial models and 0.64/0.70 for open-source models. The fact that many correlations fall short of 1 indicates that aggregate accuracy can obscure meaningful differences in how models perform across competencies. Overall, the competency-level performance provides information that is not captured by average accuracy alone, and the effect is especially pronounced for open-source models, whose cross-competency rankings are less consistent.

\section{Benchmark Generation and Inference Cost}
\label{app:all_gen_cost}
\subsection{Benchmark Generation}
\label{app:benchmark_gen_cost}
We report the overall benchmark-generation cost, together with breakdowns by model and provider; chapter-level averages are also reported. The benchmark generation over 84 chapters produced 1,854 final tasks, while discarding 151 candidate tasks, and required an estimated 10 days to complete. Overall, the pipeline consumed 98.74M tokens, comprising 80.59M input tokens and 18.16M output tokens, for a total estimated cost of \$514.76 USD. On average, this corresponds to 1.18M tokens and \$6.13 per chapter. Breaking the cost down by model, the designer model, \texttt{\geminipro}, consumed 79.36M tokens at a total cost of \$262.70, whereas the verifier model, \texttt{\claudeopus}, consumed fewer tokens overall, 19.39M, but accounted for a slightly smaller share of cost, \$252.06. The provider-level breakdown follows the same pattern, with Google accounting for \$262.70 and Anthropic accounting for \$252.06. Averaged over the 5 books, benchmark generation required 19.75M tokens and \$102.95 per book.

\begin{table}[h]
\centering
\small
\setlength{\tabcolsep}{7pt}
\begin{tabular}{lrrr}
\toprule
\textbf{Category} & \textbf{Input Tokens} & \textbf{Output Tokens} & \textbf{Cost (USD)} \\
\midrule
\geminipro (Designer) & 68,956,642 & 10,398,683 & 262.70 \\
\claudeopus (Verifier) & 11,630,566 & 7,756,435 & 252.06 \\
\midrule
Average per book & 16,117,442 & 3,631,024 & 102.95 \\
Average per chapter & 959,372 & 216,132 & 6.13 \\
\midrule
Overall & 80,587,208 & 18,155,118 & 514.76 \\
\bottomrule
\end{tabular}
\caption{Total compute cost of benchmark generation.}
\label{tab:benchmark_generation_cost}
\end{table}

\subsection{Commercial Models' Inference Cost}
\label{app:inference_cost}
We also estimate the inference cost of evaluating the generated benchmarks. Across the five books and 84 chapters, the currently completed evaluation logs comprise 504 model-evaluation runs. Overall, inference consumed 52.43M tokens, including 25.45M input tokens and 26.98M output tokens, for a total estimated cost of \$129.90 USD. This corresponds to an average of 0.62M tokens and \$1.55 per chapter, or 10.49M tokens and \$25.98 per book.

Although the subject models are evaluated on the same benchmark items, their input-token counts are not exactly identical. This is expected because different providers use different tokenizers and accounting conventions: the same prompt string can be segmented differently by OpenAI, Google, and Anthropic models, and provider APIs may include slightly different chat wrappers or message-formatting overhead. The judge model, \texttt{Gemini-3.1-Flash-Lite-Preview}, has substantially higher input-token usage because it receives not only the original question, but also the target answer, the subject model's generated response, and the grading prompt for each evaluated sample.

\begin{table}[h]
\centering
\small
\setlength{\tabcolsep}{7pt}
\begin{tabular}{lrrr}
\toprule
\textbf{Category} & \textbf{Input Tokens} & \textbf{Output Tokens} & \textbf{Cost (USD)} \\
\midrule
\claudehaiku & 996,200 & 2,004,826 & 11.02 \\
\claudeopus & 996,200 & 2,065,430 & 56.62 \\
\geminiflashlite & 885,902 & 12,349,782 & 5.03 \\
\geminipro & 885,902 & 1,901,561 & 24.59 \\
\gptmini & 861,326 & 3,885,776 & 6.56 \\
\gptlarge & 858,740 & 1,371,732 & 22.72 \\
\midrule
Subject models only & 5,484,270 & 23,579,107 & 126.54 \\
Gemini-3.1-Flash-Lite-Preview (Judge, estimated) & 19,969,458 & 3,400,745 & 3.36 \\
Average per book & 5,090,746 & 5,395,970 & 25.98 \\
Average per chapter & 303,021 & 321,189 & 1.55 \\
\midrule
Overall & 25,453,728 & 26,979,852 & 129.90 \\
\bottomrule
\end{tabular}
\caption{Estimated inference cost of evaluating the generated benchmarks. The average rows include both subject-model and judge-model usage.}
\label{tab:benchmark_inference_cost}
\end{table}

Breaking down the cost by model, the largest share comes from \texttt{\claudeopus}, which accounts for \$56.62, followed by \texttt{\geminipro} at \$24.59 and \texttt{\gptlarge} at \$22.72. The low-cost models are substantially cheaper: \texttt{\gptmini} costs \$6.56, \texttt{\geminiflashlite} costs \$5.03, and \texttt{\claudehaiku} costs \$11.02. The judge model, \texttt{Gemini-3.1-Flash-Lite-Preview}, contributes an estimated \$3.36. At the provider level, Anthropic accounts for \$67.64, OpenAI for \$29.28, and Google for \$32.98. Excluding the judge model, subject-model inference alone costs \$126.54.

\section{Verification Pass Rate and Retry Behavior by Bloom's Category}
\label{app:verification_stats}
We further analyze verification outcomes by Bloom's category using the verifier logs. Overall, pass rates were high across all categories, although clear differences emerged in how often candidates required revision before passing. \textit{Evaluate} exhibited the highest overall-verdict pass rate across verification attempts at 97.71\%, followed by \textit{Apply} at 96.01\%, \textit{Analyze} at 95.62\%, and \textit{Create} at 80.46\%. A similar pattern appears in retry behavior. 
\begin{table}[h]
\centering
\small
\setlength{\tabcolsep}{6pt}
\begin{tabular}{lrrrr}
\toprule
\textbf{Bloom Category} & \textbf{Candidates} & \textbf{Pass Rate (\%)} & \textbf{Final Pass Rate (\%)} & \textbf{Avg. Retries} \\
\midrule
Analyze  & 502 & 95.62 & 100.00 & 0.05 \\
Apply    & 505 & 96.01 & 100.00 & 0.04 \\
Create   & 494 & 80.46 & 98.38 & 0.22 \\
Evaluate & 513 & 97.71 & 100.00 & 0.02 \\
\bottomrule
\end{tabular}
\caption{Verification performance by Bloom's category. Pass Rate refers to the overall-verdict pass rate across verification logs, Final Pass Rate refers to the fraction of candidates that ultimately passed after exhausting the retry budget, and Avg. Retries is computed as average attempts required.}
\label{tab:bloom_pass_retry}
\end{table}
On average, \textit{Evaluate} required only 0.02 retries per candidate, while \textit{Apply} required 0.04 retries, indicating that these categories usually passed on the first verification attempt. \textit{Analyze} required slightly more revision at 0.05 average retries per candidate, whereas \textit{Create} was substantially more difficult, requiring 0.22 average retries and showing the only non-perfect final pass rate at 98.38\%. These results suggest that \textit{Create} tasks place greater pressure on the verification-and-repair loop, while \textit{Analyze}, \textit{Apply}, and \textit{Evaluate} tasks are comparatively easier to validate successfully.

\section{\framework Framework Pipeline Prompts}
\label{app:pipeline_prompts}
\subsection{Knowledge Structuring}
\label{app:kw_struct_prompt}
Below we provide the Prompt for Knowledge Structuring.
\begin{tcblisting}{
  enhanced,
  breakable,
  colback=gray!3,
  colframe=black!80,
  boxrule=0.5pt,
  arc=2pt,
  title={Prompt for Knowledge Structuring},
  fonttitle=\bfseries,
  listing only,
  listing options={
    basicstyle=\small\ttfamily,
    breaklines=true,
    columns=fullflexible,
    keepspaces=true
  }
}
SYSTEM_PROMPT = """
You are extracting detailed structured knowledge from a textbook chapter excerpt.

PHASE 1 - Extract and Structure Knowledge from the external source:
1) Identify:
- Core concepts
- Definitions
- Theorems or rules
- Procedures
- Algorithms
- Derived relationships
- Subtle constraints or caveats

2) Construct (internally) a dependency graph of how concepts rely on each other.

OUTPUT RULES:
- Output VALID JSON ONLY.
- Do NOT include any markdown fences.
- Do NOT include any commentary outside JSON.

Respond EXACTLY in the following format, including the JSON start and end markers:
{
  "core_concepts": [{"name": "...", "description": "..."}],
  "definitions": [{"term": "...", "definition": "..."}],
  "theorems_or_rules": [{"name": "...", "statement": "...", "conditions": ["..."], "implications": ["..."]}],
  "procedures": [{"name": "...", "steps": ["..."], "inputs": ["..."], "outputs": ["..."], "common_pitfalls": ["..."]}],
  "algorithms": [{"name": "...", "goal": "...", "steps": ["..."], "constraints": ["..."]}],
  "derived_relationships": [{"relationship": "...", "explanation": "..."}],
  "subtle_constraints_or_caveats": ["..."],
  "dependency_graph": {
    "nodes": [{"id": "C1", "label": "...", "type": "concept|definition|rule|procedure|algorithm"}],
    "edges": [{"from": "C1", "to": "C2", "relation": "requires|depends_on|uses"}]
  }
}
"""

USER_PROMPT = """
Input Details:
Chapter Material:
{chapter_excerpts}
"""
\end{tcblisting}

\subsection{Seed Task Generation}
\label{app:seed_prompt}
Below we provide the Prompt for Seed Task Generation.
\begin{tcblisting}{
  enhanced,
  breakable,
  colback=gray!3,
  colframe=black!80,
  boxrule=0.5pt,
  arc=2pt,
  title={Prompt for Seed Task Generation},
  fonttitle=\bfseries,
  listing only,
  listing options={
    basicstyle=\small\ttfamily,
    breaklines=true,
    columns=fullflexible,
    keepspaces=true
  }
}
SYSTEM_PROMPT = """
You are an expert problem designer and subject-matter specialist.
Your task is to design an extremely challenging multiple-choice problem grounded in the material provided in the external source below (e.g., a book chapter). The problem must be correct, unambiguous, and verifiably solvable using only knowledge from that source.
The problem must be difficult enough that even a strong model might fail to solve it reliably. The problem must be unique and different from the provided list of previously designed problems from the same source.
You will be provided with a difficulty level and a Bloom's level as design constraints. You must ensure that the problem you design meets the specified difficulty and targets the specified Bloom's level cognitive action.

Method You Must Follow (solution-graph-driven Design):
You must follow a solution-first design approach, consisting of the following phases:

Phase 1 - FIRST, in an incremental fashion construct a hard solution deep reasoning graph with some nodes fan-in/fan-out > 1. To make the problem challenging, if possible design graphs with width > 1. The reasoning graph, G = (V, E) which is a directed acyclic graph consists of nodes (V), where each node V[i] represents an intermediate solution, and an edge E[i, j] represents an operation by applying a core concept, definition, theorem, algorithm, etc. (the above knowledge list) from the external source content to node V[i] to obtain V[j].

Before writing the problem in text, explicitly construct a solution trace graph that:
1. Requires multiple non-trivial reasoning steps.
2. Combines two or more distinct concepts or results from different parts of the provided chapter.
3. Includes at least one of the following:
 - A non-obvious dependency
 - A hidden constraint
 - A delayed-use intermediate result
 - A reasoning-mode shift (e.g., conceptual -> algebraic -> conceptual)
4. Has at least one plausible but incorrect alternative reasoning path.
5. Cannot be solved by a single direct formula lookup.

The trace must be logically correct and lead to a unique final answer.
You must ensure:
 - Every step is justified using only the provided source.
 - The trace is internally consistent.
 - No external knowledge beyond the provided source is required.

Phase 2 - Design the Problem From the Trace
Now construct a self-contained multiple-choice problem such that:
1. Solving the problem correctly requires following the designed trace (or an equally complex equivalent).
2. The problem statement:
 - Does NOT reference any section numbers, subsections, example numbers, or explicit mentions of the chapter structure.
 - Does NOT say "according to the chapter" or similar phrases.
 - Is fully self-contained.
 - Defines all necessary notation.
 - Includes all required assumptions.

3. The problem cannot be solved by trivial pattern matching.
4. The reasoning chain is necessary for correctness.

Phase 3 - Construct High-Quality Answer Choices
The problem must have exactly 5 answer options:
A.
B.
C.
D.
E. None of the above.

Requirements for answer choices:
1. Exactly one option must be correct.
2. Distractors must be:
 - Derived from realistic but incorrect reasoning paths.
 - Based on common misunderstandings of the material.
 - Close in structure or value to the correct answer.
 - Not trivially eliminable.

3. "None of the above" must be a viable option (i.e., the other options should not trivially rule it out).
4. Avoid obviously absurd or dimensionally inconsistent distractors.
5. Do not include meta-commentary in the options.

Phase 4 - Internal Verification
Before outputting:
1. Independently verify the solution step-by-step.
2. Check that:
 - The problem is unambiguous.
 - The answer is uniquely correct.
 - No shortcut makes the problem trivial.
 - The reasoning genuinely requires multiple structured steps.
3. Ensure the problem depends only on the provided source.

Required Output:
Output must contain the following sections in order:
1. The constructed solution graph specifying nodes and edges
2. The Problem (Provide the complete, self-contained multiple-choice problem here.)
3. Answer Choices
    A.
    B.
    C.
    D.
    E. None of the above.
4. Correct Answer (Provide only the letter.)
5. Complete Solution
6. Option E MUST remain EXACTLY: "None of the above".
Provide a fully rigorous, step-by-step solution that follows the intended reasoning trace.
Do NOT reference any section numbers or structural elements of the source in the solution.

Difficulty Requirements
The problem must:
- Require at least 4-6 logically connected reasoning steps.
- Combine multiple concepts.
- Be resistant to shallow pattern matching.
- Be non-trivial even for a strong model.

Finally, you are provided the following examples ONLY to illustrate the desired STYLE and DEPTH of knowledge to solve the problems (creative, multi-concept, multi-step reasoning).
Do NOT copy the example content, numbers, entities, wording, or scenarios.
Do not include references to the chapter content (section number, theorem number, examples, etc.) in the problem. Each problem must be self-contained.

Example - 1:
<example from HLE CS/AI category>
Example - 2:
<example from HLE CS/AI category>
Example - 3:
<example from HLE CS/AI category>

IMPORTANT OUTPUT REQUIREMENT:
Respond EXACTLY in the following format, including the JSON start and end markers:
{
  "solution_graph": {
    "nodes": [{"id": "V1", "content": "..."}, {"id": "V2", "content": "..."}],
    "edges": [{"from": "V1", "to": "V2", "operation": "..."}]
  },
  "question": "<self-contained MCQ stem>",
  "options": { "A": "...", "B": "...", "C": "...", "D": "...", "E": "None of the above" },
  "correct_answer": "<one of: A|B|C|D|E>",
  "complete_solution": "<rigorous step-by-step solution text>"
}
"""

USER_PROMPT = """
Input Details:
{difficulty_and_blooms_guidance}

Textbook Chapter Excerpts:
{chapter_excerpts}

Textbook Chapter Knowledge Summary:
{chapter_knowledge_text}
"""
\end{tcblisting}

\begin{tcblisting}{
  enhanced,
  breakable,
  colback=gray!3,
  colframe=black!80,
  boxrule=0.5pt,
  arc=2pt,
  title={Prompt for Difficulty-Blooms level Guidance},
  fonttitle=\bfseries,
  listing only,
  listing options={
    basicstyle=\small\ttfamily,
    breaklines=true,
    columns=fullflexible,
    keepspaces=true
  }
}
Requested Cognitive Target:
- Difficulty: {difficulty}
- Bloom's Level: {blooms_level}

Requirements and Guidelines:
- Treat the requested difficulty as a design constraint on reasoning complexity, not as metadata.
- Preserve the requested Bloom's level throughout generation and revision.

Bloom's Level definitions:
1. Apply - Use knowledge or methods in new but familiar situations. Example verbs: calculate, demonstrate, use, implement.
2. Analyze - Break information into parts and examine relationships or patterns. Example verbs: differentiate, compare, examine, infer.
3. Evaluate - Make judgments based on criteria and standards. Example verbs: justify, critique, assess, argue.
4. Create - Combine elements to form a new pattern, structure, or product. Example verbs: design, compose, formulate, generate.
\end{tcblisting}

\subsection{Refinement Stage Prompts}
Below we provide the Refinement Stage prompts.

\subsubsection{Self-Containment Repair}
\label{app:self_contain_prompt}
Below we provide the Prompt for Self-Containment Repair.
\begin{tcblisting}{
  enhanced,
  breakable,
  colback=gray!3,
  colframe=black!80,
  boxrule=0.5pt,
  arc=2pt,
  title={Prompt for Self-Containment Repair},
  fonttitle=\bfseries,
  listing only,
  listing options={
    basicstyle=\small\ttfamily,
    breaklines=true,
    columns=fullflexible,
    keepspaces=true
  }
}
You will be given a question in the input.

Requirements:
- Inspect the question statement for any undefined notation, missing definitions, unstated assumptions, or missing information that could lead to ambiguity or multiple interpretations.
- If such issues are found, update the question by explicitly defining all notation and assumptions and by adding only the minimal necessary information to make the question self-contained and unambiguous.
- Do not introduce new assumptions beyond what is required for clarity.
- Do not change the difficulty, the core concept, or what the student/subject model is asked to do.
- Do not include definitions that a student with perfect knowledge and understanding of the subject matter is supposed to know.

IMPORTANT:
- The subject model is supposed to know the information provided in the chapter (algorithms, methods, formulas, theorems, etc.). Do not include such information in the problem statement.

Output:
Return valid JSON only, in the SAME STRUCTURE as the input.

Input Details:
Candidate_question:
{candidate_question}
\end{tcblisting}

\subsubsection{Trace-Aware Integrity Check}
\label{app:trace_integrity_prompt}
Below we provide the Prompt for Trace-Aware Integrity Check.
\begin{tcblisting}{
  enhanced,
  breakable,
  colback=gray!3,
  colframe=black!80,
  boxrule=0.5pt,
  arc=2pt,
  title={Prompt for Trace-Aware Integrity Check},
  fonttitle=\bfseries,
  listing only,
  listing options={
    basicstyle=\small\ttfamily,
    breaklines=true,
    columns=fullflexible,
    keepspaces=true
  }
}
SYSTEM_PROMPT = """
You will be given ONE multiple-choice question in JSON format.

You are also given:
- A solution trace (solution_graph)
- A complete solution text (step by step solution) that follows the trace.
- A Bloom's level that the question is supposed to target.

Your job is NOT to solve the problem independently from scratch.
Instead, you must validate and (if needed) REPAIR the PROVIDED solution trace so that it is:
- internally consistent,
- and matches exactly ONE answer option A-E.

You MUST enforce:
1) Trace validity:
   - Each step/node follows from prior steps/nodes and the described operation.
   - No missing jumps or unjustified claims.
   - The final answer implied by the trace is explicit.

2) Option consistency:
   - Exactly ONE option among A-E matches the trace's final answer.
   - correct_answer must point to that uniquely matching option.
   - All other options must NOT match the trace final answer.

Option E rule:
- Option E MUST remain EXACTLY: "None of the above".
- If any of A-D matches the trace final answer, then E must be incorrect.
- If none of A-D matches, then E must be the ONLY correct option.

3) Bloom's alignment:
   - The revised question must still match the requested Bloom's level.
   - Do not simplify the task into a lower-level cognitive action.
   - If the candidate drifts away from the requested Bloom's level, minimally revise it so the final MCQ matches the target.

-------------------------
WHAT TO DO IF ISSUES EXIST
-------------------------

A) If the solution trace is invalid / inconsistent / incomplete:
   - Update "question" ONLY if necessary to align with the repaired trace and to keep the problem self-contained.
     - Do NOT change the intended difficulty downward.
     - Keep the same "type" of problem; do not switch to a totally different concept.
   - Update "complete_solution" to match the repaired trace.
   - Update "options" and "correct_answer" so exactly one option matches the repaired final answer.

B) If the trace is valid BUT:
   - correct_answer does not point to the matching option, OR
   - multiple options match, OR
   - no option matches,
   then revise ONLY:
   - "options" (if needed), and/or
   - "correct_answer"
   Do NOT change: "question", "solution_graph", "solution_steps", "complete_solution".

-------------------------
REVISION POLICY
-------------------------
- Prefer minimal edits.
- Keep distractors plausible (realistic incorrect reasoning paths).
DO NOT OUTPUT YOUR PROCESS.
Make sure the output contains the following fields:
1. solution_graph (with nodes and edges)
2. question (the MCQ stem)
3. options (A, B, C, D, E)
4. correct_answer (one of A, B, C, D, E)
5. complete_solution (the rigorous step-by-step solution text)

Respond EXACTLY in the following format, including the JSON start and end markers:
{
  "solution_graph": {
    "nodes": [{"id": "V1", "content": "..."}, {"id": "V2", "content": "..."}],
    "edges": [{"from": "V1", "to": "V2", "operation": "..."}]
  },
  "question": "<self-contained MCQ stem>",
  "options": { "A": "...", "B": "...", "C": "...", "D": "...", "E": "None of the above" },
  "correct_answer": "<one of: A|B|C|D|E>",
  "complete_solution": "<rigorous step-by-step solution text>"
}

USER_PROMPT = """
Input Details:
Candidate Question JSON:
{candidate_question}

Solution Trace:
{solution_trace}

Full Solution JSON:
{solution_full}

Bloom's Level:
{blooms_level}
"""
\end{tcblisting}

\subsubsection{Conciseness Pass}
\label{app:conciseness_prompt}
Below we provide the Prompt for Conciseness Pass.
\begin{tcblisting}{
  enhanced,
  breakable,
  colback=gray!3,
  colframe=black!80,
  boxrule=0.5pt,
  arc=2pt,
  title={Prompt for Conciseness Pass},
  fonttitle=\bfseries,
  listing only,
  listing options={
    basicstyle=\small\ttfamily,
    breaklines=true,
    columns=fullflexible,
    keepspaces=true
  }
}
You will be given a question in the input.
Your task is to edit the question by removing any text that is irrelevant or redundant, while strictly preserving the original meaning, intent, and difficulty of the question.

Requirements:
- Remove redundant or non-essential wording that does not affect the semantics of the question.
- Do not add new information, rephrase technical content, or change the question's requirements.

Output:
Return valid JSON only, in the SAME STRUCTURE as the input.

Input Details:
Candidate_question:
{candidate_question}
\end{tcblisting}

\subsubsection{Source-Reference Removal}
\label{app:source_ref_remove_prompt}
Below we provide the Prompt for Source-Reference Removal.
\begin{tcblisting}{
  enhanced,
  breakable,
  colback=gray!3,
  colframe=black!80,
  boxrule=0.5pt,
  arc=2pt,
  title={Prompt for Source-Reference Removal},
  fonttitle=\bfseries,
  listing only,
  listing options={
    basicstyle=\small\ttfamily,
    breaklines=true,
    columns=fullflexible,
    keepspaces=true
  }
}
You will be given a question in the input.
Your task is to edit the question by removing any text that refers to the source material, while strictly preserving the original meaning, intent, and difficulty of the question.

Requirements:
- Remove phrases that explicitly reference the source material (e.g., "as discussed in the text," "as described in the text," "according to the chapter,", "as discussed in Section X, Y" etc.).
- Do not add new information, rephrase technical content, or change the question's requirements.

Output:
Return valid JSON only, in the SAME STRUCTURE as the input.

Input Details:
Candidate_question:
{candidate_question}
\end{tcblisting}

\subsubsection{Soundness Check}
\label{app:soundness_check_prompt}
Below we provide the Prompt for Soundness Check.
\begin{tcblisting}{
  enhanced,
  breakable,
  colback=gray!3,
  colframe=black!80,
  boxrule=0.5pt,
  arc=2pt,
  title={Prompt for Soundness Check},
  fonttitle=\bfseries,
  listing only,
  listing options={
    basicstyle=\small\ttfamily,
    breaklines=true,
    columns=fullflexible,
    keepspaces=true
  }
}
You will be given a question in the input.
Your task is to edit the question only as needed to make it sound and well-posed, while strictly preserving the original meaning, intent, and difficulty of the question.

Definition of soundness:
A question is sound if it (1) makes logical sense as stated, (2) is grammatically and semantically clear, (3) has no internal contradictions, (4) is phrased correctly, and 5) is complete.

Requirements:
- Edit any wording that makes the question unclear, incorrect, ill-formed, or logically inconsistent.
- Edit any grammatical errors that affect clarity.
- Edit the question if it is incomplete or missing essential information needed to understand what is being asked.
- Do NOT change what is being asked, do NOT change the difficulty, and do NOT introduce new constraints beyond what is necessary for soundness.
- If the question is already sound, return it UNCHANGED.

Output:
Return valid JSON only, in the SAME STRUCTURE as the input.

Input Details:
Candidate_question:
{candidate_question}
\end{tcblisting}

\subsection{Final Verification}
\label{app:final_ver_prompt}
Below we provide the Prompt for Final Verification.
\begin{tcblisting}{
  enhanced,
  breakable,
  colback=gray!3,
  colframe=black!80,
  boxrule=0.5pt,
  arc=2pt,
  title={Prompt for Final Verification},
  fonttitle=\bfseries,
  listing only,
  listing options={
    basicstyle=\small\ttfamily,
    breaklines=true,
    columns=fullflexible,
    keepspaces=true
  }
}
SYSTEM_PROMPT = """
You are an expert educational evaluator acting as an impartial **LLM-as-a-judge** for multiple-choice question generation quality.

You will be given:
- A Candidate Output (a response from question designer agent, in JSON format containing a MCQ with options and a labeled correct answer).

Your goal is to verify whether the candidate output strictly follows the required format and constraints, and whether each question is valid, self-contained, and well-designed.

You MUST check the following aspects:

1) Multiple-Choice Integrity.
For the question:
- Exactly **five** options (A, B, C, D, E) are present and non-empty strings.
- Option "E" is labeled as "None of the above".
- Distractors are plausible (reflect common misconceptions or near-misses) yet unambiguously incorrect if the concept is understood.

2) Constraint Compliance.
- Avoid vague absolutes ("always," "never," "most likely") unless explicitly required by the blueprint.
- If LaTeX appears, ensure escaped backslashes are used inside JSON strings (e.g., "$\\frac{1}{2}$").
- Must NOT explicitly refer to any section/theorem/lemma identifiers (e.g., "Section 2.1", "Theorem 2.1.1", "Lemma 2.1.1").
- Must NOT directly refer to the text like "as described in the text", "as discussed in the text", "according to the text"

3) Bloom's Alignment.
- Check whether the question genuinely matches the requested Bloom's level.
- Use the following operational definitions:
  - Apply: requires selecting and using taught methods in a concrete but non-trivial situation.
  - Analyze: requires breaking the situation into parts, tracing relationships, comparing cases, or inferring structure.
  - Evaluate: requires making and justifying a judgment using explicit criteria, trade-offs, or standards.
  - Create: requires synthesizing the best new structure, design, formulation, or plan under stated constraints.
- Mark this as "Yes" only if the target cognitive action is essential to solving the question correctly.

4) Output Format (Strict).
- STRICTLY ensure that the candidate output must be valid JSON and follow the expected structure:
  - Top-level is a single JSON object (NOT a list)
  - It has:
    - "question" (string)
    - "options" (object with keys A-E)
    - "correct_answer" (one of "A","B","C","D","E")
- Any missing key, wrong key (e.g., "questio"), wrong count, duplicate keys, or invalid JSON should result in format failure.
- VERY IMPORTANT: Ensure that the question can be easily parsed by standard JSON parsers.

Respond EXACTLY in the following format, including the JSON start and end markers:
{
  "json_format_valid": "<Yes/No>",
  "mcq_integrity": "<Yes/No>",
  "blooms_alignment": "<Yes/No>",
  "constraint_compliance": "<Yes/No>",
  "overall_verdict": "<Pass/Fail>",
  "explanation": "<2-4 sentences summarizing why it passes/fails>",
  "question_evaluation": {
        "distractors_plausible": "<Yes/No>",
        "main_issues": ["...","..."],
        "fix": "..."
    }
}
------------------------------------
DECISION RULE
------------------------------------
Set "overall_verdict" = "Pass" ONLY IF ALL of the following are "Yes":
- json_format_valid
- mcq_integrity
- blooms_alignment
- constraint_compliance
If "json_format_valid" = "No", overall_verdict MUST be "Fail" regardless of other fields.

VERY IMPORTANT: Now evaluate the Candidate Output using the inputs provided. Return ONLY the RESPONSE JSON.
"""

USER_PROMPT = """
Input Details:
Requested Bloom's Level:
{blooms_level}

Candidate Output:
{candidate_output}
"""
\end{tcblisting}

\subsection{Verification-Driven Repair}
\label{app:ver_repair_prompt}
Below we provide the Prompts for Verification-Driven Repair.
\subsubsection{Format-only Repair}
\label{app:format_repair_prompt}
\begin{tcblisting}{
  enhanced,
  breakable,
  colback=gray!3,
  colframe=black!80,
  boxrule=0.5pt,
  arc=2pt,
  title={Prompt for Format-only Repair},
  fonttitle=\bfseries,
  listing only,
  listing options={
    basicstyle=\small\ttfamily,
    breaklines=true,
    columns=fullflexible,
    keepspaces=true
  }
}
SYSTEM_PROMPT = """
You are a JSON repair tool.

INPUTS YOU WILL RECEIVE
1) Previous Candidate Output: an MCQ that may be malformed JSON (or valid JSON but wrong formatting).
2) Verifier LLM Feedback: indicates formatting/JSON issues.
3) The original MCQ content must be preserved exactly.

YOUR GOAL
Fix ONLY the JSON formatting so that the output is valid JSON and can be parsed by a standard JSON parser.

WHAT YOU MUST DO
- Produce a single valid JSON object that matches the intended schema.
- Preserve the question text, all option texts, and the correct answer EXACTLY as they appear in the input (character-for-character), except for changes strictly required to make it valid JSON (e.g., escaping quotes, backslashes, newlines).

WHAT YOU MUST NOT DO
- Do NOT change the meaning, wording, punctuation, capitalization, numbers, symbols, or spacing of any content fields (question/options/correct_answer).
- Do NOT add, remove, or rename keys beyond what is required to restore valid JSON.
- Do NOT "improve" writing, grammar, clarity, options, or correctness. This task is format repair ONLY.

REQUIRED OUTPUT SCHEMA
Output ONLY a single valid JSON object (no markdown, no commentary) with:
- "question": string
- "options": object with keys "A","B","C","D","E" (each a string)
- "correct_answer": one of "A","B","C","D","E"

If these keys exist in the input, preserve them. If the input contains additional keys, preserve them exactly as-is as well (only fixing JSON syntax/escaping).

DO NOT OUTPUT YOUR PROCESS.
Return ONLY the repaired JSON.

Respond EXACTLY in the following format, including the JSON start and end markers:
{
  "question": "<self-contained MCQ stem>",
  "options": { "A": "...", "B": "...", "C": "...", "D": "...", "E": "None of the above" },
  "correct_answer": "<one of: A|B|C|D|E>"
}
"""

USER_PROMPT = """
Input Details:

Previous Candidate Output:
{previous_candidate_output}

Verifier LLM Feedback:
{verifier_llm_feedback}
"""
\end{tcblisting}

\subsubsection{Content-level Repair}
\label{app:content_repair_prompt}

\begin{tcblisting}{
  enhanced,
  breakable,
  colback=gray!3,
  colframe=black!80,
  boxrule=0.5pt,
  arc=2pt,
  title={Prompt for Content-level Repair},
  fonttitle=\bfseries,
  listing only,
  listing options={
    basicstyle=\small\ttfamily,
    breaklines=true,
    columns=fullflexible,
    keepspaces=true
  }
}
SYSTEM_PROMPT = """
You are an expert educational scientist and MCQ quality auditor.

INPUTS YOU WILL RECEIVE
1) Previous Candidate Output: a single MCQ in JSON (may be imperfect).
2) Verifier LLM Feedback: issues to fix (MCQ correctness, integrity, ambiguity, constraints, etc.).
3) chapter_material: textbook chapter excerpt that constrains scope and facts.
4) chapter_knowledge_text: structured knowledge summary of the chapter.
5) solution_trace: the reasoning trace/solution graph associated with the question.
6) previous_questions: a list of previously generated questions for this chapter (anti-dup).

YOUR GOAL
Repair the MCQ so it is fully consistent with the provided solution_trace and grounded ONLY in chapter_material and chapter_knowledge_text.
The solution_trace is the strongest constraint: the revised question must be solvable via the trace, and the correct answer option must match the trace's final answer.
The revised question must also preserve the requested Bloom's level.

WHAT YOU ARE ALLOWED TO CHANGE
- You MAY edit: "question", "options", and/or "correct_answer" as needed to match the trace and satisfy the verifier feedback.
- You MAY edit wording to make the question self-contained and unambiguous.
- You MAY minimally adjust distractors to remain plausible but incorrect.

WHAT YOU MUST NOT CHANGE
- Do NOT introduce any facts, definitions, formulas, or claims not supported by chapter_material / chapter_knowledge_text.
- Do NOT change the underlying concept/skill tested to a different one.
- Do NOT produce a near-duplicate of any item in previous_questions (same concept + same solution method/reasoning pattern).
- Do NOT drift away from the requested Bloom's level.

STRICT OUTPUT RULES
- Output ONLY a single valid JSON object (no markdown, no commentary).
- Preserve the JSON key names exactly: "question", "options", "correct_answer".
- "options" must contain exactly five keys: "A","B","C","D","E".
- "E" MUST be exactly: "None of the above".
- "correct_answer" must be exactly one of: "A","B","C","D","E".
- If the previous JSON includes extra fields (e.g., solution_graph, complete_solution), preserve them unless the verifier feedback requires updating them for consistency with the repaired MCQ.

DO NOT OUTPUT YOUR PROCESS.
Return ONLY the revised JSON object.

Respond EXACTLY in the following format, including the JSON start and end markers:
{
  "question": "<self-contained MCQ stem>",
  "options": { "A": "...", "B": "...", "C": "...", "D": "...", "E": "None of the above" },
  "correct_answer": "<one of: A|B|C|D|E>"
}
"""

USER_PROMPT = """
Input Details:
Previous Candidate Output:
{previous_candidate_output}

Verifier LLM Feedback:
{verifier_llm_feedback}

{difficulty_and_blooms_guidance}

Previously generated questions in this chapter:
{previous_questions}

Chapter Material:
{chapter_material}

Chapter Knowledge Summary:
{chapter_knowledge_text}

Solution Trace:
{solution_trace}
"""
\end{tcblisting}

\subsection{Prompt-level anti-duplication}
\label{app:anti_duplication_prompt}
Below we provide the Anti-duplication instruction prompt.
\begin{tcblisting}{
  enhanced,
  breakable,
  colback=gray!3,
  colframe=black!80,
  boxrule=0.5pt,
  arc=2pt,
  title={Prompt for Anti-duplication Instruction},
  fonttitle=\bfseries,
  listing only,
  listing options={
    basicstyle=\small\ttfamily,
    breaklines=true,
    columns=fullflexible,
    keepspaces=true
  }
}
- You are provided a list of previously generated questions in the input as well.
- Your task is to generate new questions that are NOT near-duplicates of any prior question.
- Treat a question as a near-duplicate if it:
    - tests the same underlying concept or sub-skill, OR
    - relies on the same primary method of solution, OR
    - follows the same reasoning pattern or solution structure, even if the wording, numbers, scenario, or entities differ.
- Do NOT generate questions that could be solved using the same mental steps as any prior question.
\end{tcblisting}

\section{Benchmark Samples}
\label{app:benchmark_samples}
In this section, we present representative examples of chapter summary containing, concepts, algorithms, theorems etc. extracted from our three benchmark domains: Machine Learning, Corporate Finance, and Personal Finance. We also include example problems and corresponding solution traces drawn from the Machine Learning and Finance benchmarks.

\subsection{Example Chapter-level Concepts}
\label{app:chapter_concepts}
\subsubsection{Machine Learning}
\label{app:chapter_concepts_ml}
Below we provide the core-concepts extracted from Chapter 12 of \textit{Understanding Deep Learning} book.
\begin{tcblisting}{
  enhanced,
  breakable,
  colback=violet!3,
  colframe=violet!80,
  boxrule=0.5pt,
  arc=2pt,
  title={Core-concepts from Chapter 12 of \textit{Understanding Deep Learning} book},
  fonttitle=\bfseries,
  listing only,
  listing options={
    basicstyle=\small\ttfamily,
    breaklines=true,
    columns=fullflexible,
    keepspaces=true
  }
}
"core_concepts": [
      {
        "name": "Transformers",
        "description": "A neural network architecture initially targeted at natural language processing that relies entirely on self-attention mechanisms to draw global dependencies between inputs, dispensing with recurrence and convolutions."
      },
      {
        "name": "Self-attention",
        "description": "A mechanism that computes representations of a sequence by relating different positions of the sequence. It routes values in different proportions to create each output, where the routing weights depend on the inputs themselves."
      },
      {
        "name": "Multi-head self-attention",
        "description": "An extension of self-attention where multiple self-attention mechanisms (heads) are applied in parallel, each with its own set of learned parameters, and their outputs are concatenated and linearly transformed."
      },
      {
        "name": "Positional encoding",
        "description": "Information added to the input embeddings (either absolute or relative) to inject knowledge about the order or position of tokens, since the standard self-attention mechanism is permutation equivariant."
      },
      {
        "name": "Transformer layer",
        "description": "A building block of transformer models consisting of a multi-head self-attention unit followed by a fully connected network (MLP), both wrapped with residual connections and LayerNorm operations."
      },
      {
        "name": "Encoder model",
        "description": "A transformer model that transforms input embeddings into a rich representation suitable for downstream tasks (e.g., BERT), typically pre-trained using self-supervision like masked language modeling."
      },
      {
        "name": "Decoder model",
        "description": "A transformer model designed to generate the next token in a sequence (e.g., GPT-3). It uses masked self-attention to prevent attending to future tokens and builds an autoregressive language model."
      },
      {
        "name": "Encoder-decoder model",
        "description": "A transformer architecture used for sequence-to-sequence tasks (e.g., machine translation) where an encoder processes the source sequence and a decoder generates the target sequence while attending to the encoder's output via cross-attention."
      },
      {
        "name": "Vision Transformer (ViT)",
        "description": "An adaptation of the transformer architecture for images that divides an image into a grid of patches, linearly projects them into embeddings, and processes them with standard transformer layers."
      }
    ],
\end{tcblisting}

Below we provide the core-concepts extracted from Chapter 15 of \textit{Probabilistic Machine Learning: An Introduction} book.
\begin{tcblisting}{
  enhanced,
  breakable,
  colback=violet!3,
  colframe=violet!80,
  boxrule=0.5pt,
  arc=2pt,
  title={Core-concepts from Chapter 15 of \textit{Probabilistic Machine Learning: An Introduction} book},
  fonttitle=\bfseries,
  listing only,
  listing options={
    basicstyle=\small\ttfamily,
    breaklines=true,
    columns=fullflexible,
    keepspaces=true
  }
}
"core_concepts": [
      {
        "name": "Recurrent Neural Networks (RNNs)",
        "description": "Neural networks that map input sequences to output sequences in a stateful way, maintaining a hidden state updated over time."
      },
      {
        "name": "Vec2Seq",
        "description": "A sequence generation model mapping a fixed-length input vector to an arbitrary-length sequence of vectors."
      },
      {
        "name": "Seq2Vec",
        "description": "A sequence classification model mapping a variable-length sequence to a single fixed-length output vector."
      },
      {
        "name": "Seq2Seq",
        "description": "A sequence translation model mapping an input sequence to an output sequence, which can be aligned (same length) or unaligned (encoder-decoder architecture)."
      },
      {
        "name": "Bidirectional RNN",
        "description": "An RNN architecture that computes hidden states in both forward and backward directions to capture past and future context."
      },
      {
        "name": "Teacher Forcing",
        "description": "A training technique for language models where the ground truth previous tokens are fed as inputs instead of the model's own generated tokens."
      },
      {
        "name": "Backpropagation Through Time (BPTT)",
        "description": "The algorithm used to compute gradients in RNNs by unrolling the network backwards in time."
      },
      {
        "name": "Gated Recurrent Unit (GRU)",
        "description": "An RNN variant that uses reset and update gates to selectively remember or forget information, mitigating the vanishing gradient problem."
      },
      {
        "name": "Long Short-Term Memory (LSTM)",
        "description": "An RNN variant featuring a memory cell controlled by input, output, and forget gates to capture long-term dependencies."
      },
      {
        "name": "Attention",
        "description": "A mechanism allowing a model to dynamically compute a weighted average of values based on the similarity between a query and a set of keys."
      },
      {
        "name": "Self-Attention",
        "description": "An attention mechanism where the queries, keys, and values are all derived from the same input sequence."
      },
      {
        "name": "Multi-Headed Attention (MHA)",
        "description": "An extension of attention that computes multiple independent attention matrices (heads) in parallel to capture different notions of similarity."
      },
      {
        "name": "Positional Encoding",
        "description": "A technique to inject sequence order information into permutation-invariant models like transformers, often using sinusoidal functions."
      },
      {
        "name": "Transformer",
        "description": "A seq2seq architecture relying entirely on self-attention and feedforward layers, dispensing with recurrence and convolutions."
      },
      {
        "name": "Masked Language Model (MLM)",
        "description": "A training objective where random tokens in a sequence are masked, and the model must predict them using bidirectional context (e.g., BERT)."
      },
      {
        "name": "Large Language Models (LLMs)",
        "description": "Generative (causal) language models based on decoder-only transformers, trained on massive text corpora (e.g., GPT)."
      }
    ]
\end{tcblisting}

\subsubsection{Corporate Finance}
\label{app:chapter_concepts_cp}
Below we provide the core-concepts extracted from Chapter 16 of \textit{Corporate Finance} book.
\begin{tcblisting}{
  enhanced,
  breakable,
  colback=violet!3,
  colframe=violet!80,
  boxrule=0.5pt,
  arc=2pt,
  title={Core-concepts from Chapter 16 of \textit{Corporate Finance} book},
  fonttitle=\bfseries,
  listing only,
  listing options={
    basicstyle=\small\ttfamily,
    breaklines=true,
    columns=fullflexible,
    keepspaces=true
  }
}
"core_concepts": [
      {
        "name": "Capital Structure",
        "description": "The sum total of all claims on the assets of a firm, representing the rights that own all the firm's assets."
      },
      {
        "name": "Cash Flow Rights",
        "description": "The rights that describe how firm-generated cash will be allocated among different claim holders."
      },
      {
        "name": "Control Rights",
        "description": "The rights that allow claim owners to enforce their cash flow rights, such as forcing bankruptcy or electing the corporate board."
      },
      {
        "name": "State-Contingent Claims",
        "description": "Claims, such as debt and equity, whose future values depend on the future state (value) of the underlying firm."
      },
      {
        "name": "Payoff Diagrams",
        "description": "Visual representations that plot the payoffs of financial claims as a function of the underlying firm value at one fixed point in time."
      }
    ],
\end{tcblisting}

\subsubsection{Personal Finance}
\label{app:chapter_concepts_pf}
Below we provide the core-concepts extracted from Chapter 10 of \textit{Strategic Financial Planning Over the Lifecycle} book.
\begin{tcblisting}{
  enhanced,
  breakable,
  colback=violet!3,
  colframe=violet!80,
  boxrule=0.5pt,
  arc=2pt,
  title={Core-concepts from Chapter 10 of \textit{Strategic Financial Planning Over the Lifecycle} book},
  fonttitle=\bfseries,
  listing only,
  listing options={
    basicstyle=\small\ttfamily,
    breaklines=true,
    columns=fullflexible,
    keepspaces=true
  }
}
"core_concepts": [
      {
        "name": "Survivorship Bias",
        "description": "A distortion in performance measurement that occurs when only successful or surviving entities (like mutual funds or lucky investors) are observed, while failed ones are ignored, making it difficult to distinguish true ability from chance."
      },
      {
        "name": "Portfolio Diversification",
        "description": "The practice of spreading investments across multiple assets to reduce the dispersion of possible outcomes (risk) without necessarily sacrificing the expected payoff."
      },
      {
        "name": "Correlation Coefficient",
        "description": "A statistical measure ranging from -1 to +1 that indicates how the returns of two assets co-move with each other. It is a critical factor in determining the risk of a combined portfolio."
      },
      {
        "name": "Modern Portfolio Theory",
        "description": "A theory stating that as the number of assets in a portfolio increases, individual asset variances become negligible, and the portfolio's total risk is primarily driven by the covariances (co-movements) among the assets."
      },
      {
        "name": "Portfolio Frontier",
        "description": "A hyperbolic line on a graph of expected return versus standard deviation that represents all possible risk/return profiles for different weight combinations of a given set of assets."
      }
    ],
\end{tcblisting}

\subsection{Examples from the Benchmarks}
\label{app:solution_graph_examples}
\subsubsection{Machine Learning}
\label{app:examples_ml}
Below we provide an example grounded in Chapter 12 of \textit{Understanding Deep Learning} book.
\begin{tcblisting}{
  enhanced,
  breakable,
  colback=seagreen!3,
  colframe=seagreen!80,
  boxrule=0.5pt,
  arc=2pt,
  title={An example grounded in Chapter 12 of \textit{Understanding Deep Learning} book},
  fonttitle=\bfseries,
  listing only,
  listing options={
    basicstyle=\scriptsize\ttfamily,
    breaklines=true,
    columns=fullflexible,
    keepspaces=true
  }
}
{
  "task_id": "ml_dl_task_000821",
  "task_statement": "Consider a transformer decoder processing a sequence of $N=3$ tokens. The input to a transformer layer is the embedding matrix $X \\in \\mathbb{R}^{D \\times N}$ with $D=4$:\n\n$X = \\begin{bmatrix} 1 & 0 & 0 \\\\ 0 & 1 & 0 \\\\ 0 & 0 & 1 \\\\ 0 & 0 & 0 \\end{bmatrix}$\n\nThe layer uses masked multi-head scaled dot-product self-attention with $H=2$ heads and a causal mask. For each head $i \\in \\{1, 2\\}$, $Q_i = \\Omega_{qi} X$, $K_i = \\Omega_{ki} X$, and $V_i = \\Omega_{vi} X$, with dimension $D_q = D/H = 2$. All biases are zero.\n\nThe weight matrices for Head 1 are:\n$\\Omega_{q1} = \\begin{bmatrix} \\sqrt{2} & 0 & 0 & 0 \\\\ 0 & \\sqrt{2} & 0 & 0 \\end{bmatrix}$\n$\\Omega_{k1} = \\begin{bmatrix} \\ln(2) & \\ln(3) & \\ln(4) & 0 \\\\ \\ln(5) & \\ln(6) & \\ln(7) & 0 \\end{bmatrix}$\n$\\Omega_{v1} = \\begin{bmatrix} 11 & 22 & 33 & 0 \\\\ -11 & -22 & -33 & 0 \\end{bmatrix}$\n\nThe weight matrices for Head 2 are:\n$\\Omega_{q2} = \\begin{bmatrix} 0 & \\sqrt{2} & 0 & 0 \\\\ 0 & 0 & \\sqrt{2} & 0 \\end{bmatrix}$\n$\\Omega_{k2} = \\begin{bmatrix} 0 & \\ln(2) & \\ln(3) & 0 \\\\ \\ln(2) & \\ln(4) & \\ln(6) & 0 \\end{bmatrix}$\n$\\Omega_{v2} = \\begin{bmatrix} 3 & 6 & 9 & 0 \\\\ 6 & 12 & 18 & 0 \\end{bmatrix}$\n\nThe output linear transformation matrix is:\n$\\Omega_c = \\begin{bmatrix} 1 & 1 & 0 & 0 \\\\ 0 & 0 & 1 & -1 \\\\ 1 & 0 & 1 & 0 \\\\ 0 & 1 & 0 & 1 \\end{bmatrix}$\n\nWhat is the second column of the matrix immediately after the first residual connection (i.e., $X + \\Omega_c Z$, where $Z = \\begin{bmatrix} Z_1 \\\\ Z_2 \\end{bmatrix}$ is the vertical concatenation of the head outputs)?\n\nOptions:\nA. $[0, -4, 22, -7]^T$\nB. $[0, -5, 22, -7]^T$\nC. $[1, -5, 22, -7]^T$\nD. $[0, -6, 272/9, -83/9]^T$\nE. None of the above",
  "task_type": "multiple_choice",
  "difficulty": "Hard - Involves complex reasoning, integration of several sub-topics, or solving non-trivial problems that demand deeper conceptual understanding.",
  "bloom_level": "Apply - Use knowledge or methods in new but familiar situations. Example verbs: calculate, demonstrate, use, implement.",
  "choices": [
    {
      "label": "A",
      "solution": "$[0, -4, 22, -7]^T$"
    },
    {
      "label": "B",
      "solution": "$[0, -5, 22, -7]^T$"
    },
    {
      "label": "C",
      "solution": "$[1, -5, 22, -7]^T$"
    },
    {
      "label": "D",
      "solution": "$[0, -6, 272/9, -83/9]^T$"
    },
    {
      "label": "E",
      "solution": "None of the above"
    }
  ],
  "competency": "Transformers & Attention Mechanisms",
  "area_name": "Neural Network Architectures",
  "domain": "ML/DL",
  "book_name": "Understanding Deep Learning (Prince)",
  "chapter_id": "12",
  "correct_answer": "A",
  "solution_graph": {
    "nodes": [
      {
        "id": "V1",
        "content": "Identify that a decoder model uses masked self-attention, meaning token 2 can only attend to tokens 1 and 2. Dot products for token 3 will be set to -infinity."
      },
      {
        "id": "V2",
        "content": "Extract the input vector for the second token: x_2 = [0, 1, 0, 0]^T."
      },
      {
        "id": "V3",
        "content": "Compute queries, keys, and values for Head 1: q_2^(1) = \\Omega_{q1} x_2, K_1 = \\Omega_{k1} X, V_1 = \\Omega_{v1} X."
      },
      {
        "id": "V4",
        "content": "Compute scaled dot products for Head 1: K_1^T q_2^(1) / \\sqrt{D_q} = [\\ln(5), \\ln(6), \\ln(7)]^T."
      },
      {
        "id": "V5",
        "content": "Apply masking and softmax for Head 1: Masked vector is [\\ln(5), \\ln(6), -\\infty]^T. Softmax yields [5/11, 6/11, 0]^T."
      },
      {
        "id": "V6",
        "content": "Compute Head 1 output for token 2: Sa_1[X]_2 = V_1 [5/11, 6/11, 0]^T = [17, -17]^T."
      },
      {
        "id": "V7",
        "content": "Compute queries, keys, and values for Head 2: q_2^(2) = \\Omega_{q2} x_2, K_2 = \\Omega_{k2} X, V_2 = \\Omega_{v2} X."
      },
      {
        "id": "V8",
        "content": "Compute scaled dot products for Head 2: K_2^T q_2^(2) / \\sqrt{D_q} = [0, \\ln(2), \\ln(3)]^T."
      },
      {
        "id": "V9",
        "content": "Apply masking and softmax for Head 2: Masked vector is [0, \\ln(2), -\\infty]^T. Softmax yields [1/3, 2/3, 0]^T."
      },
      {
        "id": "V10",
        "content": "Compute Head 2 output for token 2: Sa_2[X]_2 = V_2 [1/3, 2/3, 0]^T = [5, 10]^T."
      },
      {
        "id": "V11",
        "content": "Concatenate head outputs and apply \\Omega_c: MhSa[X]_2 = \\Omega_c [17, -17, 5, 10]^T = [0, -5, 22, -7]^T."
      },
      {
        "id": "V12",
        "content": "Add the residual connection: x_2 + MhSa[X]_2 = [0, -4, 22, -7]^T."
      }
    ],
    "edges": [
      {
        "from": "V1",
        "to": "V5",
        "operation": "Apply masking rule to Head 1"
      },
      {
        "from": "V1",
        "to": "V9",
        "operation": "Apply masking rule to Head 2"
      },
      {
        "from": "V2",
        "to": "V3",
        "operation": "Use x_2 for Head 1 query"
      },
      {
        "from": "V2",
        "to": "V7",
        "operation": "Use x_2 for Head 2 query"
      },
      {
        "from": "V3",
        "to": "V4",
        "operation": "Matrix multiplication and scaling"
      },
      {
        "from": "V4",
        "to": "V5",
        "operation": "Masking and Softmax"
      },
      {
        "from": "V3",
        "to": "V6",
        "operation": "Use V_1"
      },
      {
        "from": "V5",
        "to": "V6",
        "operation": "Weighted sum of values"
      },
      {
        "from": "V7",
        "to": "V8",
        "operation": "Matrix multiplication and scaling"
      },
      {
        "from": "V8",
        "to": "V9",
        "operation": "Masking and Softmax"
      },
      {
        "from": "V7",
        "to": "V10",
        "operation": "Use V_2"
      },
      {
        "from": "V9",
        "to": "V10",
        "operation": "Weighted sum of values"
      },
      {
        "from": "V6",
        "to": "V11",
        "operation": "Concatenate"
      },
      {
        "from": "V10",
        "to": "V11",
        "operation": "Concatenate"
      },
      {
        "from": "V2",
        "to": "V12",
        "operation": "Add original input"
      },
      {
        "from": "V11",
        "to": "V12",
        "operation": "Add to original input"
      }
    ]
  },
  "complete_solution": "To find the second column of the matrix after the first residual connection, we must compute $x_2 + \\text{MhSa}[X]_2$, where $x_2$ is the second column of $X$ and $\\text{MhSa}[X]_2$ is the output of the multi-head self-attention block for the second token.\n\n**Step 1: Identify the masking rule**\nBecause this is a decoder model, it uses masked self-attention. This means a token can only attend to itself and previous tokens. For the second token ($n=2$), it can attend to tokens 1 and 2. The dot product for token 3 will be set to $-\\infty$ before the softmax operation.\n\n**Step 2: Extract the input vector**\nThe second column of $X$ is $x_2 = [0, 1, 0, 0]^T$.\n\n**Step 3: Compute Head 1 output**\nFirst, compute the query for token 2:\n$q_2^{(1)} = \\Omega_{q1} x_2 = [0, \\sqrt{2}]^T$\nNext, compute the keys and values for all tokens:\n$K_1 = \\Omega_{k1} X = \\begin{bmatrix} \\ln(2) & \\ln(3) & \\ln(4) \\\\ \\ln(5) & \\ln(6) & \\ln(7) \\end{bmatrix}$\n$V_1 = \\Omega_{v1} X = \\begin{bmatrix} 11 & 22 & 33 \\\\ -11 & -22 & -33 \\end{bmatrix}$\nCompute the dot products $K_1^T q_2^{(1)}$:\n$K_1^T q_2^{(1)} = [\\sqrt{2}\\ln(5), \\sqrt{2}\\ln(6), \\sqrt{2}\\ln(7)]^T$\nScale the dot products by $\\sqrt{D_q} = \\sqrt{2}$:\nScaled dot products = $[\\ln(5), \\ln(6), \\ln(7)]^T$\nApply masking (set the 3rd element to $-\\infty$) and then softmax:\nMasked = $[\\ln(5), \\ln(6), -\\infty]^T$\nSoftmax weights = $[\\frac{e^{\\ln(5)}}{5+6}, \\frac{e^{\\ln(6)}}{5+6}, 0]^T = [5/11, 6/11, 0]^T$\nCompute the weighted sum of values for Head 1:\n$Sa_1[X]_2 = V_1 [5/11, 6/11, 0]^T = \\frac{5}{11}\\begin{bmatrix} 11 \\\\ -11 \\end{bmatrix} + \\frac{6}{11}\\begin{bmatrix} 22 \\\\ -22 \\end{bmatrix} = \\begin{bmatrix} 5 \\\\ -5 \\end{bmatrix} + \\begin{bmatrix} 12 \\\\ -12 \\end{bmatrix} = \\begin{bmatrix} 17 \\\\ -17 \\end{bmatrix}$\n\n**Step 4: Compute Head 2 output**\nCompute the query for token 2:\n$q_2^{(2)} = \\Omega_{q2} x_2 = [\\sqrt{2}, 0]^T$\nCompute the keys and values for all tokens:\n$K_2 = \\Omega_{k2} X = \\begin{bmatrix} 0 & \\ln(2) & \\ln(3) \\\\ \\ln(2) & \\ln(4) & \\ln(6) \\end{bmatrix}$\n$V_2 = \\Omega_{v2} X = \\begin{bmatrix} 3 & 6 & 9 \\\\ 6 & 12 & 18 \\end{bmatrix}$\nCompute the dot products $K_2^T q_2^{(2)}$:\n$K_2^T q_2^{(2)} = [0, \\sqrt{2}\\ln(2), \\sqrt{2}\\ln(3)]^T$\nScale the dot products by $\\sqrt{D_q} = \\sqrt{2}$:\nScaled dot products = $[0, \\ln(2), \\ln(3)]^T$\nApply masking and softmax:\nMasked = $[0, \\ln(2), -\\infty]^T$\nSoftmax weights = $[\\frac{e^0}{1+2}, \\frac{e^{\\ln(2)}}{1+2}, 0]^T = [1/3, 2/3, 0]^T$\nCompute the weighted sum of values for Head 2:\n$Sa_2[X]_2 = V_2 [1/3, 2/3, 0]^T = \\frac{1}{3}\\begin{bmatrix} 3 \\\\ 6 \\end{bmatrix} + \\frac{2}{3}\\begin{bmatrix} 6 \\\\ 12 \\end{bmatrix} = \\begin{bmatrix} 1 \\\\ 2 \\end{bmatrix} + \\begin{bmatrix} 4 \\\\ 8 \\end{bmatrix} = \\begin{bmatrix} 5 \\\\ 10 \\end{bmatrix}$\n\n**Step 5: Concatenate and apply $\\Omega_c$**\nVertically concatenate the outputs of the two heads:\n$[Sa_1[X]_2^T, Sa_2[X]_2^T]^T = [17, -17, 5, 10]^T$\nApply the linear transformation $\\Omega_c$:\n$\\text{MhSa}[X]_2 = \\Omega_c \\begin{bmatrix} 17 \\\\ -17 \\\\ 5 \\\\ 10 \\end{bmatrix} = \\begin{bmatrix} 1\\cdot17 + 1\\cdot(-17) + 0\\cdot5 + 0\\cdot10 \\\\ 0\\cdot17 + 0\\cdot(-17) + 1\\cdot5 + (-1)\\cdot10 \\\\ 1\\cdot17 + 0\\cdot(-17) + 1\\cdot5 + 0\\cdot10 \\\\ 0\\cdot17 + 1\\cdot(-17) + 0\\cdot5 + 1\\cdot10 \\end{bmatrix} = \\begin{bmatrix} 0 \\\\ -5 \\\\ 22 \\\\ -7 \\end{bmatrix}$\n\n**Step 6: Add the residual connection**\nThe first residual connection adds the original input back to the output of the multi-head self-attention block:\n$x_2 + \\text{MhSa}[X]_2 = \\begin{bmatrix} 0 \\\\ 1 \\\\ 0 \\\\ 0 \\end{bmatrix} + \\begin{bmatrix} 0 \\\\ -5 \\\\ 22 \\\\ -7 \\end{bmatrix} = \\begin{bmatrix} 0 \\\\ -4 \\\\ 22 \\\\ -7 \\end{bmatrix}$\n\nThe answer is $[0, -4, 22, -7]^T$, which matches option A."
}
\end{tcblisting}

Below we provide an example grounded in Chapter 15 of \textit{Probabilistic Machine Learning: An Introduction} book.
\begin{tcblisting}{
  enhanced,
  breakable,
  colback=seagreen!3,
  colframe=seagreen!80,
  boxrule=0.5pt,
  arc=2pt,
  title={An example grounded in Chapter 15 of \textit{Probabilistic Machine Learning: An Introduction} book},
  fonttitle=\bfseries,
  listing only,
  listing options={
    basicstyle=\scriptsize\ttfamily,
    breaklines=true,
    columns=fullflexible,
    keepspaces=true
  }
}
{
  "task_id": "ml_dl_task_000480",
  "task_statement": "The input matrix $X \\in \\mathbb{R}^{3 \\times 4}$ to a self-attention head has rows:\n$x_1 = [1, 0, 0, 0]$\n$x_2 = [0, 1, 0, 0]$\n$x_3 = [0, 0, 1, 0]$\n\nThe projection matrices are $W_q = W_k = W_v = I_4$.\nThe scaled dot-product attention output is $Z = \\text{softmax}\\left(\\frac{Q K^T}{\\sqrt{d_k}}\\right) V$, where $Q = X W_q$, $K = X W_k$, $V = X W_v$, and $d_k = 4$.\nLet $E = \\text{LayerNorm}(Z + X)$.\nLayer Normalization computes the population mean and variance across the feature dimension for each token independently, with no learned affine parameters ($\\gamma=1$, $\\beta=0$) or smoothing parameter ($\\epsilon=0$).\n\nWhat is the exact value of $E_{1,4}$?\n\nOptions:\nA. $\\frac{-(\\sqrt{e}+2)}{\\sqrt{3e + 4\\sqrt{e} + 2}}$\nB. $\\frac{-(\\sqrt{e}+2)}{\\sqrt{3e - 4\\sqrt{e} + 4}}$\nC. $\\frac{-(e+2)}{\\sqrt{3e^2 + 4e + 2}}$\nD. $\\frac{-(e+2)}{\\sqrt{3e^2 - 4e + 4}}$\nE. None of the above",
  "task_type": "multiple_choice",
  "difficulty": "Hard - Involves complex reasoning, integration of several sub-topics, or solving non-trivial problems that demand deeper conceptual understanding.",
  "bloom_level": "Apply - Use knowledge or methods in new but familiar situations. Example verbs: calculate, demonstrate, use, implement.",
  "choices": [
    {
      "label": "A",
      "solution": "$\\frac{-(\\sqrt{e}+2)}{\\sqrt{3e + 4\\sqrt{e} + 2}}$"
    },
    {
      "label": "B",
      "solution": "$\\frac{-(\\sqrt{e}+2)}{\\sqrt{3e - 4\\sqrt{e} + 4}}$"
    },
    {
      "label": "C",
      "solution": "$\\frac{-(e+2)}{\\sqrt{3e^2 + 4e + 2}}$"
    },
    {
      "label": "D",
      "solution": "$\\frac{-(e+2)}{\\sqrt{3e^2 - 4e + 4}}$"
    },
    {
      "label": "E",
      "solution": "None of the above"
    }
  ],
  "competency": "Recurrent Neural Networks & Sequence Modeling",
  "area_name": "Neural Network Architectures",
  "domain": "ML/DL",
  "book_name": "Probabilistic Machine Learning: An Introduction (Murphy)",
  "chapter_id": "15",
  "correct_answer": "A",
  "solution_graph": {
    "nodes": [
      {
        "id": "V1",
        "content": "Identify Q, K, V matrices from input X and identity projection matrices (Q = K = V = X)."
      },
      {
        "id": "V2",
        "content": "Compute the unscaled dot-product attention scores X X^T, resulting in the 3x3 identity matrix I_3."
      },
      {
        "id": "V3",
        "content": "Scale the attention scores by 1/sqrt(d) = 1/sqrt(4) = 0.5, yielding S = 0.5 * I_3."
      },
      {
        "id": "V4",
        "content": "Apply the softmax function row-wise to S to obtain the attention weights matrix A. For row 1, A_11 = sqrt(e)/(sqrt(e)+2) and A_12 = A_13 = 1/(sqrt(e)+2)."
      },
      {
        "id": "V5",
        "content": "Compute the attention output Z = A X. The first row z_1 is a linear combination of the basis vectors: [p, q, q, 0]."
      },
      {
        "id": "V6",
        "content": "Add the residual connection U = Z + X. The first row becomes u_1 = [1+p, q, q, 0]."
      },
      {
        "id": "V7",
        "content": "Calculate the population mean of u_1 across the feature dimension (d=4), yielding mu = 0.5."
      },
      {
        "id": "V8",
        "content": "Calculate the population variance of u_1 across the feature dimension, yielding sigma^2 = (6q^2 - 8q + 3)/4."
      },
      {
        "id": "V9",
        "content": "Compute the LayerNorm output for the fourth component: E_{1,4} = (u_{1,4} - mu) / sigma = -1 / sqrt(6q^2 - 8q + 3)."
      },
      {
        "id": "V10",
        "content": "Substitute q = 1/(sqrt(e)+2) into the expression and algebraically simplify to obtain the final exact value: -(sqrt(e)+2)/sqrt(3e + 4*sqrt(e) + 2)."
      }
    ],
    "edges": [
      {
        "from": "V1",
        "to": "V2",
        "operation": "Matrix multiplication for dot products"
      },
      {
        "from": "V2",
        "to": "V3",
        "operation": "Scalar division by sqrt(d)"
      },
      {
        "from": "V3",
        "to": "V4",
        "operation": "Row-wise softmax computation"
      },
      {
        "from": "V1",
        "to": "V5",
        "operation": "Matrix multiplication with values"
      },
      {
        "from": "V4",
        "to": "V5",
        "operation": "Matrix multiplication with values"
      },
      {
        "from": "V1",
        "to": "V6",
        "operation": "Vector addition (residual)"
      },
      {
        "from": "V5",
        "to": "V6",
        "operation": "Vector addition (residual)"
      },
      {
        "from": "V6",
        "to": "V7",
        "operation": "Mean calculation"
      },
      {
        "from": "V6",
        "to": "V8",
        "operation": "Variance calculation"
      },
      {
        "from": "V7",
        "to": "V8",
        "operation": "Variance calculation"
      },
      {
        "from": "V6",
        "to": "V9",
        "operation": "Standardization"
      },
      {
        "from": "V7",
        "to": "V9",
        "operation": "Standardization"
      },
      {
        "from": "V8",
        "to": "V9",
        "operation": "Standardization"
      },
      {
        "from": "V9",
        "to": "V10",
        "operation": "Algebraic substitution and simplification"
      }
    ]
  },
  "complete_solution": "Step 1: Compute the queries, keys, and values.\nSince $W_q = W_k = W_v = I_4$, we have $Q = K = V = X$.\n\nStep 2: Compute the scaled dot-product attention scores.\nThe dot products between the queries and keys are given by $X X^\\top$.\nSince the rows of $X$ are orthogonal standard basis vectors, $x_i^\\top x_j = 1$ if $i=j$ and $0$ otherwise.\nThus, $X X^\\top = I_3$ (the $3 \\times 3$ identity matrix).\nThe scaled attention scores are $S = \\frac{X X^\\top}{\\sqrt{d}} = \\frac{1}{\\sqrt{4}} I_3 = 0.5 I_3$.\n\nStep 3: Compute the attention weights.\nWe apply the softmax function row-wise to $S$.\nFor the first row, the scores are $[0.5, 0, 0]$.\nThe softmax probabilities are:\n$p = \\frac{e^{0.5}}{e^{0.5} + e^0 + e^0} = \\frac{\\sqrt{e}}{\\sqrt{e} + 2}$\n$q = \\frac{e^0}{e^{0.5} + e^0 + e^0} = \\frac{1}{\\sqrt{e} + 2}$\nSo the first row of the attention matrix $A$ is $[p, q, q]$.\n\nStep 4: Compute the attention output $Z$.\nThe first row of $Z$ is $z_1 = p x_1 + q x_2 + q x_3$.\nSubstituting the basis vectors:\n$z_1 = p[1, 0, 0, 0] + q[0, 1, 0, 0] + q[0, 0, 1, 0] = [p, q, q, 0]$.\n\nStep 5: Apply the residual connection.\nLet $U = Z + X$. The first row is $u_1 = z_1 + x_1 = [1+p, q, q, 0]$.\n\nStep 6: Apply Layer Normalization.\nWe compute the population mean $\\mu$ and variance $\\sigma^2$ of $u_1$ across its $d=4$ features.\nNote that $p + 2q = \\frac{\\sqrt{e}}{\\sqrt{e} + 2} + \\frac{2}{\\sqrt{e} + 2} = 1$.\nMean: $\\mu = \\frac{1+p + q + q + 0}{4} = \\frac{1 + (p + 2q)}{4} = \\frac{1 + 1}{4} = 0.5$.\nVariance: $\\sigma^2 = \\frac{(1+p - 0.5)^2 + (q - 0.5)^2 + (q - 0.5)^2 + (0 - 0.5)^2}{4}$\n$\\sigma^2 = \\frac{(p + 0.5)^2 + 2(q - 0.5)^2 + 0.25}{4}$.\nSubstitute $p = 1 - 2q$:\n$\\sigma^2 = \\frac{(1.5 - 2q)^2 + 2(q - 0.5)^2 + 0.25}{4}$\n$= \\frac{(2.25 - 6q + 4q^2) + 2(q^2 - q + 0.25) + 0.25}{4}$\n$= \\frac{2.25 - 6q + 4q^2 + 2q^2 - 2q + 0.5 + 0.25}{4}$\n$= \\frac{6q^2 - 8q + 3}{4}$.\nThe standard deviation is $\\sigma = \\frac{\\sqrt{6q^2 - 8q + 3}}{2}$.\n\nStep 7: Compute the normalized value for the fourth component.\n$E_{1,4} = \\frac{u_{1,4} - \\mu}{\\sigma} = \\frac{0 - 0.5}{\\sigma} = \\frac{-0.5}{\\sqrt{6q^2 - 8q + 3}/2} = \\frac{-1}{\\sqrt{6q^2 - 8q + 3}}$.\n\nStep 8: Simplify the expression.\nSubstitute $q = \\frac{1}{\\sqrt{e} + 2}$:\n$6q^2 - 8q + 3 = 6\\left(\\frac{1}{\\sqrt{e} + 2}\\right)^2 - 8\\left(\\frac{1}{\\sqrt{e} + 2}\\right) + 3$\n$= \\frac{6 - 8(\\sqrt{e} + 2) + 3(\\sqrt{e} + 2)^2}{(\\sqrt{e} + 2)^2}$\n$= \\frac{6 - 8\\sqrt{e} - 16 + 3(e + 4\\sqrt{e} + 4)}{(\\sqrt{e} + 2)^2}$\n$= \\frac{-10 - 8\\sqrt{e} + 3e + 12\\sqrt{e} + 12}{(\\sqrt{e} + 2)^2}$\n$= \\frac{3e + 4\\sqrt{e} + 2}{(\\sqrt{e} + 2)^2}$.\nTaking the square root gives $\\sqrt{6q^2 - 8q + 3} = \\frac{\\sqrt{3e + 4\\sqrt{e} + 2}}{\\sqrt{e} + 2}$.\nThus, $E_{1,4} = \\frac{-1}{\\frac{\\sqrt{3e + 4\\sqrt{e} + 2}}{\\sqrt{e} + 2}} = \\frac{-(\\sqrt{e} + 2)}{\\sqrt{3e + 4\\sqrt{e} + 2}}$.\n\nThis matches Option A."
}
\end{tcblisting}

\subsubsection{Corporate Finance}
\label{app:example_cp}
Below we provide an example grounded in Chapter 16 of \textit{Corporate Finance} book.
\begin{tcblisting}{
  enhanced,
  breakable,
  colback=seagreen!3,
  colframe=seagreen!80,
  boxrule=0.5pt,
  arc=2pt,
  title={An example grounded in Chapter 16 of \textit{Corporate Finance} book},
  fonttitle=\bfseries,
  listing only,
  listing options={
    basicstyle=\scriptsize\ttfamily,
    breaklines=true,
    columns=fullflexible,
    keepspaces=true
  }
}
{
  "task_id": "finance_task_000696",
  "task_statement": "A firm's capital structure consists of the following. The market value of all non-convertible liabilities and preferred equity equals their book value:\n- Accounts payable: $150 million\n- Short-term financial debt: $50 million\n- Deferred long-term nonfinancial liabilities: $50 million\n- Senior long-term financial debt: $150 million\n- Preferred equity: $100 million\n- Convertible bonds: 200,000 bonds outstanding\n- Common stock: 15 million shares outstanding\n\nThe current stock price is $20 per share. Each convertible bond can be converted into 50 shares of common stock. The market value of the convertible bonds equals their conversion value.\n\nAll convertible bonds are converted into common stock. The total market value of the firm's assets remains unchanged.\n\nWhat are the firm's market-based total-liabilities-to-total-assets ratio and financial-debt-to-capital ratio immediately after conversion? (Capital is the sum of financial debt and total equity).\n\nOptions:\nA. Total-liabilities-to-total-assets: 40.0%; Financial-debt-to-capital: 25.0%\nB. Total-liabilities-to-total-assets: 40.0%; Financial-debt-to-capital: 40.0%\nC. Total-liabilities-to-total-assets: 60.0%; Financial-debt-to-capital: 50.0%\nD. Total-liabilities-to-total-assets: 50.0%; Financial-debt-to-capital: 37.5%\nE. None of the above",
  "task_type": "multiple_choice",
  "difficulty": "Hard - Involves complex reasoning, integration of several sub-topics, or solving non-trivial problems that demand deeper conceptual understanding.",
  "bloom_level": "Apply - Use knowledge or methods in new but familiar situations. Example verbs: calculate, demonstrate, use, implement.",
  "choices": [
    {
      "label": "A",
      "solution": "Total-liabilities-to-total-assets: 40.0%; Financial-debt-to-capital: 25.0%"
    },
    {
      "label": "B",
      "solution": "Total-liabilities-to-total-assets: 40.0%; Financial-debt-to-capital: 40.0%"
    },
    {
      "label": "C",
      "solution": "Total-liabilities-to-total-assets: 60.0%; Financial-debt-to-capital: 50.0%"
    },
    {
      "label": "D",
      "solution": "Total-liabilities-to-total-assets: 50.0%; Financial-debt-to-capital: 37.5%"
    },
    {
      "label": "E",
      "solution": "None of the above"
    }
  ],
  "competency": "Corporate Securities and Claims",
  "area_name": "Capital Structure",
  "domain": "Finance",
  "book_name": "Corporate_Finance_by_Ivo_Welch",
  "chapter_id": "16",
  "correct_answer": "A",
  "solution_graph": {
    "nodes": [
      {
        "id": "V1",
        "content": "Calculate Market Value of Common Stock: 15 million shares * $20 = $300 million"
      },
      {
        "id": "V2",
        "content": "Calculate Conversion Value of Convertible Bonds: 200,000 bonds * 50 shares/bond * $20/share = $200 million"
      },
      {
        "id": "V3",
        "content": "Calculate Total Market Value of Assets before conversion: $300M (Common) + $200M (Convertible) + $100M (Preferred) + $150M (AP) + $50M (STFD) + $50M (Def) + $150M (LTFD) = $1,000 million"
      },
      {
        "id": "V4",
        "content": "Identify Total Liabilities after conversion: Exclude Convertibles (now equity) and Preferred (equity). Sum remaining = $150M (AP) + $50M (STFD) + $50M (Def) + $150M (LTFD) = $400 million"
      },
      {
        "id": "V5",
        "content": "Calculate Market-based Total-Liabilities-to-Total-Assets Ratio: $400M / $1,000M = 40.0%"
      },
      {
        "id": "V6",
        "content": "Identify Financial Debt after conversion: Sum Short-term financial debt ($50M) and Senior long-term financial debt ($150M) = $200 million"
      },
      {
        "id": "V7",
        "content": "Calculate Equity after conversion: Total Assets ($1,000M) - Total Liabilities ($400M) = $600 million"
      },
      {
        "id": "V8",
        "content": "Calculate Financial Capital: Financial Debt ($200M) + Equity ($600M) = $800 million"
      },
      {
        "id": "V9",
        "content": "Calculate Financial-Debt-to-Capital Ratio: $200M / $800M = 25.0%"
      }
    ],
    "edges": [
      {
        "from": "V1",
        "to": "V3",
        "operation": "Sum market values of all claims to find total asset value"
      },
      {
        "from": "V2",
        "to": "V3",
        "operation": "Sum market values of all claims to find total asset value"
      },
      {
        "from": "V3",
        "to": "V4",
        "operation": "Reclassify convertible bonds as equity; retain non-convertible liabilities"
      },
      {
        "from": "V3",
        "to": "V5",
        "operation": "Apply total-liabilities-to-total-assets ratio formula"
      },
      {
        "from": "V4",
        "to": "V5",
        "operation": "Apply total-liabilities-to-total-assets ratio formula"
      },
      {
        "from": "V4",
        "to": "V6",
        "operation": "Filter total liabilities to isolate only financial debt"
      },
      {
        "from": "V3",
        "to": "V7",
        "operation": "Apply accounting identity: Equity = Assets - Liabilities"
      },
      {
        "from": "V4",
        "to": "V7",
        "operation": "Apply accounting identity: Equity = Assets - Liabilities"
      },
      {
        "from": "V6",
        "to": "V8",
        "operation": "Apply financial capital definition"
      },
      {
        "from": "V7",
        "to": "V8",
        "operation": "Apply financial capital definition"
      },
      {
        "from": "V6",
        "to": "V9",
        "operation": "Apply financial-debt-to-capital ratio formula"
      },
      {
        "from": "V8",
        "to": "V9",
        "operation": "Apply financial-debt-to-capital ratio formula"
      }
    ]
  },
  "complete_solution": "Step 1: Calculate the total market value of the firm's assets before conversion.\nThe total market value of the firm's assets is the sum of the market values of all its claims.\n- Market value of common stock = 15 million shares x $20/share = $300 million.\n- Market value of convertible bonds = 200,000 bonds x 50 shares/bond x $20/share = $200 million.\n- Market value of preferred equity = $100 million.\n- Market value of non-convertible liabilities = $150M (Accounts payable) + $50M (Short-term financial debt) + $50M (Deferred nonfinancial) + $150M (Senior long-term financial debt) = $400 million.\nTotal market value of assets = $300M + $200M + $100M + $400M = $1,000 million.\n\nStep 2: Determine the total liabilities after conversion.\nWhen the convertible bonds are converted, they become equity and are no longer liabilities. Furthermore, preferred equity is classified as equity, not a liability.\nThe remaining liabilities are the non-convertible liabilities: Accounts payable ($150M), Short-term financial debt ($50M), Deferred nonfinancial ($50M), and Senior long-term financial debt ($150M).\nTotal liabilities = $400 million.\n\nStep 3: Calculate the market-based total-liabilities-to-total-assets ratio.\nRatio = Total Liabilities / Total Market Value of Assets = $400M / $1,000M = 40.0%.\n\nStep 4: Determine the financial debt after conversion.\nFinancial debt consists only of short-term financial debt and long-term financial debt. Accounts payable and deferred long-term nonfinancial liabilities are nonfinancial claims and must be excluded.\nFinancial debt = $50M (Short-term) + $150M (Senior long-term) = $200 million.\n\nStep 5: Determine the equity and financial capital after conversion.\nEquity is the residual claim on the firm's assets after all liabilities have been honored.\nEquity = Total Market Value of Assets - Total Liabilities = $1,000M - $400M = $600 million.\n(Alternatively, Equity = $300M original common + $200M new common from conversion + $100M preferred = $600 million.)\nFinancial capital = Financial debt + Equity = $200M + $600M = $800 million.\n\nStep 6: Calculate the financial-debt-to-capital ratio.\nRatio = Financial Debt / Financial Capital = $200M / $800M = 25.0%.\n\nTherefore, the correct option is A: Total-liabilities-to-total-assets: 40.0%; Financial-debt-to-capital: 25.0%."
}
\end{tcblisting}

\subsubsection{Personal Finance}
\label{app:example_pf}
Below we provide an example grounded in Chapter 10 of \textit{Strategic Financial Planning Over the Lifecycle} book.
\begin{tcblisting}{
  enhanced,
  breakable,
  colback=seagreen!3,
  colframe=seagreen!80,
  boxrule=0.5pt,
  arc=2pt,
  title={An example grounded in Chapter 10 of \textit{Strategic Financial Planning Over the Lifecycle} book},
  fonttitle=\bfseries,
  listing only,
  listing options={
    basicstyle=\scriptsize\ttfamily,
    breaklines=true,
    columns=fullflexible,
    keepspaces=true
  }
}
{
  "task_id": "finance_task_000142",
  "task_statement": "An investor has $100,000 and considers two strategies.\n\nStrategy 1:\nThe $100,000 is allocated evenly across $N$ independent roulette tables. At each table, the probability of winning is 0.55. Winning pays twice the bet amount, and losing pays zero. $N$ is chosen such that the probability of a total payoff less than $100,000 is exactly 2.28%.\n\nStrategy 2:\nThe $100,000 is allocated evenly between Asset A and Asset B for one year.\n- Asset A: expected return 6% p.a., standard deviation 8% p.a.\n- Asset B: expected return 14% p.a., standard deviation 12% p.a.\nThe probability of a negative portfolio return ($\\le 0\\%$) is exactly 2.28%.\n\nAssume the total payoff in Strategy 1 and the portfolio return in Strategy 2 are normally distributed, and the z-score for a cumulative probability of 2.28% is -2.00.\n\nWhat are the correlation coefficient ($\\rho_{A,B}$) between the returns of Asset A and Asset B, and the number of tables ($N$) in Strategy 1?\n\nOptions:\nA. $\\rho_{A,B} = -0.5625$, $N = 396$\nB. $\\rho_{A,B} = -0.5625$, $N = 400$\nC. $\\rho_{A,B} = -0.9531$, $N = 396$\nD. $\\rho_{A,B} = 1.0000$, $N = 99$\nE. None of the above",
  "task_type": "multiple_choice",
  "difficulty": "Hard - Involves complex reasoning, integration of several sub-topics, or solving non-trivial problems that demand deeper conceptual understanding.",
  "bloom_level": "Apply - Use knowledge or methods in new but familiar situations. Example verbs: calculate, demonstrate, use, implement.",
  "choices": [
    {
      "label": "A",
      "solution": "$\\rho_{A,B} = -0.5625$, $N = 396$"
    },
    {
      "label": "B",
      "solution": "$\\rho_{A,B} = -0.5625$, $N = 400$"
    },
    {
      "label": "C",
      "solution": "$\\rho_{A,B} = -0.9531$, $N = 396$"
    },
    {
      "label": "D",
      "solution": "$\\rho_{A,B} = 1.0000$, $N = 99$"
    },
    {
      "label": "E",
      "solution": "None of the above"
    }
  ],
  "competency": "Portfolio Diversification and Construction",
  "area_name": "Investing and Portfolio Management",
  "domain": "Finance",
  "book_name": "Strategic_Financial_Planning_Over_the_Lifecycle_Huang_Milevsky_Charupat",
  "chapter_id": "10",
  "correct_answer": "A",
  "solution_graph": {
    "nodes": [
      {
        "id": "V1",
        "content": "Identify target z-score = -2.00 for both strategies based on the 2.28% probability."
      },
      {
        "id": "V2",
        "content": "Calculate expected total payoff E[W] = 110,000 for Strategy 1."
      },
      {
        "id": "V3",
        "content": "Use z-score formula to find required SD[W] = 5,000 for Strategy 1."
      },
      {
        "id": "V4",
        "content": "Set up SD[W] formula in terms of N: SD[W] = (100,000 / sqrt(N)) * sqrt(0.99)."
      },
      {
        "id": "V5",
        "content": "Solve for N = 396."
      },
      {
        "id": "V6",
        "content": "Calculate expected portfolio return E[R_P] = 10% for Strategy 2."
      },
      {
        "id": "V7",
        "content": "Use z-score formula to find required portfolio standard deviation = 5%."
      },
      {
        "id": "V8",
        "content": "Set up portfolio variance formula with given weights (0.5) and standard deviations."
      },
      {
        "id": "V9",
        "content": "Solve for correlation coefficient rho_{A,B} = -0.5625."
      }
    ],
    "edges": [
      {
        "from": "V1",
        "to": "V3",
        "operation": "Apply z-score to Strategy 1 target"
      },
      {
        "from": "V2",
        "to": "V3",
        "operation": "Substitute E[W] into z-score formula"
      },
      {
        "from": "V3",
        "to": "V5",
        "operation": "Equate required SD[W] to formula"
      },
      {
        "from": "V4",
        "to": "V5",
        "operation": "Solve algebraic equation for N"
      },
      {
        "from": "V1",
        "to": "V7",
        "operation": "Apply z-score to Strategy 2 target"
      },
      {
        "from": "V6",
        "to": "V7",
        "operation": "Substitute E[R_P] into z-score formula"
      },
      {
        "from": "V7",
        "to": "V9",
        "operation": "Equate required variance to formula"
      },
      {
        "from": "V8",
        "to": "V9",
        "operation": "Solve algebraic equation for rho_{A,B}"
      }
    ]
  },
  "complete_solution": "Step 1: Analyze Strategy 1 (Roulette Tables)\n- The total investment is $100,000 spread evenly across $N$ tables, so the bet per table is $M = \\$100,000 / N$.\n- The probability of winning is $p = 0.55$.\n- The expected payoff per table is $E[w_i] = 2pM = 2(0.55)(\\$100,000 / N) = \\$110,000 / N$.\n- The total expected payoff is $E[W] = N \\cdot E[w_i] = \\$110,000$.\n- The target is a total payoff of less than $100,000. The z-score for a 2.28% probability is given as -2.00.\n- Using the z-score formula: $z = \\frac{Y - E[W]}{SD[W]} \\implies -2.00 = \\frac{100,000 - 110,000}{SD[W]} \\implies SD[W] = \\frac{-10,000}{-2.00} = 5,000$.\n- The standard deviation of the total payoff is $SD[W] = M \\sqrt{N \\cdot 4p(1-p)}$.\n- Substitute $M = 100,000 / N$ and $p = 0.55$:\n  $SD[W] = \\frac{100,000}{N} \\sqrt{N \\cdot 4(0.55)(0.45)} = \\frac{100,000}{\\sqrt{N}} \\sqrt{0.99}$.\n- Set $SD[W] = 5,000$:\n  $5,000 = \\frac{100,000 \\sqrt{0.99}}{\\sqrt{N}} \\implies \\sqrt{N} = \\frac{100,000 \\sqrt{0.99}}{5,000} = 20 \\sqrt{0.99}$.\n- Square both sides: $N = 400 \\times 0.99 = 396$.\n\nStep 2: Analyze Strategy 2 (Two-Asset Portfolio)\n- The investment is divided evenly, so the weights are $f_A = 0.5$ and $f_B = 0.5$.\n- The expected portfolio return is $E[R_P] = f_A E[r_A] + f_B E[r_B] = 0.5(6\\%) + 0.5(14\\%) = 3\\% + 7\\% = 10\\%$.\n- The target is a negative return (return $\\le 0\\%$). The z-score for a 2.28% probability is -2.00.\n- Using the z-score formula: $z = \\frac{0\\% - E[R_P]}{\\sigma_{Port}} \\implies -2.00 = \\frac{0\\% - 10\\%}{\\sigma_{Port}} \\implies \\sigma_{Port} = \\frac{-10\\%}{-2.00} = 5\\% = 0.05$.\n- The portfolio variance is $\\sigma_{Port}^2 = (0.05)^2 = 0.0025$.\n- The formula for portfolio variance is:\n  $\\sigma_{Port}^2 = f_A^2 \\sigma_A^2 + f_B^2 \\sigma_B^2 + 2 f_A f_B \\rho_{A,B} \\sigma_A \\sigma_B$.\n- Substitute the known values:\n  $0.0025 = (0.5)^2(0.08)^2 + (0.5)^2(0.12)^2 + 2(0.5)(0.5)\\rho_{A,B}(0.08)(0.12)$\n  $0.0025 = 0.25(0.0064) + 0.25(0.0144) + 0.5 \\rho_{A,B} (0.0096)$\n  $0.0025 = 0.0016 + 0.0036 + 0.0048 \\rho_{A,B}$\n  $0.0025 = 0.0052 + 0.0048 \\rho_{A,B}$\n- Solve for $\\rho_{A,B}$:\n  $-0.0027 = 0.0048 \\rho_{A,B} \\implies \\rho_{A,B} = \\frac{-0.0027}{0.0048} = -0.5625$.\n\nConclusion: $\\rho_{A,B} = -0.5625$ and $N = 396$. This matches Option A."
}
\end{tcblisting}



\end{document}